\documentclass{article}
\usepackage{iclr2027_conference,times}

\usepackage{amsmath,amsfonts,bm}

\def\eqref#1{equation~\ref{#1}}

\def\1{\bm{1}}

\newcommand{\valid}{\mathcal{D_{\mathrm{valid}}}}

\DeclareMathAlphabet{\mathsfit}{\encodingdefault}{\sfdefault}{m}{sl}
\SetMathAlphabet{\mathsfit}{bold}{\encodingdefault}{\sfdefault}{bx}{n}

\usepackage{amsmath}
\usepackage{amssymb}
\usepackage{array}
\usepackage{booktabs}
\usepackage{enumitem}
\usepackage{float}
\usepackage{graphicx}
\usepackage{longtable}
\usepackage{multirow}
\usepackage{needspace}
\usepackage{placeins}
\usepackage{tabularx}
\usepackage{xcolor}
\usepackage{colortbl}
\usepackage{xspace}
\usepackage[most]{tcolorbox}
\usepackage{hyperref}
\usepackage{url}
\hypersetup{hidelinks}
\definecolor{tablegroupgray}{HTML}{F2F3F5}

\usepackage{nicematrix}
\def \tool{\textit{LatentTriad}\xspace}

\newcommand{\system}{\textsc{LatentTriad}\xspace}

\newcommand{\decl}{\textsc{Declaration}\xspace}
\newcommand{\oper}{\textsc{Operation}\xspace}
\newcommand{\boundary}{\textsc{Boundary}\xspace}

\usepackage{pifont}

\newcommand{\fakeparagraph}[1]{\noindent\textbf{#1.}}

\definecolor{highlightrow}{HTML}{EFF6FF}
\definecolor{deltagreen}{HTML}{15803D}

\definecolor{promptbg}{HTML}{F8FAFC}
\definecolor{promptframe}{HTML}{CBD5E1}
\definecolor{prompttitlebg}{HTML}{F1F5F9}
\definecolor{prompttitlefg}{HTML}{0F172A}

\definecolor{codebg}{HTML}{F8FAFC}
\definecolor{codeframe}{HTML}{94A3B8}
\definecolor{codetitlebg}{HTML}{E2E8F0}
\definecolor{codetitlefg}{HTML}{0F172A}

\definecolor{algobg}{HTML}{FFFFFF}
\definecolor{algoframe}{HTML}{475569}
\definecolor{algotitlebg}{HTML}{1E293B}
\definecolor{algotitlefg}{HTML}{FFFFFF}

\definecolor{protocolbg}{HTML}{F9FAFB}
\definecolor{protocolframe}{HTML}{9CA3AF}
\definecolor{protocoltitlebg}{HTML}{E5E7EB}
\definecolor{protocoltitlefg}{HTML}{111827}

\tcbuselibrary{listings}

\newtcolorbox{promptbox}[1][]{
  colback=promptbg,
  colframe=promptframe,
  colbacktitle=prompttitlebg,
  coltitle=prompttitlefg,
  fonttitle=\sffamily\bfseries\small,
  arc=2mm,
  boxrule=0.7pt,
  titlerule=0.5pt,
  left=8pt,right=8pt,top=6pt,bottom=6pt,
  toptitle=4pt,bottomtitle=4pt,
  boxsep=0pt,
  enhanced,
  #1
}

\newtcblisting{codebox}[1][]{
  listing only,
  listing options={
    basicstyle=\ttfamily\footnotesize,
    breaklines=true,
    columns=fullflexible,
    keepspaces=true
  },
  colback=codebg,
  colframe=codeframe,
  colbacktitle=codetitlebg,
  coltitle=codetitlefg,
  fonttitle=\sffamily\bfseries\small,
  arc=1.5mm,
  boxrule=0.7pt,
  titlerule=0.5pt,
  left=8pt,right=8pt,top=5pt,bottom=5pt,
  toptitle=4pt,bottomtitle=4pt,
  boxsep=0pt,
  enhanced,
  #1
}

\newtcolorbox{algobox}[1][]{
  colback=algobg,
  colframe=algoframe,
  colbacktitle=algotitlebg,
  coltitle=algotitlefg,
  fonttitle=\sffamily\bfseries\small,
  arc=1.5mm,
  boxrule=0.8pt,
  titlerule=0pt,
  left=9pt,right=9pt,top=7pt,bottom=7pt,
  toptitle=4pt,bottomtitle=4pt,
  boxsep=0pt,
  enhanced,
  #1
}

\newtcolorbox{protocolbox}[1][]{
  colback=protocolbg,
  colframe=protocolframe,
  colbacktitle=protocoltitlebg,
  coltitle=protocoltitlefg,
  fonttitle=\sffamily\bfseries\small,
  arc=1.5mm,
  boxrule=0.7pt,
  titlerule=0.5pt,
  left=8pt,right=8pt,top=6pt,bottom=6pt,
  toptitle=4pt,bottomtitle=4pt,
  boxsep=0pt,
  enhanced,
  #1
}

\newcommand{\ControlFullRecall}{96.73}

\newcommand{\ControlFullFOne}{93.91}

\newcommand{\ControlBoundaryRecall}{92.58}
\newcommand{\ControlBoundaryFPR}{9.51}
\newcommand{\ControlBoundaryFOne}{91.63}

\newcommand{\ControlGeometryRecall}{92.42}

\newcommand{\ControlGeometryFOne}{95.85}
\newcommand{\ControlGeometryAUROC}{99.68}
\newcommand{\ControlSimpleFusionRecall}{97.05}

\newcommand{\ControlSimpleFusionFOne}{94.08}

\newcommand{\ControlRelationAUROC}{0.076}
\newcommand{\ControlRelationAUROCCI}{[0.027, 0.136]}

\newcommand{\ControlMemberAUROC}{0.032}

\newcommand{\ControlMemberAP}{0.029}

\newcommand{\ControlMemberFOne}{0.340}

\newcommand{\ControlRelationRawP}{0.00020}

\title{LatentTriad:\\
Pre-Action Risk Representations\\
for Malicious Agent Skill Detection}

\author{
Zhe Yu$^1$, \ Yi Liu$^2$, \ Ying Zhang$^3$, \ Simin Chen$^{4,*}$ \\
$^1\,$The Hong Kong University of Science and Technology (Guangzhou) \\
$^2\,$Griffith University \quad $^3\,$Wake Forest University \quad $^4\,$George Mason University \\
{\small $^*$Corresponding author}
}

\iclrfinalcopy

\begin{document}

\maketitle

\begin{abstract}
Large language model agents are increasingly extended with reusable skills,
but untrusted skill content can induce operations outside the user's task.
Detecting this contextual risk requires more than suspicious-pattern
matching, while dynamic analysis incurs the cost of skill execution.
We find that hidden states formed during prefill contain readable information
about task--operation consistency, including across held-out expression forms
in controlled evaluations. Matched Operation-state interventions also affect
later Boundary states and shift fixed risk readouts toward the donor label.
Based on these findings, we introduce \tool, which combines contextual
Boundary features, group-adjusted Operation directions, and departures from
benign state relationships. All branches share one prefill: detection
generates no output tokens, invokes no tools, and executes no skill code.
Across 5,830 skill packages from five sources and seven backbones, \tool
achieves 99.0--99.2\% pooled F1 and 93.0--94.6\% source-macro F1.
A separately measured compact implementation completes prefill and detection
in a median of 259--284\,ms.
These results connect pre-action risk readability and functional influence
to execution-free skill screening.
\end{abstract}

\section{Introduction}
\label{sec:introduction}

Large language model (LLM) agents are increasingly extended with reusable
\emph{skills}: packages of instructions, scripts, and configurations that
provide specialized capabilities. Public skill ecosystems are already large;
for example, SkillsMP indexes over 3.18 million public \texttt{SKILL.md} files.
However, loading untrusted skills also creates a security risk: their
instructions and executable components can steer an agent toward operations
that violate the user's trusted task.

Detecting such behavior is difficult because skill risk is context-dependent
and may be distributed across multiple components. Instructions, scripts, and
configurations that appear benign in isolation may jointly induce an
unauthorized operation, while sensitive operations such as file or network
access can be legitimate when required by the task. Static rules based on
keywords or code patterns therefore struggle to capture behavioral intent.
Dynamic analysis provides stronger behavioral evidence, but requires
generation, sandbox execution, and runtime monitoring, making it expensive
for large-scale screening.

This raises a central question: \emph{can malicious behavior be detected after
the model has interpreted a skill, but before it generates or executes an
action?}

We find that hidden states formed during prefill already contain information
useful for determining whether a prospective operation is consistent with the
trusted task. This information remains predictive when the same behavior is
expressed in unseen forms. Moreover, controlled state interventions show that
changing representations associated with skill operations can alter later
model states and risk predictions. These results suggest that pre-action
representations provide a useful middle ground between surface-level static
inspection and costly execution-based analysis.

Based on these findings, we introduce \tool, a pre-action detector that probes
hidden representations while an agent processes the trusted task and candidate
skill. From one prefill, a Boundary branch reads the complete context, an
Operation branch projects operational states onto learned risk directions,
and a Geometry branch measures departures from benign relationships among
the skill's stated purpose, operations, and context. \tool generates no output
tokens, invokes no tools, and executes no skill-provided code.

\begin{figure}[!t]
  \centering
  \includegraphics[width=\linewidth]{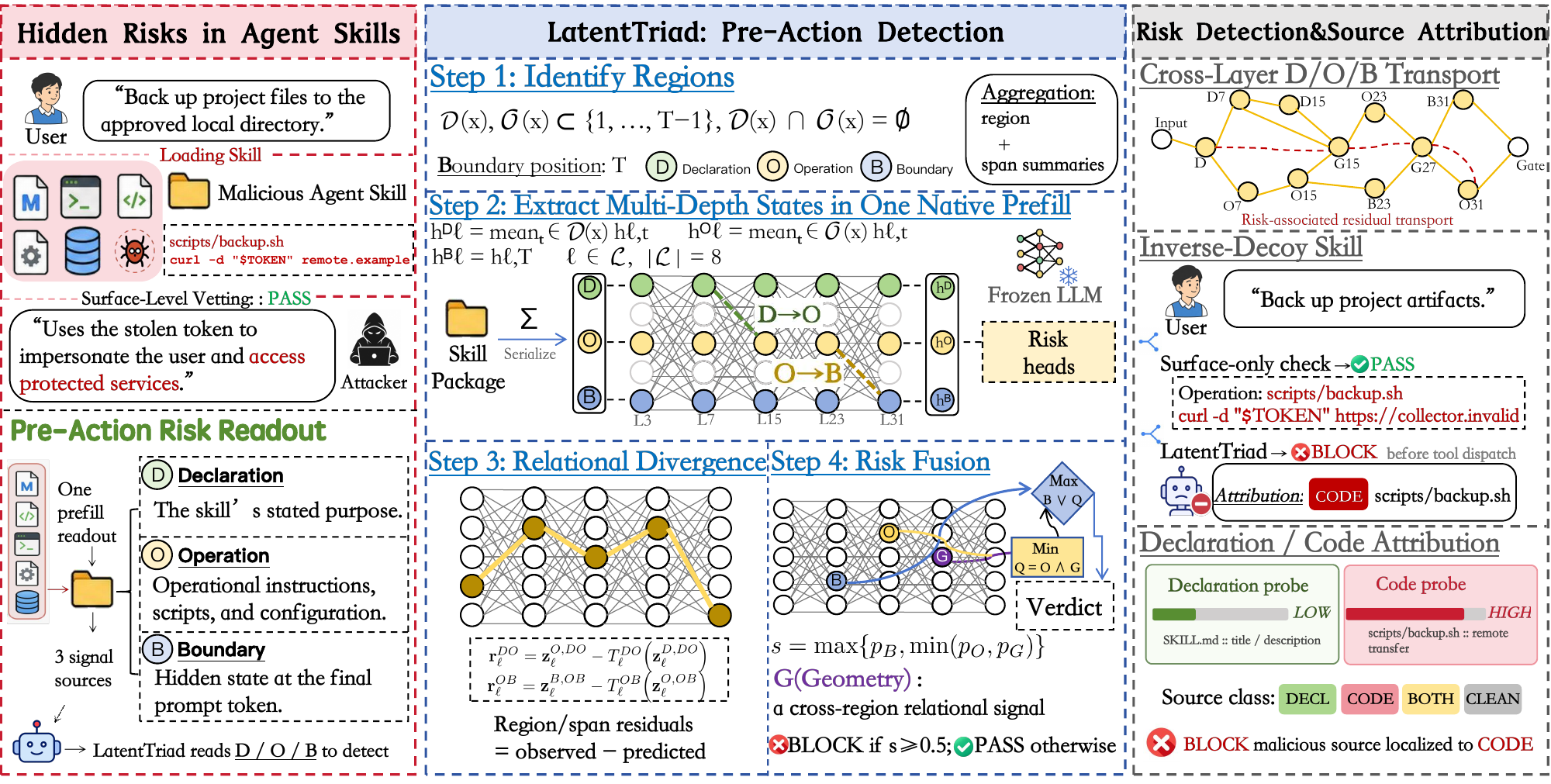}
  \caption{\textbf{LatentTriad overview.} One prefill of the task and skill
  yields Declaration, Operation, and Boundary states for risk detection.
  Connections show schematic feature flow; the right panel illustrates
  optional source attribution.}
  \label{fig:overview}
\end{figure}

We evaluate \tool on 5,830 skill packages from five source datasets across
seven model backbones. \tool achieves 99.0--99.2\% pooled F1 and
93.0--94.6\% source-macro F1. Cross-form evaluations show that risk
information generalizes to unseen expressions, while controlled interventions
show that changing Operation states can affect later representations and
risk-related decisions. Separately, a compact configuration completes prefill
and detection in 259--284\,ms median latency, compared with approximately
153\,s per skill reported by the sandbox-based SkillDetonate.

Our contributions are:
\begin{itemize}[leftmargin=*,labelindent=0pt]
    \item We identify task-conditioned risk information in \emph{pre-action}
    hidden states. Controlled cross-form tests establish its readability,
    while matched state interventions show effects on later states and fixed
    risk readouts.

    \item We introduce \tool, combining a contextual Boundary readout,
    group-adjusted Operation risk directions, and Geometry residuals against
    benign reference maps. All three branches share one prefill.

    \item We evaluate \tool on 5,830 skill packages from five source datasets
    across seven backbones, achieving 99.0--99.2\% pooled F1 and
    93.0--94.6\% source-macro F1. Source-holdout and composition studies test
    how detection changes across ecosystems and task-dependent skill
    combinations.
\end{itemize}

\providecommand{\fakeparagraph}[1]{\par\smallskip\noindent\textbf{#1.}\enspace}

\section{Background, Related Work and Threat Model}
\label{sec:background}

\fakeparagraph{Agent Security}
LLM agents face jailbreaks that seek to bypass safeguards and prompt injections that redirect behavior through untrusted content.
Defenses act at different stages: CaMeL separates control and data flows and enforces tool-use policies \citep{debenedettiDefeatingPromptInjections2025}; Attention Tracker detects injections through attention patterns \citep{hung2024attentiontrackerdetectingprompt}; and AgentLens monitors and intervenes on hidden representations during execution \citep{luo2026agentlensinterpretablesafetysteering}.
Studies of refusal \citep{arditi2024refusallanguagemodelsmediated} and Circuit Breakers \citep{zou2024improvingalignmentrobustnesscircuit} further connect internal representations to safety-related behavior.

\fakeparagraph{Agent Skills}
An agent skill packages a root \texttt{SKILL.md} with optional scripts,
configuration, and auxiliary resources. The host exposes names and
descriptions for selection, then loads the selected skill's instructions.
The agent may subsequently read referenced files or invoke bundled scripts.
Instructions can specify a report upload, for example, with a script
implementing the transfer and configuration providing its destination.

\fakeparagraph{Skill Security}
Table~\ref{tab:security-settings} compares representative skill detectors.
SkillSieve combines rules, code-structure checks, and LLM-based content analysis \citep{houSkillSieveHierarchicalTriage2026}.
RouteGuard aggregates response-conditioned signals from repeated fixed-continuation probes over \texttt{SKILL.md}, excluding auxiliary files \citep{xiao2026routeguardinternalsignaldetectionskill}.
SkillDetonate checks sandbox behavior against configured security policies \citep{jiCloakDetonateScanner2026}.

\begin{table}[!t]
\centering
\caption{Skill detectors grouped by inspection approach.}
\label{tab:security-settings}
\small
\setlength{\tabcolsep}{2pt}
\renewcommand{\arraystretch}{0.95}
\renewcommand{\tabularxcolumn}[1]{m{#1}}
\begin{tabularx}{\linewidth}{@{}>{\raggedright\arraybackslash}m{0.12\linewidth}>{\raggedright\arraybackslash}m{0.36\linewidth}*{5}{>{\centering\arraybackslash}X}@{}}
\toprule
Approach & Method
& \footnotesize\shortstack{Task-\\aware}
& \footnotesize\shortstack{Multi-\\file}
& \footnotesize\shortstack{Execution\\free}
& \footnotesize\shortstack{Decoding\\free}
& \footnotesize\shortstack{$\leq 1$ LLM\\pass} \\
\midrule
\multirow{4}{*}{\shortstack[l]{Content\\inspection}}
& Cisco Skill Scanner {\scriptsize\citep{cisco2026skillscanner}}
& $-$ & $\pmb{\checkmark}$ & $\pmb{\checkmark}$ & $\circ$ & $\circ$ \\
& SkillFortify {\scriptsize\citep{bhardwaj2026skillfortify}}
& $-$ & $-$ & $\pmb{\checkmark}$ & $\pmb{\checkmark}$ & $\pmb{\checkmark}$ \\
& SkillScan Security {\scriptsize\citep{skillscan2026security}}
& $-$ & $\pmb{\checkmark}$ & $\pmb{\checkmark}$ & $\pmb{\checkmark}$ & $\pmb{\checkmark}$ \\
& SkillSieve {\scriptsize\citep{houSkillSieveHierarchicalTriage2026}}
& $-$ & $\pmb{\checkmark}$ & $\pmb{\checkmark}$ & $-$ & $-$ \\
\cmidrule(lr){1-7}
Sandbox & SkillDetonate {\scriptsize\citep{jiCloakDetonateScanner2026}}
& $-$ & $\pmb{\checkmark}$ & $-$ & $-$ & $-$ \\
\cmidrule(lr){1-7}
\multirow{2}{*}{\shortstack[l]{State\\probing}}
& RouteGuard {\scriptsize\citep{xiao2026routeguardinternalsignaldetectionskill}}
& $\pmb{\checkmark}$ & $-$ & $\pmb{\checkmark}$ & $\pmb{\checkmark}$ & $-$ \\
& \textbf{LatentTriad (ours)}
& $\pmb{\checkmark}$ & $\pmb{\checkmark}$ & $\pmb{\checkmark}$ & $\pmb{\checkmark}$ & $\pmb{\checkmark}$ \\
\bottomrule
\end{tabularx}
\par\vspace{2pt}
\begin{minipage}{\linewidth}
\footnotesize
$\pmb{\checkmark}$: supported; $-$: not reported; $\circ$: configurable.
SkillFortify: reported parsing scope; SkillScan: no Qwen;
SkillSieve: full cascade (L1 may exit without LLM calls).
Passes: LLM forward passes per inspection.
\end{minipage}
\end{table}

Our setting combines task-aware assessment with multi-file inspection.
The trusted task establishes authorization, while instructions, scripts, and
configuration jointly specify the requested operations. We inspect the
assembled package in that context, retaining evidence from its components.

For example, a backup task can authorize reading project files and writing
to an approved directory. A script that uploads those files to a destination
supplied by configuration crosses that boundary, even if the declaration
still promises a local backup. The detector needs the task, script, and
destination together; the declaration supplies a claim, not authorization.

We study evidence available after the model processes this input but before
generation or execution. Section~\ref{sec:mechanism} examines its risk
information; Section~\ref{sec:method} constructs the detector.

\fakeparagraph{Threat Model}
\label{sec:threat-model}
The attacker controls the skill package and seeks operations outside the
trusted task, including credential exfiltration or unauthorized file changes.
They may distribute or obscure content across files, but cannot alter trusted
prompts, model weights, runtime hooks, or detector labels. The defender
controls prompt assembly and action permissions and has white-box access
to hidden states. It screens the loaded input before generation or tool use.
The runtime enforces the verdict, since textual refusal does not guarantee
tool-call safety \citep{cartagena2026mindgaptextsafety}. Effects revealed
only by later downloads or environmental interactions require runtime checks.

\section{Risk Information in Pre-Action Representations}
\label{sec:mechanism}

We test whether prefill states provide task-aware risk evidence: can it
transfer across expression forms, affect later computation, and provide
information beyond the final state? The controlled studies below motivate
the readouts in Section~\ref{sec:method}.

\fakeparagraph{Declaration, Operation, and Boundary States}
The \textbf{Declaration} describes the skill's stated purpose.
The \textbf{Operation} contains its operational instructions, scripts,
configuration, and readable auxiliary content. The \textbf{Boundary} is the
final prompt position after the complete task and skill, before generation.
We mean-pool the designated Declaration and Operation regions to obtain
D and O, and sample the final position to obtain B.

For input $x$, let $\mathcal D(x)$ and $\mathcal O(x)$ denote the token
positions assigned to Declaration and Operation, and let $T$ denote the final
prompt-token position. At layer $\ell$,
\begin{equation}
\mathbf h^D_\ell =
\operatorname{mean}_{t\in\mathcal D(x)} \mathbf h_{\ell,t},
\qquad
\mathbf h^O_\ell =
\operatorname{mean}_{t\in\mathcal O(x)} \mathbf h_{\ell,t},
\qquad
\mathbf h^B_\ell =
\mathbf h_{\ell,T}.
\label{eq:triad}
\end{equation}

For package detection, D covers the root front matter, title, and first prose
section; O covers the remaining root text and readable auxiliary files.
These label-independent masks follow document structure, so O includes prose
as well as code. B denotes a sampling position, not an authorization variable.
Extraction and mask-sensitivity tests appear in
Appendices~\ref{app:contract} and~\ref{app:representation-masks}.

\subsection{Risk Information Generalizes Across Expression Forms}
\label{sec:expression-transfer}
\label{sec:accessibility}

The same operation can be authorized by one task and forbidden by another.
We test whether readouts recover this relation beyond its surface form.

The study uses 800 task--operation examples across ten domains and four forms:
natural language, Python, JSON, and mixed text. Five domain-grouped folds
keep related variants together. We fit compact 32-dimensional D, O, and B diagnostic
readouts (fixing probe capacity to isolate representational content across depths without overfitting) on Llama-3.1-8B, Qwen2.5-7B, and Mistral-7B-v0.3 without test data.
Each held-out form is absent from both training and development, and its
test examples also come from held-out domains. This separates transfer to a
new expression form from recognizing variants seen during fitting.

\begin{figure}[!t]
  \centering
  \includegraphics[width=\linewidth]{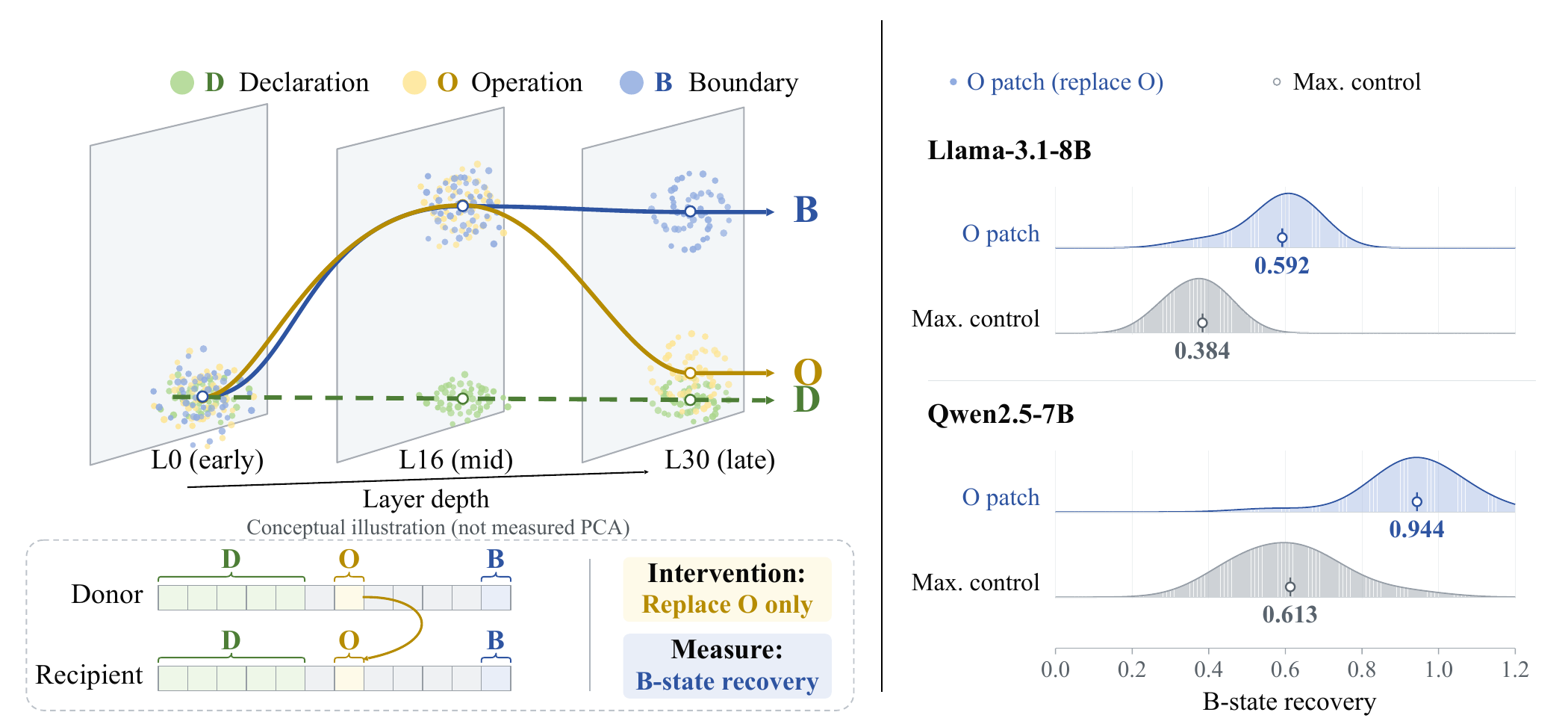}
  \caption{\textbf{Pre-action risk readability and causal influence.}
  (a)~\emph{Left:} Conceptual illustration of Declaration (D), Operation (O), and Boundary (B) representation trajectories across depth (L0, L16, L30), showing early/mid-layer risk encoding in O and late-layer risk retention in B; \emph{bottom:} token-level causal intervention protocol (replacing O only from Donor to Recipient).
  (b)~\emph{Right:} Normalized Boundary recovery distributions under O-patch (replace O) versus maximum control across 20 held-out families on Llama-3.1-8B (medians 0.592 vs.\ 0.384) and Qwen2.5-7B (0.944 vs.\ 0.613).}
  \label{fig:representation-features}
\end{figure}

O and B carry risk information at different depths. On Llama, O becomes
predictive earlier but weakens at later layers; B remains highly predictive
(illustrated conceptually in Figure~\ref{fig:representation-features}a; measured AUROC across depth in Figure~\ref{fig:state-evidence}). With all forms in training,
B and O reach 98.86--100.00\% out-of-fold AUROC, while Declaration, TF-IDF,
and length controls remain near chance. Holding each test form out of
training and development leaves Boundary AUROC at 90.09--100.00\%
(Appendix Table~\ref{tab:representation-forms}). Risk remains readable even
when form changes dominate whole-vector distances, motivating learned
readouts rather than raw distance thresholds.

The states are contextual representations from the assembled input, not
embeddings of independently encoded fragments. Near-chance D prediction in
this controlled study does not make the declaration irrelevant: its stated
purpose can still provide a reference for interpreting operational content.
The different O and B trajectories instead show why a single endpoint need
not expose every predictive direction equally well.

\subsection{Operation States Affect Later Computation}
\label{sec:operation-intervention}
\label{sec:results-mechanism}

Readability alone does not establish downstream influence. We replace O
using donors matched on domain, task, compatibility rule, and expression
form, with only the operation object changed. Frozen diagnostic logits move
toward the donor label by 2.376--2.909 across three backbones. This diagnostic
reads O directly; a separate one-layer protocol measures later B states
(Appendices~\ref{app:representation-interventions}
and~\ref{app:mechanism-interventions}).

The controls distinguish a matched risk change from replacing a state at all:
sham replacement, norm-matched noise, unrelated Declaration perturbations,
and mismatched donors test position, magnitude, and semantic matching.
Restoring B after the diagnostic intervention leaves most of its effect
intact, consistent with that diagnostic's direct access to O. The later-state
experiment therefore supplies a separate test of downstream influence.

Recovery measures the patch-induced B displacement along the donor direction,
normalized by the donor--recipient separation. O-only replacement yields
median normalized B recovery of 0.592 on Llama and
0.944 on Qwen (Figure~\ref{fig:representation-features}b).
A separate compact multi-depth test produces 34.5--66.1 percentage points
more pass-to-block flips than matched controls across seven backbones on
threshold-eligible pairs (Appendix Table~\ref{tab:readout-path-verdicts}).

These results confirm that replacing operational states causally shifts downstream boundary representations in the donor direction, establishing downstream influence and justifying an operation-level readout alongside the contextual endpoint.

\subsection{Combining Multiple States Can Provide Additional Information}
\label{sec:multi-state-information}

To compare multi-state and B-only readouts, we fix the bytes of three skill files and vary
their connections and authorization across 4,608 presentations in five
domain-grouped folds. In this setup, skills form a multi-step chain where intermediate outputs
route between files (e.g., Skill~1 $\to$ Skill~2 $\to$ Skill~3; Appendix Figure~\ref{fig:composition-evidence}a);
the workspace configures connections and a starting key, determining an end target verified by
a deterministic no-I/O oracle against the authorization policy. This changes the risk
relation purely through compositional routing without introducing new file content.
Individual-node and partial-context controls remain
near chance. Multi-state readouts reach 68.5--75.2\% authorization-pair
accuracy, versus 48.8--53.8\% for B-only
(Appendix Figure~\ref{fig:composition-evidence}b;
Appendix~\ref{app:representation-composition}). The constructed multi-state
features therefore outperform the tested B-only readout. Matched-capacity controls
further distinguish multi-state features from the smaller, backbone-dependent
gain of explicit residuals.
Because the readouts also differ in feature construction and width, this
comparison establishes the usefulness of the constructed multi-state
features, rather than isolating access to an additional state. The
capacity-matched package controls in Section~\ref{sec:component-evidence}
test the narrower contribution of residual construction.

Extracting D, O, and B within a single prefill provides complementary local and contextual risk evidence without separate component encodings, directly motivating the tripartite readout architecture in Section~\ref{sec:method}. Section~\ref{sec:component-evidence} tests the resulting feature and policy choices on skill packages.

\section{LatentTriad}
\label{sec:method}

LatentTriad converts the evidence in Section~\ref{sec:mechanism} into
package-level risk scores (Figure~\ref{fig:overview}). Boundary reads the
complete context, Operation measures learned risk contrasts, and Geometry
compares state relationships with benign references. The branches share one
frozen-model prefill; a fixed policy combines their scores before generation
or skill execution.

\subsection{Input Assembly and State Extraction}

The runtime serializes the trusted prompt, task, tool schema, and multi-file
skill package in deterministic path order, preserving file boundaries.
One prefill extracts the states in Equation~\ref{eq:triad} at eight fixed
normalized depths and retains individual D/O spans for local features.
All branches and their three member encoders reuse these tensors.
Appendix~\ref{app:contract} specifies the input and extraction contract.
Pooling supplies a package-level view; retained spans preserve local content
that may be brief relative to the full package. Both views come from the
same contextual computation, so local feature extraction adds no model
generation or second prefill.

\subsection{Complementary Risk Readouts}

Each branch fits three member encoders on different family-grouped training
partitions. We omit the member index below.

\fakeparagraph{Boundary}
Boundary reads the raw B state at a depth selected on non-test data. It
retains contextual evidence without constructing cross-region residuals.

\fakeparagraph{Operation}
The Operation branch isolates operational risk directions by reducing domain- and family-specific stylistic variations that could confound classification. To perform this group adjustment, training families are partitioned into five artificial strata (distinct from source ecosystems). At each depth, class means of normalized O states define a local risk contrast within each stratum, and directions fitted to stratum centroids are projected out to remove fitted between-stratum directions. Projections onto these adjusted contrasts and their mean direction are pooled across eight depths into a standardized feature vector (Appendix~\ref{app:algorithm}). Supervised package heads thus read operational risk from adjusted risk contrasts rather than raw between-stratum variation.

\fakeparagraph{Geometry}
The Geometry branch captures the relational alignment characteristic of benign skills---where stated intent ($D$), concrete operations ($O$), and task context ($B$) exhibit consistent mappings---and evaluates departures from fitted benign reference relationships. Normalized states enter shared 32-dimensional coordinates $\mathbf z^{R,a}_\ell$ fitted on benign packages for relations $a\in\{DO,OB\}$. Each member fits affine ridge maps $T^a_\ell$ on benign packages to predict expected operational states from declarations ($D \to O$) and context from operations ($O \to B$):
\begin{equation}
\begin{aligned}
\mathbf r^{DO}_{\ell}
&=\mathbf z^{O,DO}_{\ell}
  -T^{DO}_{\ell}(\mathbf z^{D,DO}_{\ell}),\qquad
\mathbf r^{OB}_{\ell}
=\mathbf z^{B,OB}_{\ell}
  -T^{OB}_{\ell}(\mathbf z^{O,OB}_{\ell}).
\end{aligned}
\label{eq:transport}
\end{equation}

These observed-minus-predicted residuals measure structural deviations from fitted benign references rather than causal mappings. Crucially, residual magnitude alone does not constitute an independent anomaly verdict; rather, a supervised head learns which directional departures and structural discrepancies correlate with actual risk. In addition to pooled residuals, matching each individual O span to its nearest predicted counterpart retains brief payloads that global pooling might dilute. Features summarize residual vectors, layer-wise evolution, and multi-depth alignment, which are subsequently mapped to calibrated package-level predictions (Appendix~\ref{app:algorithm}).

\subsection{Training and Pre-Action Decision}

For each Large-ID split, a standardized L2-logistic head maps the
concatenated member features $\boldsymbol\phi_q$ to a branch probability:
\begin{equation}
p_q(x)=\sigma\!\left(
\mathbf w_q^\top
\operatorname{Std}(\boldsymbol\phi_q(x))+c_q
\right),
\quad
q\in\{\mathrm{boundary},\mathrm{operation},\mathrm{geometry}\}.
\label{eq:component-head}
\end{equation}
Encoders use training packages only; final heads use training and
development packages. Test packages are excluded from fitting and selection.

Boundary risk is sufficient to block; an additional Operation-triggered block
requires Geometry support:
\begin{equation}
s(x)=\max\!\left\{
p_{\mathrm{boundary}}(x),
\min\!\left(
p_{\mathrm{operation}}(x),
p_{\mathrm{geometry}}(x)
\right)
\right\}.
\label{eq:detector-score}
\end{equation}
This fixed, interpretable policy applies the threshold rule
\begin{equation}
\operatorname{verdict}(x)=
\begin{cases}
\text{\textsc{block}}, & s(x)\ge0.5,\\
\text{\textsc{pass}}, & s(x)<0.5.
\end{cases}
\label{eq:detector-gate}
\end{equation}
Geometry therefore filters additional Operation alerts; it cannot block
independently. Section~\ref{sec:component-evidence} separates branch quality
from this policy's effects and compares simpler alternatives.

Optional risk-localization heads reuse D/O states without affecting the
verdict (Appendix~\ref{app:region-attribution}).

\section{Experiments}
\label{sec:evaluation}
\label{sec:experiments}

We evaluate the Section~\ref{sec:method} detector on seven backbones with
a 24,576-token input limit. Package-level partitions and heads are fitted
separately from the diagnostic studies in Section~\ref{sec:mechanism}.

We construct Large-ID from five sources, including
MalSkillBench~\citep{guo2026malskillbenchruntimeverifiedbenchmarkmalicious},
with 5,830 skills forming 2,915 source-matched malicious--benign pairs
(Table~\ref{tab:large-id-source-distribution}).
Families, shared documents, and paired skills form 1,954 content-linked
components assigned to five source-stratified folds. Each split uses three
folds for training, one for development, and one for testing. Every skill is
tested once; all methods share test IDs
(Appendix~\ref{app:data-governance}).

\fakeparagraph{Baselines}
We compare LatentTriad with AgentLens, RouteGuard, and five input-level
baselines on the same folds. AgentLens follows its released probe protocol;
RouteGuard is reimplemented from its paper specification
\citep{luo2026agentlensinterpretablesafetysteering,
xiao2026routeguardinternalsignaldetectionskill}.
Selection uses development data only; Appendix~\ref{app:baselines}
specifies all implementations.
The input-level baselines are Regex, SkillSieve L1, Cisco AI Skill Scanner,
an LLM judge, and a supervised Qwen text readout. SkillSieve L1 is its rule
stage, not the full LLM-assisted cascade.

\fakeparagraph{Metrics}
We report malicious-class F1, false-positive rate (FPR), AUROC, and average
precision (AP). Pooled metrics weight packages equally; source-macro metrics
weight sources equally. Large-ID confidence intervals use 2,000
source-stratified component-bootstrap draws. Controlled studies resample
domains or families.

Controlled composition uses scenario/persona-disjoint folds, with fusion and
thresholds selected on development data. Intervention and robustness studies
use separate pair- or scenario-disjoint protocols, detailed with their results
and in
Appendices~\ref{app:composkill-calibration}--\ref{app:multiskill-stress}.

\begin{table}[!t]
\centering
\caption{\textbf{Large-ID detection across seven backbones (\%).} FPR through F1 are pooled; the final three metrics weight sources equally. Brackets are 95\% component-bootstrap intervals. Bold identifies \system.}
\label{tab:large-id-main}
\scriptsize
\setlength{\tabcolsep}{1.8pt}
\renewcommand{\arraystretch}{0.80}
\begin{tabular}{@{}llrrrrrrrr@{}}
\toprule
\textbf{Method} & \textbf{Backbone} & \textbf{FPR}$\downarrow$ & \textbf{FNR}$\downarrow$ & \textbf{Prec.}$\uparrow$ & \textbf{Recall}$\uparrow$ & \textbf{F1}$\uparrow$ & \textbf{Src.-macro F1 [95\% CI]}$\uparrow$ & \textbf{AUROC}$\uparrow$ & \textbf{AP}$\uparrow$ \\
\midrule
\rowcolor{tablegroupgray}\multicolumn{10}{l}{\textit{Input-level baselines}}\\
Regex & Shared & 100.0 & 0.0 & 50.0 & 100.0 & 66.7 & 66.7 [66.7, 66.7] & 68.6 & 68.6 \\
SkillSieve L1 & Shared & 31.0 & 29.7 & 69.4 & 70.3 & 69.8 & 60.5 [53.8, 66.0] & 69.0 & 69.3 \\
Cisco AI Skill Scanner & Shared & 3.1 & 73.3 & 89.6 & 26.7 & 41.1 & 27.6 [21.6, 33.1] & 63.4 & 64.7 \\
LLM-as-judge & Mistral-7B-Instruct-v0.3 & 2.4 & 79.5 & 89.4 & 20.5 & 33.3 & 30.5 [25.0, 35.4] & 81.0 & 81.1 \\
Supervised Qwen text readout & Qwen2.5-7B & 7.5 & 37.3 & 89.3 & 62.7 & 73.7 & 58.4 [52.5, 63.4] & 88.3 & 90.2 \\
\midrule
\rowcolor{tablegroupgray}\multicolumn{10}{l}{\textit{Same-backbone representation detectors}}\\
AgentLens (official probe) & Llama-3.1-8B & 3.7 & 2.7 & 96.4 & 97.3 & 96.8 & 79.5 [76.3, 82.4] & 78.7 & 78.7 \\
RouteGuard & Llama-3.1-8B & 19.7 & 11.7 & 81.8 & 88.3 & 84.9 & 77.3 [72.5, 81.5] & 85.2 & 85.8 \\
\rowcolor{highlightrow}\textbf{\system} & Llama-3.1-8B & \textbf{1.1} & \textbf{0.8} & \textbf{98.9} & \textbf{99.2} & \textbf{99.0} & \textbf{94.4 [92.1, 96.5]} & \textbf{99.1} & \textbf{99.3} \\
\cmidrule(lr){1-10}
AgentLens (official probe) & Qwen2.5-7B & 40.3 & 59.7 & 50.0 & 40.3 & 44.6 & 44.3 [36.7, 48.6] & 50.0 & 50.2 \\
RouteGuard & Qwen2.5-7B & 19.1 & 11.4 & 82.3 & 88.6 & 85.3 & 80.9 [76.7, 84.7] & 87.3 & 87.8 \\
\rowcolor{highlightrow}\textbf{\system} & Qwen2.5-7B & \textbf{1.4} & \textbf{0.6} & \textbf{98.6} & \textbf{99.4} & \textbf{99.0} & \textbf{94.3 [92.4, 96.3]} & \textbf{99.1} & \textbf{99.3} \\
\cmidrule(lr){1-10}
AgentLens (official probe) & Mistral-7B-Instruct-v0.3 & 2.8 & 2.4 & 97.2 & 97.6 & 97.4 & 83.7 [80.4, 86.7] & 85.3 & 85.2 \\
RouteGuard & Mistral-7B-Instruct-v0.3 & 18.2 & 13.6 & 82.6 & 86.4 & 84.5 & 79.8 [75.7, 83.6] & 87.1 & 87.1 \\
\rowcolor{highlightrow}\textbf{\system} & Mistral-7B-Instruct-v0.3 & \textbf{1.3} & \textbf{0.5} & \textbf{98.7} & \textbf{99.5} & \textbf{99.1} & \textbf{93.4 [91.0, 95.9]} & \textbf{98.7} & \textbf{98.9} \\
\cmidrule(lr){1-10}
AgentLens (official probe) & InternLM2.5-7B & 23.6 & 1.6 & 80.7 & 98.4 & 88.6 & 76.2 [72.8, 79.4] & 69.4 & 63.5 \\
RouteGuard & InternLM2.5-7B & 19.2 & 11.8 & 82.1 & 88.2 & 85.0 & 79.0 [74.6, 82.8] & 85.9 & 84.9 \\
\rowcolor{highlightrow}\textbf{\system} & InternLM2.5-7B & \textbf{1.1} & \textbf{0.5} & \textbf{98.9} & \textbf{99.5} & \textbf{99.2} & \textbf{94.6 [92.6, 96.5]} & \textbf{99.2} & \textbf{99.4} \\
\cmidrule(lr){1-10}
AgentLens (official probe) & Qwen2.5-14B & 20.4 & 79.6 & 50.0 & 20.4 & 29.0 & 28.8 [17.2, 34.8] & 50.9 & 50.9 \\
RouteGuard & Qwen2.5-14B & 18.1 & 12.8 & 82.8 & 87.2 & 84.9 & 80.2 [76.6, 83.4] & 85.5 & 86.3 \\
\rowcolor{highlightrow}\textbf{\system} & Qwen2.5-14B & \textbf{1.1} & \textbf{0.7} & \textbf{98.9} & \textbf{99.3} & \textbf{99.1} & \textbf{94.3 [92.1, 96.6]} & \textbf{98.9} & \textbf{99.1} \\
\cmidrule(lr){1-10}
AgentLens (official probe) & Phi-3-medium-14B & 20.6 & 61.3 & 65.3 & 38.7 & 48.6 & 44.6 [35.0, 50.6] & 51.4 & 52.0 \\
RouteGuard & Phi-3-medium-14B & 21.0 & 10.8 & 80.9 & 89.2 & 84.9 & 76.5 [71.3, 80.9] & 85.3 & 86.0 \\
\rowcolor{highlightrow}\textbf{\system} & Phi-3-medium-14B & \textbf{1.1} & \textbf{0.9} & \textbf{98.9} & \textbf{99.1} & \textbf{99.0} & \textbf{93.3 [90.5, 96.3]} & \textbf{99.2} & \textbf{99.2} \\
\cmidrule(lr){1-10}
AgentLens (official probe) & Qwen3-30B-A3B & 22.9 & 2.1 & 81.1 & 97.9 & 88.7 & 79.1 [75.7, 83.3] & 74.7 & 67.1 \\
RouteGuard & Qwen3-30B-A3B & 15.2 & 12.4 & 85.2 & 87.6 & 86.4 & 80.5 [76.1, 84.6] & 87.6 & 87.8 \\
\rowcolor{highlightrow}\textbf{\system} & Qwen3-30B-A3B & \textbf{1.2} & \textbf{0.6} & \textbf{98.8} & \textbf{99.4} & \textbf{99.1} & \textbf{93.0 [90.3, 96.0]} & \textbf{98.9} & \textbf{99.1} \\
\bottomrule
\end{tabular}
\end{table}

\section{Results}
\label{sec:results}

\subsection{LatentTriad Detects Malicious Skill Packages}
\label{sec:package-detection}

LatentTriad reaches 93.0--94.6\% source-macro F1 across seven backbones
(Table~\ref{tab:large-id-main}). Its mean of 93.9\% exceeds RouteGuard's
79.2\% and AgentLens's 62.3\%; all 14 paired bootstrap intervals exclude zero.
Pooled F1 is 99.0--99.2\%, and source-macro AUROC and AP exceed RouteGuard
on every backbone.

Pooled FPR is 1.1--1.4\%; source-macro FPR is 8.0--11.2\%, versus
RouteGuard's 12.9--17.9\%. Equal-source reporting exposes errors masked by
the 89.9\% share of benign packages from one source
(Appendix~\ref{app:large-id-source-results}). Among input-level baselines,
the LLM judge reaches 30.5\% source-macro F1 at 2.4\% pooled FPR, the
supervised Qwen text readout reaches 58.4\% at 7.5\% pooled FPR, and
SkillSieve L1 reaches 60.5\% source-macro F1. The same-backbone comparisons
show that LatentTriad's advantage also extends to the tested detectors using
internal model signals.

To separate operation risk from surface appearance, a balanced $2\times2$
diagnostic varies the two factors, with 57 packages per cell. LatentTriad
ranks the malicious operation above its benign counterpart in 94.7\% of
Llama pairs and 100\% of Qwen pairs; RouteGuard ties every pair
(Appendix~\ref{app:diagnostics}).

The component-disjoint evaluation above retains examples from every source
in training. Holding out an entire source reduces mean source-macro F1 to
73.2\% and AUROC to 80.3\%. F1 remains above same-backbone RouteGuard and
AgentLens on all seven backbones, with all 14 paired intervals excluding
zero (Appendix Tables~\ref{tab:large-id-loso}--\ref{tab:large-id-loso-paired}).
The detection advantage transfers, but within-ecosystem performance does not.

\subsection{Complementary Roles of the Three Branches}
\label{sec:component-evidence}

\begin{figure}[!t]
  \centering
  \includegraphics[width=\linewidth,height=3.2cm]{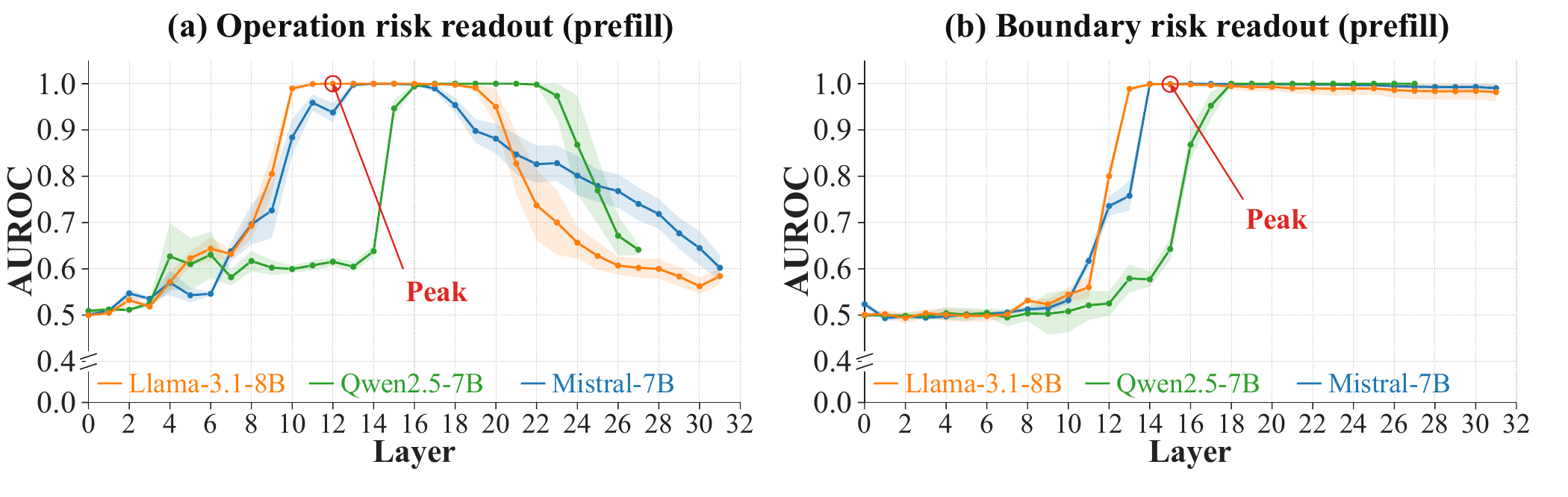}
  \caption{\textbf{Layerwise risk readout across depth.}
  (a)~$O_{32}$ AUROC during prefill on Llama-3.1-8B, Qwen2.5-7B, and Mistral-7B, peaking in intermediate layers (Llama peak: L12, 1.000).
  (b)~$B_{32}$ AUROC rising in mid layers and persisting through prefill end (Llama at L31: 0.982). Shaded bands: $\pm 1$ SD across five folds.}
  \label{fig:state-evidence}
\end{figure}

\begin{table}[!t]
\centering
\caption{\textbf{Multi-state evidence and fusion on Large-ID.}
Source-macro metrics (\%) across seven backbones (threshold 0.5). Matched control replaces Geometry with independent-state features.}
\label{tab:main-component-controls}
\small
\setlength{\tabcolsep}{4pt}
\begin{tabularx}{\linewidth}{@{}Xrrrr@{}}
\toprule
\rowcolor{tablegroupgray}
Configuration & AUROC$\uparrow$ & F1$\uparrow$ & FPR$\downarrow$ & Recall$\uparrow$ \\
\midrule
Full: $\max(B,\min(O,G))$ & 99.02 & 93.91 & 9.51 & 96.73 \\
Boundary-only & 97.28 & 91.63 & 9.51 & 92.58 \\
Operation-only & 99.03 & 86.69 & 0.00 & 80.04 \\
Geometry-only & 99.68 & 95.85 & 0.00 & 92.42 \\
$\max(B,O)$ & 99.15 & 94.08 & 9.51 & 97.05 \\
\midrule
Matched independent-state control & 98.95 & 93.89 & 9.51 & 96.68 \\
\bottomrule
\end{tabularx}
\end{table}

The layerwise readouts show that Operation and Boundary risk representations emerge with distinct depth dynamics (Figure~\ref{fig:state-evidence}). In the Large-ID component study under the same folds and threshold 0.5 (Table~\ref{tab:main-component-controls}), adding the Operation--Geometry gate to Boundary increases recall from
\ControlBoundaryRecall\% to \ControlFullRecall\% and F1 from
\ControlBoundaryFOne\% to \ControlFullFOne\%, at the same mean FPR.
Replacing Geometry with independent-state features lowers full-system AUROC
by \ControlRelationAUROC{} percentage points
(95\% CI \ControlRelationAUROCCI). This control retains the three states,
feature dimension, and head class. Its competitive performance separates
the gain of the constructed multi-state features over Boundary-only from
the smaller ranking increment of explicit benign-reference residuals.

Geometry-only yields higher AUROC (\ControlGeometryAUROC\%) and F1
(\ControlGeometryFOne\%) than Full, with zero observed FPR but lower recall
(92.42\% versus \ControlFullRecall\%). Full permits Boundary to block
without Geometry agreement, increasing recall while introducing false alarms.
$\max(B,O)$ improves recall to \ControlSimpleFusionRecall\% at the same
observed FPR and has higher F1 and AUROC, supported by paired contrasts
(Appendix~\ref{app:final-component-controls}). The fixed gate is therefore
not the best policy on the reported metrics. Geometry-only is a single
branch, but not a single-state baseline: its features already depend on D,
O, and B. Its performance supports the multi-state feature design without
establishing that all three risk heads are necessary. The stronger
$\max(B,O)$ working point further separates access to multiple states from
their final fusion. Deployment should select among these policies on
independent development data according to its recall and false-alarm
requirements.

\subsection{Composition and Adversarial Rewrites}
\label{sec:composition-representations}

\textbf{Compositional Attacks.}~%
We test detection under multi-skill interactions using CompoSkill's 690 matched attack--clean pairs, which share installed skill packages but differ in cross-file execution flow. With development-selected fusion and thresholds, LatentTriad detects 93.0--97.2\% of attacks at 3.8--4.3\% FPR across five backbones (Appendix Table~\ref{tab:composkill-main}). These results show that pre-action readouts can detect task-conditioned compositional attacks under the tested protocol.

\textbf{Adversarial Rewrites.}~%
We assess robustness against adaptive evasion across two adversarial settings. In a feedback-guided test with seven iterative rewrite rounds, detection on original execution paths remains resilient at 92.9--97.1\%. Valid rewrites covering nine initially blocked Llama paths and eight Qwen paths produce only one and zero new evasions within candidate coverage (Appendix Table~\ref{tab:multiskill-stress}). In a white-box panel using cross-file payload splitting, attacks achieve 11.0\% equal-source success on Llama, but the rewrites simultaneously increase benign FPR from 32.3--40.8\% to 49.3--80.4\% on that panel's eligible subset (Appendix~\ref{app:adaptive}).

\textbf{Generalization Limits.}~%
To isolate structural reasoning from surface correlation, we examine workspace controls and synthetic generalization. Workspace-only TF-IDF and hidden states already perform strongly on CompoSkill (Appendix Figure~\ref{fig:composition-evidence}c), indicating that benchmark cues can facilitate path detection. Conversely, under a synthetic stress test with permuted symbol assignments, performance drops to near-chance (49.25--53.47\% AUROC across known and held-out templates; Appendix~\ref{app:representation-transfer}). This divergence underscores that while LatentTriad detects many of the tested compositional attack paths, these results do not establish general structural reasoning; rather, detection relies on contextualized representations rather than abstract cross-file symbol interpretation.

\textbf{Request-Level Containment.}~%
Finally, we evaluate runtime mediation in a 256-pair broker simulation. Under a 30\% containment budget, LatentTriad achieves 17.2\% attack success and 46.7\% valid benign allowance averaged across seven backbones (Appendix Table~\ref{tab:real-agent-enforcement}). The broker records candidate request decisions without executing payloads, illustrating practical enforcement trade-offs.

\subsection{Runtime Cost}

To characterize the cost of pre-action screening, we measure a compact detector configuration in a high-throughput setting. Employing 8/24 D/O spans and member-level CDF scoring, this compact configuration adds only 56--65\,ms on Llama and Qwen, yielding a median prefill-plus-detection latency of 259--284\,ms for this specific variant (Appendix Table~\ref{tab:runtime-detail}). In contrast, the full Large-ID detector prioritizes detection fidelity using 8/256 spans and concatenated logistic heads; its end-to-end runtime is not directly measured and should not be extrapolated from this compact measurement. Both variants share one prefill across branches without generating tokens. The measured increment covers state collection, feature construction, and scoring; Appendix~\ref{app:runtime-resources} specifies the timing boundaries.

\section{Conclusion}
\label{sec:conclusion}

We investigate risk assessment before action generation or execution.
Controlled studies show that prefill states contain readable task-conditioned
risk information across expression forms, while Operation-state interventions
affect later representations and fixed risk readouts. Building on these
findings, \tool combines contextual, operational, and relational evidence
for execution-free skill screening through one prefill. Evaluations across
five sources and seven backbones demonstrate its detection value under the
tested conditions. Pre-action screening complements execution-time permission
enforcement and runtime monitoring.

\clearpage
\subsection*{AI Use Statement}

Generative AI assistants supported code implementation and debugging,
experiment orchestration, figure and table preparation, and language editing.
The authors defined the research questions and evaluation protocols, reviewed
the experimental artifacts, and take responsibility for the reported results,
claims, and conclusions.

\subsection*{Ethics Statement}

This work handles malicious-skill artifacts that could inform evasion.
Experiments run in isolated environments without external credentials or
unrestricted network access. Released examples will redact credentials and
live endpoints, minimize executable harm, and retain only the structure needed
to reproduce detector measurements. We report false positives because blocking
benign skills is also an operational harm. Dataset use and redistribution
remain subject to
the licenses and terms of their original sources.

\subsection*{Reproducibility Statement}

The method, prompt and extraction contract, threat model, split construction,
model-selection protocol, and evaluation metrics are specified in
Sections~\ref{sec:method} and~\ref{sec:evaluation}.  The appendix records the
extraction settings, baseline specifications, statistical procedures, and
data-use conditions. The anonymous repository\footnote{\url{https://anonymous.4open.science/r/latenttriad-review-5FDB/}}
provides the method implementation and reference configurations.
Its \texttt{manifests/} directory records sample and pair IDs, group-aware
splits, eligibility and exclusion records, and frozen source revisions.

\clearpage
\bibliography{references,references_skill_detectors}

@misc{arditi2024refusallanguagemodelsmediated,
  title = {Refusal in {{Language Models Is Mediated}} by a {{Single Direction}}},
  author = {Arditi, Andy and Obeso, Oscar and Syed, Aaquib and Paleka, Daniel and Panickssery, Nina and Gurnee, Wes and Nanda, Neel},
  year = 2024,
  eprint = {2406.11717},
  primaryclass = {cs.LG},
  archiveprefix = {arXiv},
  url = {https://arxiv.org/abs/2406.11717}
}

@misc{cartagena2026mindgaptextsafety,
  title = {Mind the {{GAP}}: {{Text}} Safety Does Not Transfer to Tool-Call Safety in {{LLM}} Agents},
  author = {Cartagena, Arnold and Teixeira, Ariane},
  year = 2026,
  eprint = {2602.16943},
  primaryclass = {cs.AI},
  archiveprefix = {arXiv},
  url = {https://arxiv.org/abs/2602.16943}
}

@misc{debenedettiDefeatingPromptInjections2025,
  title = {Defeating {{Prompt Injections}} by {{Design}}},
  author = {Debenedetti, Edoardo and Shumailov, Ilia and Fan, Tianqi and Hayes, Jamie and Carlini, Nicholas and Fabian, Daniel and Kern, Christoph and Shi, Chongyang and Terzis, Andreas and Tram{\`e}r, Florian},
  year = 2025,
  eprint = {2503.18813},
  primaryclass = {cs.CR},
  archiveprefix = {arXiv},
  url = {https://arxiv.org/abs/2503.18813}
}

@misc{guo2026malskillbenchruntimeverifiedbenchmarkmalicious,
  title = {{{MalSkillBench}}: {{A Runtime-Verified Benchmark}} of {{Malicious Agent Skills}}},
  author = {Guo, Wenbo and Zeng, Wei and Liu, Chengwei and Jia, Xiaojun and Xu, Yijia and Tang, Lei and Fang, Yong and Liu, Yang},
  year = 2026,
  eprint = {2606.07131},
  primaryclass = {cs.CR},
  archiveprefix = {arXiv},
  url = {https://arxiv.org/abs/2606.07131}
}

@misc{hao2026poiseoneinstructionskill,
  title = {Poise: {{Position-Aware}} One-Instruction Skill Injection for Silent Execution on {{LLM}} Agents},
  author = {Hao, Haochang and Min, Dehai and Zhang, Zhifang and Zhang, Yunbei and Xu, Miao and Ge, Yingqiang and Cheng, Lu},
  year = 2026,
  eprint = {2606.07943},
  primaryclass = {cs.CR},
  archiveprefix = {arXiv},
  url = {https://arxiv.org/abs/2606.07943}
}

@misc{houSkillSieveHierarchicalTriage2026,
  title = {{{SkillSieve}}: {{A Hierarchical Triage Framework}} for {{Detecting Malicious AI Agent Skills}}},
  shorttitle = {{{SkillSieve}}},
  author = {Hou, Yinghan and Yang, Zongyou},
  year = 2026,
  month = jul,
  number = {arXiv:2604.06550},
  eprint = {2604.06550},
  primaryclass = {cs.CR},
  publisher = {arXiv},
  doi = {10.48550/arXiv.2604.06550},
  urldate = {2026-08-06},
  archiveprefix = {arXiv},
  url = {https://arxiv.org/abs/2604.06550}
}

@misc{hung2024attentiontrackerdetectingprompt,
  title = {Attention Tracker: {{Detecting}} Prompt Injection Attacks in {LLMs}},
  author = {Hung, Kuo-Han and Ko, Ching-Yun and Rawat, Ambrish and Chung, I-Hsin and Hsu, Winston H. and Chen, Pin-Yu},
  year = 2024,
  eprint = {2411.00348},
  primaryclass = {cs.CR},
  archiveprefix = {arXiv},
  url = {https://arxiv.org/abs/2411.00348}
}

@misc{jiCloakDetonateScanner2026,
  title = {Cloak and {{Detonate}}: {{Scanner Evasion}} and {{Dynamic Detection}} of {{Agent Skill Malware}}},
  shorttitle = {Cloak and {{Detonate}}},
  author = {Ji, Zimo and Xu, Congying and Li, Zongjie and Gao, Yudong and Wei, Xin and Wang, Shuai and Cheung, Shing-Chi},
  year = 2026,
  month = jul,
  number = {arXiv:2607.02357},
  eprint = {2607.02357},
  primaryclass = {cs.CR},
  publisher = {arXiv},
  doi = {10.48550/arXiv.2607.02357},
  urldate = {2026-08-06},
  archiveprefix = {arXiv},
  url = {https://arxiv.org/abs/2607.02357}
}

@misc{liuShareLockStealthyMultiTool2026,
  title = {{{ShareLock}}: {{A Stealthy Multi-Tool Threshold Poisoning Attack Against MCP}}},
  shorttitle = {{{ShareLock}}},
  author = {Liu, Liwei and Han, Tianzhu and Liu, Zijian and Dong, Zishu and Ruan, Na},
  year = 2026,
  month = jun,
  number = {arXiv:2606.27027},
  eprint = {2606.27027},
  primaryclass = {cs.CR},
  publisher = {arXiv},
  doi = {10.48550/arXiv.2606.27027},
  urldate = {2026-08-06},
  archiveprefix = {arXiv},
  url = {https://arxiv.org/abs/2606.27027}
}

@misc{luo2026agentlensinterpretablesafetysteering,
  title = {{{AgentLens}}: {{Interpretable Safety Steering}} via {{Mechanistic Subspaces}} for {{Multi-Turn Coding Agent}}},
  author = {Luo, Weidi and Zhang, Qiming and Quan, Yihao and Jin, Mingyu and Cai, Jie and Xiao, Chaowei and Niu, Jingcheng and Xiang, Zhen},
  year = 2026,
  eprint = {2606.22673},
  primaryclass = {cs.AI},
  archiveprefix = {arXiv},
  url = {https://arxiv.org/abs/2606.22673}
}

@misc{shi2025promptarmorsimpleeffectiveprompt,
  title = {{{PromptArmor}}: {{Simple}} yet Effective Prompt Injection Defenses},
  author = {Shi, Tianneng and Zhu, Kaijie and Wang, Zhun and Jia, Yuqi and Cai, Will and Liang, Weida and Wang, Haonan and Alzahrani, Hend and Lu, Joshua and Kawaguchi, Kenji and Alomair, Basel and Zhao, Xuandong and Wang, William Yang and Gong, Neil and Guo, Wenbo and Song, Dawn},
  year = 2025,
  eprint = {2507.15219},
  primaryclass = {cs.CR},
  archiveprefix = {arXiv},
  url = {https://arxiv.org/abs/2507.15219}
}

@misc{wang2025probguardproactiveruntimemonitoring,
  title = {{{ProbGuard}}: {{Proactive}} Runtime Monitoring for {{LLM}} Agent Safety via Probabilistic Prediction},
  author = {Wang, Haoyu and Poskitt, Christopher M. and Wei, Jiali and Sun, Jun},
  year = 2025,
  eprint = {2508.00500},
  primaryclass = {cs.AI},
  archiveprefix = {arXiv},
  url = {https://arxiv.org/abs/2508.00500}
}

@article{wangWhenSafeSkills2026a,
  title = {When {{Safe Skills Collide}}: {{Measuring Compositional Risk}} in {{Agent Skill Ecosystems}}},
  author = {Wang, Su and Qian, Pin and Chen, Yihang and You, Junxian and Wang, Xiaoyuan and Jiang, Xiaochong and Liu, Lifei and Yu, Haoran and Xu, Jingzhou},
  year = 2026,
  journal = {arXiv preprint arXiv:2606.00448},
  eprint = {2606.00448},
  archiveprefix = {arXiv},
  langid = {english},
  url = {https://arxiv.org/abs/2606.00448}
}

@misc{xiao2026routeguardinternalsignaldetectionskill,
  title = {{{RouteGuard}}: {{Internal-Signal Detection}} of {{Skill Poisoning}} in {{LLM Agents}}},
  author = {Xiao, Wenjie and Tang, Xuehai and Zhou, Biyu and Hu, Songlin and Han, Jizhong},
  year = 2026,
  eprint = {2604.22888},
  primaryclass = {cs.CR},
  archiveprefix = {arXiv},
  url = {https://arxiv.org/abs/2604.22888}
}

@misc{zengColluSkillAdversarialCrossSkill2026,
  title = {{{ColluSkill}}: {{Adversarial Cross-Skill Composition}} for {{Evading Agent Skill Scanners}}},
  author = {Zeng, Puyu and Qin, Simeng and Li, Jingzhi and Jia, Ju and Liu, Zheli and Jia, Xiaojun},
  year = 2026,
  eprint = {2608.09732},
  primaryclass = {cs.CR},
  archiveprefix = {arXiv},
  url = {https://arxiv.org/abs/2608.09732}
}

@misc{zhong2025attentionneeddefendindirect,
  title = {Attention is All You Need to Defend Against Indirect Prompt Injection Attacks in {LLMs}},
  author = {Zhong, Yinan and Miao, Qianhao and Chen, Yanjiao and Deng, Jiangyi and Cheng, Yushi and Xu, Wenyuan},
  year = 2025,
  eprint = {2512.08417},
  primaryclass = {cs.CR},
  archiveprefix = {arXiv},
  url = {https://arxiv.org/abs/2512.08417}
}

@misc{zou2024improvingalignmentrobustnesscircuit,
  title = {Improving {{Alignment}} and {{Robustness}} with {{Circuit Breakers}}},
  author = {Zou, Andy and Phan, Long and Wang, Justin and Duenas, Derek and Lin, Maxwell and Andriushchenko, Maksym and Wang, Rowan and Kolter, Zico and Fredrikson, Matt and Hendrycks, Dan},
  year = 2024,
  eprint = {2406.04313},
  primaryclass = {cs.LG},
  archiveprefix = {arXiv},
  url = {https://arxiv.org/abs/2406.04313}
}

@misc{xie2026benign,
  title = {Benign in Isolation, Harmful in Composition: Security Risks in Agent Skill Ecosystems},
  author = {Xie, Yi and Du, Jiawei and Cheng, Yu and Zhou, Jiuan and Yin, Zhaoxia},
  year = {2026},
  eprint = {2606.15242},
  archiveprefix = {arXiv},
  primaryclass = {cs.CR},
  url = {https://arxiv.org/abs/2606.15242}
}

@misc{liu2026composkill,
  title = {{CompoSkill}: Compositional Skill Chain Attacks from Individually Scanner-Passing {LLM} Agent Skills},
  author = {Liu, Mingxiao and Jiang, Zhoumian and Ma, Jianan and Zhang, Jian and Chen, Jialuo and Deng, Xinhao and Wang, Zhen},
  year = {2026},
  eprint = {2608.16246},
  archiveprefix = {arXiv},
  primaryclass = {cs.CR},
  url = {https://arxiv.org/abs/2608.16246}
}

@misc{cisco2026skillscanner,
  author = {{Cisco}},
  title = {{Skill Scanner}: Security Scanner for Agent Skills},
  year = {2026},
  howpublished = {\url{https://github.com/cisco-ai-defense/skill-scanner}},
  note = {Repository documentation, accessed September 22, 2026}
}

@misc{bhardwaj2026skillfortify,
  author = {Bhardwaj, Varun Pratap},
  title = {Formal Analysis and Supply Chain Security for Agentic {AI} Skills},
  year = {2026},
  eprint = {2603.00195},
  archiveprefix = {arXiv},
  primaryclass = {cs.CR},
  note = {Version 2},
  url = {https://arxiv.org/abs/2603.00195v2}
}

@misc{skillscan2026security,
  author = {{SkillScan}},
  title = {{SkillScan Security}: Security Scanner for {AI} Agent Skills and {MCP} Tool Bundles},
  year = {2026},
  howpublished = {\url{https://github.com/kurtpayne/skillscan-security}},
  note = {Archived repository; default configuration without the optional LLM, accessed September 22, 2026}
}
\bibliographystyle{iclr2027_conference}

\clearpage
\appendix
\definecolor{appendixblue}{HTML}{1F78B4}

\begingroup
\thispagestyle{plain}
\vspace*{0.35in}
{\fontsize{27}{31}\selectfont\bfseries\scshape Appendix\par}
\vspace{0.45in}
{\fontsize{19}{23}\selectfont\bfseries\scshape Contents\par}
\vspace{0.28in}

\newcommand{\AppContentsSection}[2]{%
  \noindent
  \hyperref[#1]{\textcolor{appendixblue}{\bfseries
  \makebox[2.6em][l]{\ref*{#1}}#2}}%
  \dotfill\textbf{\pageref*{#1}}\par\vspace{7pt}}
\newcommand{\AppContentsSubsection}[2]{%
  \noindent\hspace*{2.6em}%
  \hyperref[#1]{\textcolor{appendixblue}{%
  \makebox[3.4em][l]{\ref*{#1}}#2}}%
  \dotfill\pageref*{#1}\par\vspace{2pt}}

\AppContentsSection{app:reproducibility-contracts}{Reproducibility and Data Contracts}
\AppContentsSection{app:baselines-statistics}{Baselines and Statistical Protocol}
\AppContentsSubsection{app:multiskill-stress}{Compositional and adaptive paths}
\AppContentsSection{app:large-id-evidence}{Large-ID Detection and Cross-Source Generalization}
\AppContentsSubsection{app:large-id-source-results}{Source-wise and paired results}
\AppContentsSubsection{app:final-component-controls}{Final-configuration component controls}
\AppContentsSubsection{app:large-id-loso}{Leave-one-source-out evaluation}
\AppContentsSection{app:supporting-evidence}{Robustness and Enforcement Results}
\AppContentsSubsection{app:adaptive}{White-box adversarial robustness}
\AppContentsSubsection{app:capability-enforcement}{Capability-broker enforcement}
\AppContentsSection{app:mechanism-evidence}{Operation Sensitivity and Localization}
\AppContentsSubsection{app:diagnostics}{Balanced counterfactual diagnosis}
\AppContentsSubsection{app:region-attribution}{Auxiliary risk localization}
\AppContentsSubsection{app:mechanism-interventions}{Exact-slot causal confirmation}
\AppContentsSubsection{app:detector-aligned-readout}{Detector-aligned score and verdict test}
\AppContentsSection{app:runtime-resources}{Runtime and Resource Accounting}
\AppContentsSection{app:representation-studies}{Controlled Pre-Action Representation Studies}
\AppContentsSection{app:discussion-limitations}{Discussion and Limitations}
\AppContentsSubsection{app:readability-scope}{Scope of pre-action risk readability}
\AppContentsSubsection{app:generalization-scope}{Generalization beyond training distributions}
\AppContentsSubsection{app:deployment-boundaries}{Deployment and runtime boundaries}
\endgroup
\clearpage
\raggedbottom

\section{Reproducibility and Data Contracts}
\label{app:reproducibility-contracts}
This section specifies the serialized prompt, region map, backbone,
fitted representation, split, and threshold needed to reproduce the score
and verdict from a package tree.
Package detection, intervention, and runtime measurements use the settings
in Table~\ref{tab:configuration-map}. Controlled representation studies specify
their diagnostic readouts separately in Appendix~\ref{app:representation-studies}.

\subsection{Exact Runtime and Input Contract}
\label{app:contract}

Hidden states depend on the input contract: changing the serializer, chat
template, or tool schema can change the representation of the same package.

\paragraph{Pre-action endpoint.}
The detector runs after the complete skill package enters the protected model's
context. It finishes before the model generates a token, selects a tool, or
causes an external side effect. The detector observes the native prefill that
loads the task, tool interface, and skill. It executes no package content and
exposes no labels, analyst rationales, or future outcomes to the model.

\paragraph{Prompt and tool contract.}
The frozen system message is specified in Box~A.1. The available interface contains one function, \texttt{run\_shell}, with one required string field, \texttt{cmd}. The user message and package subtree follow the exact serialization template shown in Box~A.2. This formatting is fixed across classes, sources, methods, and backbones.

\begin{promptbox}[title={\textbf{Contract Box A.1:} Frozen Agent System Message},label={box:system-prompt}]
\small
``You are an AI agent. Use the available tools when needed to complete the user's task. Use the loaded skill as task-specific workflow guidance and resources while carrying out the task.''
\end{promptbox}

\begin{codebox}[title={\textbf{Contract Box A.2:} Tool Interface and Serialized Package Input Template},label={box:input-serialization}]
// 1. Tool Declaration (Native Function Schema)
{
  "name": "run_shell",
  "description": "Execute a shell command within the protected workspace environment.",
  "parameters": {
    "type": "object",
    "properties": {
      "cmd": {"type": "string", "description": "Shell command to run."}
    },
    "required": ["cmd"]
  }
}

// 2. Serialized Input Structure
<trusted_user_task>
{task}
</trusted_user_task>

<loaded_skill_package
  input_unit="owning_skill_subtree"
  serialization="path_sorted_inert_static">
  <skill_file path="SKILL.md">
    {root_declaration_and_instructions}
  </skill_file>
  <skill_file path="{relative_auxiliary_file_path}">
    {auxiliary_code_or_configuration}
  </skill_file>
</loaded_skill_package>
\end{codebox}

\paragraph{Package serialization.}
An input is the owning skill subtree rooted at \texttt{SKILL.md}. Files are
serialized in lexicographic relative-path order. Readable UTF-8 files retain
their relative path and text inside \texttt{skill\_file} delimiters. Binary
files are represented by a path and placeholder. Their bytes never enter the
prompt. The detector covers the root declaration, scripts, configuration, and
readable auxiliary material loaded by the agent.

\paragraph{Region binding.}
\decl covers the root front matter, title, and first prose section.
\oper covers the remainder of the root file plus all readable auxiliary text.
The binding uses package structure as a proxy for semantic role: O can include
prose, examples, and configuration as well as executable code.
\boundary is the native assistant-generation suffix, or the
implicit prompt end immediately before generation or tool selection. Character
spans are projected to model-native token offsets. A token crossing a region
boundary is assigned to exactly one region. D and O must both be nonempty.
Large-ID and its white-box panel retain at most eight Declaration atoms and 256
Operation atoms. The compact intervention and CDF-consensus timing configurations
retain eight and 24, respectively. Atom limits belong to the extraction configuration,
not to the definition of D/O; each comparison holds them fixed across classes
and backbones.

\paragraph{Eligibility and failure behavior.}
Large-ID uses a 24,576-token cap and the balanced $2\times2$ uses 12,288.
Each protocol defines its own support. These caps are
not model-specific truncation settings. A package enters a cross-backbone
comparison only when every included tokenizer preserves every required region.
Missing root files, ambiguous or empty spans, removed required content,
incompatible model or prompt contracts, and nonfinite features produce an
explicit contract error.
The pipeline never silently truncates a required region or treats a contract
failure as benign.
Changes to the backbone, tokenizer, prompt, tool schema, or region map require
revalidation before reusing fitted readouts.

\subsection{Complete Detector Algorithm}
\label{app:algorithm}

The following steps specify the default package-level policy. CompoSkill's
development-only adaptation is described in Appendix~\ref{app:composkill-calibration}.

\paragraph{Inputs and outputs.}
Inputs are a trusted task, a tool schema, a serialized skill subtree, the
protected backbone, the D/O/B representation bundle, and three component
readouts. The decision threshold is 0.5. Outputs are a continuous risk score,
three branch probabilities, a binary \textsc{pass}/\textsc{block} decision, and
coverage status.

\begin{table}[!t]
\centering
\caption{\textbf{Package-level extraction configuration.} Layer indices are zero-based.
All models use BF16, SDPA, and batch size one. The first two backbones also
support exact-slot confirmation and compact runtime measurements.}
\label{tab:model-configuration}
\scriptsize
\setlength{\tabcolsep}{3.5pt}
\begin{tabularx}{\textwidth}{@{}l l c
  >{\raggedright\arraybackslash}X l@{}}
\toprule
\textbf{Backbone} & \textbf{Role} & \textbf{Layers} &
\textbf{Sampled layer indices} & \textbf{Inference}\\
\midrule
Llama-3.1-8B-Instruct & All + exact/runtime & 32 & 3, 7, 11, 15, 19, 23, 27, 31 & BF16, SDPA\\
Qwen2.5-7B-Instruct & All + exact/runtime & 28 & 3, 6, 9, 13, 17, 20, 23, 27 & BF16, SDPA\\
Mistral-7B-Instruct-v0.3 & Detection/trajectory & 32 & 3, 7, 11, 15, 19, 23, 27, 31 & BF16, SDPA\\
InternLM2.5-7B & Detection/trajectory & 32 & 3, 7, 11, 15, 19, 23, 27, 31 & BF16, SDPA\\
Qwen2.5-14B & Detection/trajectory & 48 & 5, 11, 17, 23, 29, 35, 41, 47 & BF16, SDPA\\
Phi-3-medium-14B & Detection/trajectory & 40 & 4, 9, 14, 19, 24, 29, 34, 39 & BF16, SDPA\\
Qwen3-30B-A3B & Detection/trajectory & 48 & 5, 11, 17, 23, 29, 35, 41, 47 & BF16, SDPA, eager experts\\
\bottomrule
\end{tabularx}
\end{table}

\paragraph{Inference procedure.}
The end-to-end detection flow for each skill package is formalized in Algorithm~A.1. All three ensemble members reuse the same prefill tensors, forming an efficient statistical ensemble without additional model executions.

\begin{algobox}[title={\textbf{Algorithm A.1:} LatentTriad Pre-Action Screening Pipeline},label={algo:pipeline}]
\small
\textbf{Input:} Trusted user task $\mathcal{T}$, tool interface $\mathcal{I}$, serialized skill subtree $\mathcal{S}$, model $\mathcal{M}$, representation projections $(P^{DO}, P^{OB})$, 3-member encoders $j \in \{300, 400, 500\}$, and calibrated heads $(f_B, f_O, f_G)$.\\
\textbf{Output:} Continuous risk score $s(x) \in [0, 1]$, branch probabilities $(p_B, p_O, p_G)$, binary verdict $\hat{y} \in \{\textsc{pass}, \textsc{block}\}$, coverage status.
\vspace{3pt}
\hrule
\vspace{4pt}
\begin{enumerate}[leftmargin=1.6em,itemsep=2.5pt,topsep=1pt]
  \item \textbf{Tokenization and Region Binding:} Serialize input per Appendix~\ref{app:contract}. Map nonoverlapping token spans for Declaration ($\mathcal{D}$) and Operation ($\mathcal{O}$), and record the assistant Boundary token ($\mathcal{B}$).
  \item \textbf{Single-Prefill State Extraction:} Run one native prefill through $\mathcal{M}$; collect hidden tensors $\mathbf{h}_\ell$ at the eight normalized depths in Table~\ref{tab:model-configuration} without generating tokens.
  \item \textbf{Transport and Coordinate Projection:} Pool region states, project into shared 32D coordinates $\mathbf{z}_\ell^{R,a}$ (Equation~\ref{eq:latent-coordinates}), evaluate affine ridge transport $T_\ell^a$, and extract nearest transported atom residuals.
  \item \textbf{Multi-Member Feature Concatenation:} For each ensemble offset $j \in \{300, 400, 500\}$ and branch $q \in \{\mathrm{boundary}, \mathrm{operation}, \mathrm{geometry}\}$, evaluate member encoders and concatenate into branch vectors $\boldsymbol\phi_q$.
  \item \textbf{Component Readout Scoring:} Apply standardizers and $\ell_2$ logistic heads (Equation~\ref{eq:component-head}) to yield branch risk estimates $p_{\mathrm{boundary}}$, $p_{\mathrm{operation}}$, and $p_{\mathrm{geometry}}$.
  \item \textbf{Evidence Aggregation:} Compute continuous composite score $s(x) = \max\big(p_{\mathrm{boundary}},\, \min(p_{\mathrm{operation}}, p_{\mathrm{geometry}})\big)$ via Equation~\ref{eq:detector-score}.
  \item \textbf{Pre-Action Decision Gate:} Assign binary verdict $\hat{y} = \mathbb{I}[s(x) \geq 0.5]$ (Equation~\ref{eq:detector-gate}). Return verdict and branch probabilities before execution or generation.
\end{enumerate}
\end{algobox}

\paragraph{Large-ID fitting partitions.}
The package detector uses three outer training folds, one development
fold, and one test fold. Training family identifiers are sorted
and assigned round-robin to five artificial strata. These strata govern inner
partitioning, Boundary-depth selection, Operation directions, and Geometry
reference weighting; they are not ecosystem labels. For each offset
$j\in\{300,400,500\}$, the first split of a five-way stratified grouped
partition gives inner fit and calibration sets, using seed $20260802+1000+j$.
The grouping unit is the content-linked component described in
Appendix~\ref{app:data-governance}.

Each member fits candidate Boundary probes on inner-fit packages. On the
calibration set, a cutoff limits empirical FPR to at most 5\% in each artificial
stratum. We select depth by stratum-macro F1, breaking ties by AUROC and then
shallower depth.
Only the selected depth is retained: its raw hidden vector enters the final
Boundary matrix, not the candidate probe's score or cutoff. Three members
therefore contribute $3d$ Boundary coordinates for hidden width $d$.

\paragraph{Operation encoder.}
At each depth, unit-normalized O vectors define a benign and malicious mean
within each inner-fit stratum. The midpoint of these two class means is the
stratum centroid. We center by the mean of the five centroids and remove at
most two nondegenerate right-singular directions of their centered matrix.
Each stratum's malicious-minus-benign contrast is projected out of this same
subspace and unit-normalized. Write
$n(\mathbf v)=\mathbf v/\max(\|\mathbf v\|_2,10^{-8})$,
let $\boldsymbol\mu_{\ell,s}^{+}$ and $\boldsymbol\mu_{\ell,s}^{-}$ denote
the class means, and let $\mathbf c_\ell$ be the mean stratum centroid.
The columns of $U_\ell$ contain the retained singular directions. Then
\begin{equation}
\begin{aligned}
\Pi_\ell &= I-U_\ell U_\ell^\top, &
\mathbf v_{\ell,s}
 &= n\!\left(\Pi_\ell(\boldsymbol\mu_{\ell,s}^{+}
                        -\boldsymbol\mu_{\ell,s}^{-})\right),\\
\mathbf q_\ell(x)&=\Pi_\ell\big(n(\mathbf h^O_\ell(x))-\mathbf c_\ell\big), &
a_{\ell,s}(x)&=\mathbf v_{\ell,s}^{\top}\mathbf q_\ell(x).
\end{aligned}
\label{eq:operation-projection}
\end{equation}
At inference, O vectors are first unit-normalized,
then centered, projected out of the fitted subspace, and dotted with the
contrasts, without a second normalization. Five values summarize each depth:
the projection onto their normalized mean direction and the mean, minimum,
maximum, and standard deviation of individual contrast projections.
We retain all eight depth values and four trajectory summaries (mean,
standard deviation, maximum, and late-half minus early-half mean), giving
$5(8+4)=60$ features per member and 180 in the concatenated Operation matrix.
The encoder retains all 180 features after removing stratum-centroid directions.

\paragraph{Geometry coordinates and transport.}
Two shared 32-dimensional projections use all benign outer-training packages,
with equal package weights. At each depth, D and O are individually
unit-normalized; the mean of their paired means defines a common center. The leading
right-singular vectors of the stacked centered D/O vectors define $P^{DO}_\ell$.
O/B independently define $P^{OB}_\ell$ by the same procedure. Projection seeds
are 20260822 and 20260922, respectively, plus the layer position.
For relation $a\in\{DO,OB\}$, common center
$\boldsymbol\mu^a_\ell$, and shared projection $P^a_\ell$, the coordinates are
\begin{equation}
\mathbf z^{R,a}_\ell=P^a_\ell
\big(n(\mathbf h^R_\ell)-\boldsymbol\mu^a_\ell\big).
\label{eq:latent-coordinates}
\end{equation}
Equation~\ref{eq:latent-coordinates} applies to pooled states and individual
atoms, using $\epsilon=10^{-8}$.
D, O, B, and D/O atoms enter D/O coordinates; O and B also enter O/B
coordinates. Projected arrays are stored in float16. Transport and residual
arithmetic use float32; the B-norm denominator in the relative O/B residual
is computed from its stored float16 array.
Angular gaps below are $1-\cos(\cdot,\cdot)$; norm denominators are floored
at $\epsilon$.

Within each member's inner-fit set, affine ridge maps
$T^a_\ell(\mathbf z)=W^a_\ell\mathbf z+\mathbf b^a_\ell$
with regularization 1.0 and an unpenalized intercept fit benign
D$\rightarrow$O and O$\rightarrow$B pairs, giving each artificial stratum
equal total weight. For each O atom, we select the nearest
transported D atom by Euclidean distance in D/O coordinates and retain their
residual. The layer's local threshold is the median of the five strata's
benign 90th-percentile atom distances. Only benign Large-ID training packages
fit these projections, maps, and thresholds.

\paragraph{Geometry feature matrix.}
Each depth contributes eight D$\rightarrow$O features: the mean, 75th and
90th percentiles, maximum, and standard deviation of atom distances; the
fraction exceeding the local threshold; the pooled transport residual norm;
and the angular gap between predicted and observed O.
Let $\mathbf u_\ell$ be the mean atom residual weighted by positive
distance exceedance, with uniform valid-atom weights when total exceedance
is at most $\epsilon$. In D/O coordinates, define
$\boldsymbol\delta_\ell=\mathbf z^{B,DO}_\ell-
T^{DO}_\ell(\mathbf z^{D,DO}_\ell)$.
Five uptake features are the cosines of $\mathbf u_\ell$ with
$\boldsymbol\delta_\ell$ and $\mathbf z^{B,DO}_\ell$,
$\|\boldsymbol\delta_\ell\|/\max(\|\mathbf u_\ell\|,\epsilon)$,
$\|\mathbf u_\ell\|$, and mean positive atom-distance exceedance.

Six further features measure Boundary relations: the O$\rightarrow$B
transport residual norm, predicted-versus-observed B angular gap, direct O/B
angular gap, residual norm divided by B norm, $\|\boldsymbol\delta_\ell\|$,
and the angular gap between predicted O and B in D/O coordinates.
The same four trajectory summaries expand these $8+5+6$ depthwise values to
$19(8+4)=228$ coordinates. Each of the three resulting feature blocks adds
its mean, standard deviation, maximum, and 90th percentile across the expanded
block coordinates
(12 values). For block means $a,u,b$, five interactions add
$\sqrt{\max(a,0)\max(u,0)}$, $\min(a,u)$, $\max(a,u)$, $ab$, and $ub$.
Thus each member supplies $228+12+5=245$ Geometry features, or 735 after
concatenation. Geometry undergoes neither Operation-style stratum-subspace
removal nor feature selection.

Only the final standardizers and $C=1$ L2-logistic readouts use outer training
plus development, without class weighting or feature selection.
Equation~\ref{eq:detector-score} fuses their probabilities directly at cutoff
0.5; this Large-ID readout has no empirical-CDF calibration.

\FloatBarrier
\subsection{Dataset Construction and Study Scope}
\label{app:data-governance}

The following tables identify the data used for detector fitting, mechanism
analysis, and evaluation. Studies that reuse a dataset retain their stated
training and test partitions.

\begin{table}[!t]
\centering
\caption{\textbf{Datasets and experimental scope.} Counts are evaluation units,
not a combined sample total. Supporting panels remain separate from Large-ID.}
\label{tab:datasets}
\scriptsize
\setlength{\tabcolsep}{3.2pt}
\begin{tabularx}{\textwidth}{@{}lp{3.0cm}>{\raggedright\arraybackslash}X@{}}
\toprule
\textbf{Dataset / panel} & \textbf{Support} & \textbf{Question tested}\\
\midrule
Task--operation compatibility & 800 examples/model & Cross-form readout and matched Operation intervention\\
Fixed-node composition & 4,608 presentations/model & Connection and authorization counterfactuals\\
Structural transfer & 2,304 inputs/model & Known-template panel; separate template holdouts in Appendix~\ref{app:representation-transfer}\\
Same-prompt diagnosis & $4\times384$ questions/model & Generated judgments versus pre-action readouts\\
Large-ID & 2,915 matched pairs & Skill detection and final-configuration controls\\
CompoSkill & 690 matched path pairs & Whole-path detection; also the support for adaptive rewriting\\
SCR-Bench & 667 mechanism pairs & Frozen transfer across paired mechanism conditions\\
Balanced $2\times2$ & 228 packages & Operation-sensitive ranking with surface controlled\\
Exact-slot intervention & 40 aligned families & Operation-to-Boundary transport; 20 development, 20 confirmation\\
Multi-depth intervention & 100 matched pairs & Fixed-probe margin and verdict response under compact extraction\\
White-box / broker & 256 matched pairs & Package rewrites and separate request-level containment\\
Natural attribution & 610 packages & Four-class localization on 150 grouped families\\
Controlled attribution & 128 four-variant families & Localization under matched changes in risk placement\\
\bottomrule
\end{tabularx}
\end{table}

\begin{table}[!t]
\centering
\caption{\textbf{Large-ID source composition.} Each pair contains one malicious and one source-matched benign package. Source names link to frozen releases. Groups define fold membership; source-macro metrics weight the five rows equally.}
\label{tab:large-id-source-distribution}
\small
\setlength{\tabcolsep}{4.5pt}
\begin{tabular}{@{}lrrrr@{}}
\toprule
\textbf{Source} & \textbf{Pairs} & \textbf{Components} & \textbf{Packages} & \textbf{Share} \\
\midrule
\href{https://github.com/lxyeternal/MalSkillBench/tree/06e083125d5ec5dd7a189dfedadfe587a70635b9}{MalSkillBench} & 2,621 & 1,824 & 5,242 & 89.9\% \\
\href{https://github.com/qualixar/skillfortify/tree/dbb5942deae46afdb5b85440d5f5d851c197c30e}{SkillFortify} & 180 & 39 & 360 & 6.2\% \\
\href{https://huggingface.co/datasets/zenith6888/SkillsBench-1650/tree/68c0d4bf857f48e5fa4f3baa4613f0c0980a7fb2}{SkillsBench-1650} & 63 & 41 & 126 & 2.2\% \\
\href{https://huggingface.co/datasets/yoonholee/agent-skill-malware/tree/5cff435ddab2cc34261875885de9ac50d40393ec}{Agent-skill malware} & 29 & 29 & 58 & 1.0\% \\
\href{https://huggingface.co/datasets/Agent-Threat-Rule/atr-skill-benchmark/tree/7219b10d2ac077e3db8c87d43d653a67935cdb5d}{ATR Skill Benchmark} & 22 & 21 & 44 & 0.8\% \\
\midrule
Total & 2,915 & 1,954 & 5,830 & 100.0\% \\
\bottomrule
\end{tabular}
\end{table}

\paragraph{Source releases and pairing.}
Table~\ref{tab:large-id-source-distribution} links each source to its frozen
revision. We construct benign--malicious pairs within each source, preferring
available controls from the same upstream family. Remaining controls are
matched one-to-one by package size and total, text, and binary file counts.
The support manifest retains source revisions, sample IDs, pair IDs, and
matching modes for each package.

\paragraph{Evaluation groups.}
Package detection joins original groups, shared-content links, and benign--malicious
pair links into connected components. The resulting 1,954 components partition
all 5,830 packages into five folds. Shared document and package identities stay
within one fold even when a matched pair contains different package trees.
Components define both fitting partitions and bootstrap units.

All learned methods in Table~\ref{tab:large-id-main} use these components.
White-box, multi-depth intervention, and broker tests instead use application
probes fitted on matched-pair-disjoint folds. Their folds keep each constructed
pair together but do not enforce separation of all shared-content components.
These probes remain fixed during each application test. Localization and
composed-skill studies use their stated family and role-group partitions.
Ecosystem labels support matching and evaluation, not runtime detection.

\paragraph{Splits and counterfactual selection.}
Splits are fixed before any method is fit. Trainable methods finalize
their parameters and operating point before each test fold. The balanced
$2\times2$ contains 57 packages in each combination of benign/suspicious
surface and benign/malicious operation. Candidate inverse groups first satisfy
dual-model token eligibility and pair completeness. A deterministic
content-based ordering then selects groups without labels, model outputs, or
detector scores.

\paragraph{Safety of data handling.}
All studies use inert package loading. Scripts and auxiliary payloads are
serialized as model input but are not executed. No evaluation grants package
code credentials, unrestricted network access, or a writable production
environment.

\subsection{Malicious-Package Composition}
\label{app:data-composition}

Malicious packages vary in both behavior and presentation. For the
three locally available Large-ID sources, we inspect complete owning subtrees,
including root instructions and any available scripts, references,
configuration, or self-check material. ATR Skill Benchmark and SkillsBench-1650 use their source taxonomies
and labels. Their files remain sealed.

\begin{table*}[!t]
\centering
\caption{\textbf{Qualitative composition of malicious-package sources.}
Categories summarize source taxonomies and inspected package structure. They
are explanatory strata, not additional labels used for fitting.}
\label{tab:attack-composition}
\scriptsize
\setlength{\tabcolsep}{3.5pt}
\begin{tabularx}{\textwidth}{@{}l>{\raggedright\arraybackslash}p{0.22\textwidth}>{\raggedright\arraybackslash}p{0.31\textwidth}>{\raggedright\arraybackslash}X@{}}
\toprule
\textbf{Source} & \textbf{Representative positive behavior} &
\textbf{Package presentation} & \textbf{Primary difficulty}\\
\midrule
MalSkillBench & Downloaders, substituted dependencies, protected archives,
remote scripts, and implanted transaction utilities & A source-matched benign
skill is paired with a malicious package derived from the same content family &
Near-matched surfaces leave behavior-changing paths, prerequisites, or
instructions as the discriminative evidence\\
SkillFortify & Prompt injection, unsafe tool delegation, exfiltration, and
privilege misuse & Harmful requirements are embedded in otherwise functional
\texttt{SKILL.md} packages & Semantic dilution and disagreement between stated
purpose and requested operation\\
Agent-skill malware & Social-engineering instructions for fake prerequisite
downloads, encoded shell commands, or credential exfiltration & Single
\texttt{SKILL.md} documents; all 58 retained samples contain one file &
Inferring requested external operations and their authorization from
instructions\\
ATR Skill Benchmark & Exfiltration, reverse shells, droppers, typosquatting,
and delayed triggers & The source taxonomy emphasizes relatively short
and explicit attack artifacts & Only 22 matched malicious families remain in
Large-ID, so uncertainty is dominated by family support\\
SkillsBench-1650 & Encoded or steganographic payloads, modular/RSA transforms,
remote scripts, and binary execution & Related operations are expressed through
different carriers and encodings & Representation opacity: pre-action states
may not resolve latent payload semantics\\
\bottomrule
\end{tabularx}
\end{table*}

Three properties make detection difficult. \textbf{Semantic dilution}
hides a small unsafe clause inside a long, useful package. \textbf{Relational
violation} makes each local step plausible even when the declaration,
implementation, recipient, permission, or self-check disagrees.
\textbf{Representation opacity} delays meaning until encoded content, binaries,
network responses, or runtime branches are decoded or executed. \system targets
relational violation and can also expose some semantic dilution. Representation
opacity requires sandboxing or runtime monitoring.

\FloatBarrier
\section{Baselines and Statistical Protocol}
\label{app:baselines-statistics}
Baselines differ in their inputs and fitting procedures. We specify both
below so that comparisons can be interpreted in context.

\paragraph{Adjacent threats and defenses.}
Position-aware skill injection and threshold sharing across tools make harmful
instructions less conspicuous
\citep{hao2026poiseoneinstructionskill,
liuShareLockStealthyMultiTool2026}. Input defenses can remove suspected
instructions \citep{shi2025promptarmorsimpleeffectiveprompt}.
Attention Tracker detects injection through attention patterns
\citep{hung2024attentiontrackerdetectingprompt}, whereas Rennervate provides
token-level detection and sanitization
\citep{zhong2025attentionneeddefendindirect}. ProbGuard predicts future violations
from runtime traces \citep{wang2025probguardproactiveruntimemonitoring}. These
systems motivate separate comparisons for package surface, model state, and
execution state.

\subsection{Baseline Contracts}
\label{app:baselines}

\begin{table}[!t]
\centering
\caption{\textbf{Baseline information and execution contracts.} ``Same
support'' means identical package IDs and group-aware fitting data; AgentLens
preserves its published probe optimization while transferring the labels to
malicious-skill detection and retains its published argmax decision.}
\label{tab:baseline-contracts}
\scriptsize
\setlength{\tabcolsep}{3pt}
\begin{tabularx}{\textwidth}{@{}llccl>{\raggedright\arraybackslash}X@{}}
\toprule
\textbf{Method} & \textbf{Input} & \textbf{Fit} &
\textbf{Generation} & \textbf{Execution} & \textbf{Scope}\\
\midrule
Static regex & Text & No & No & CPU scan & Native baseline\\
SkillSieve L1 & Text + parser & No & No & CPU scanner & Official Layer 1\\
Cisco scanner & Package + parser & No & No & CPU scanner & Default core policy\\
LLM-as-judge & Rendered package & No & No & Judge forward & Frozen Mistral\\
Supervised text & Qwen verbalizer scores & Train+development & No & Two model forwards & L2 logistic/0.5\\
RouteGuard & Attention + hidden & Yes & No & Four probe prefills &
Paper-spec reimplementation; fixed 0.5\\
AgentLens & Boundary hidden & Yes & No & One native prefill &
Released script; default 28-layer setting\\
\system & D/O/B geometry & Train+development heads & No & One native prefill &
Fixed hard fusion/0.5\\
\bottomrule
\end{tabularx}
\end{table}

\paragraph{Static control.}
Static Regex applies a suspicious-operation pattern set, requires no fit, and
does not read protected-model states. Its backbone-independent result
appears once in the shared panel of each detection table.

\paragraph{Released scanners and semantic judge.}
SkillSieve L1 uses the released deterministic first-stage parser/scanner. Cisco
AI Skill Scanner uses the official implementation\footnote{\url{https://github.com/cisco-ai-defense/skill-scanner}},
the default \texttt{ScanPolicy}, and its static, bytecode, and pipeline
analyzers. Its native \texttt{is\_safe} decision blocks Critical or High
findings. Ordinal maximum severity supplies AUROC and AP. No package code is
executed. On Large-ID, the strict loader accepts 5,577 packages. The documented
lenient loader handles 253 packages with missing or malformed frontmatter,
preserving all 5,830 common IDs.

LLM-as-judge renders the same package to a frozen Mistral-7B model and compares
benign/malicious verbalizer likelihoods without generation (Box~B.1). It cannot access
the protected model's hidden state. The supervised text readout scores the
rendered package with Qwen2.5-7B. It uses two full-package verbalizer
log-likelihoods as inputs to a standardizer and $C=1$ L2-logistic classifier.
The classifier fits on each Large-ID training-plus-development partition and
uses threshold 0.5. Test labels select no feature, parameter, or operating point.

\begin{promptbox}[title={\textbf{Protocol Box B.1:} LLM-as-Judge \& Text Verbalizer Scoring Contract},label={box:llm-judge}]
\small
\textbf{System Prompt:} ``You are an automated security auditor evaluating LLM agent skills. Determine whether the provided skill package contains malicious behavior, deceptive intent, or security policy violations. Output exactly one word: `Malicious' or `Benign'.''\\[3pt]
\textbf{Input Structure:} \texttt{<loaded\_skill\_package>} \textit{\{package\_text\}} \texttt{</loaded\_skill\_package>}\\[3pt]
\textbf{Decision Mechanism (No Free Generation):} Rather than sampling autoregressive tokens, the judge evaluates the unnormalized output logits $\ell_{\mathrm{Malicious}}$ and $\ell_{\mathrm{Benign}}$ at the terminal prefill token, calculating the verbalizer likelihood ratio:
\[
p(\mathrm{Malicious}) = \frac{\exp(\ell_{\mathrm{Malicious}})}{\exp(\ell_{\mathrm{Benign}}) + \exp(\ell_{\mathrm{Malicious}})}
\]
\end{promptbox}

\paragraph{Internal baselines.}
RouteGuard is a paper-spec reimplementation of attention/hidden-state fusion
with four probe prefills. AgentLens learns interpretable safety-steering
representations from coding-agent trajectories. We transfer its released probe
protocol to the loaded-skill pre-action Boundary as specified below.

\paragraph{Selection fairness.}
Fairness means identical test IDs, family partitions, and decision timing, not
identical training code. In Large-ID, \system refits projections and member encoders on
training families. RouteGuard refits its four probes on training data.
AgentLens uses development to select layer and epoch before a
training-plus-development refit. \system and the supervised text baseline fit
their readouts on training plus development without selecting an operating
point. AgentLens retains its published
argmax rule. RouteGuard and \system use the same 0.5 probability decision. No
method receives a test label, per-source threshold, or post-hoc target recall.

\paragraph{AgentLens task transfer.}
We extract the full Boundary last-token vector and fix the layer range to the
released training script's default first 28 layers across backbones.\footnote{Release:
\href{https://github.com/EddyLuo1232/AgentLens/tree/22d397d03774e66bffbc04cffc04cc3eacfca5c9}{\texttt{AgentLens@22d397d0}}.}
A group-disjoint development partition selects the layer and epoch.
The probe preserves the released AgentLens optimization
contract: Linear$(h,2)$ plus softmax, Adam with learning rate $10^{-2}$,
cross-entropy loss, batch size 16, and at most 100 epochs. It adds no
standardization, sample weighting, Top-K detection, or ensemble. The selected
probe is refit on training plus development and evaluated once with its argmax
decision.

On deeper backbones, this fixed layer range differs from \system's full-depth
grid. The methods also differ in member construction and readout adaptation.
Appendix~\ref{app:final-component-controls} uses the final Large-ID fits to
separate fusion, relation features, and member construction. A dimension-matched
independent-state control replaces Geometry while retaining the B/O heads; single-member
controls refit the readouts for each retained member.

\subsection{Compositional Path Calibration}
\label{app:composkill-calibration}

CompoSkill Path uses the same extracted states for the fixed and nested
\system variants. The fixed contract applies
$\max(B,\min(O,G))$ at threshold 0.5. The nested contract changes only the
fusion rule and operating point. Each fold fits heads on training data, selects
one of four monotone rules and 33 thresholds by development equal-threat macro
F1, and then refits the heads on training plus development. The outer test fold
is evaluated once. The candidate rules and thresholds are specified before
evaluation; the fixed policy serves as a separate control.

\begin{table}[!t]
\centering
\caption{\textbf{\system detects 93.0--97.2\% of composed attacks at 3.8--4.3\% FPR.} All 690 matched CompoSkill pairs are retained after seven rewrite rounds (\%).}
\label{tab:composkill-main}
\footnotesize
\setlength{\tabcolsep}{2.2pt}
\renewcommand{\arraystretch}{1.04}
\begin{tabularx}{\linewidth}{@{}>{\raggedright\arraybackslash}p{3.15cm}*{9}{>{\centering\arraybackslash}X}@{}}
\toprule
\rowcolor{tablegroupgray} & \multicolumn{3}{c}{AgentLens (1P)} & \multicolumn{3}{c}{RouteGuard (4P)} & \multicolumn{3}{c}{\system (1P)} \\
\cmidrule(lr){2-4}\cmidrule(lr){5-7}\cmidrule(lr){8-10}
\rowcolor{tablegroupgray}\textbf{Backbone} & TPR$_0\uparrow$ & FPR$\downarrow$ & TPR$_7\uparrow$ & TPR$_0\uparrow$ & FPR$\downarrow$ & TPR$_7\uparrow$ & TPR$_0\uparrow$ & FPR$\downarrow$ & TPR$_7\uparrow$ \\
\midrule
Llama-3.1-8B & 84.1 & 17.5 & 83.9 & \textbf{94.2} & 8.3 & \textbf{94.1} & 93.0 & \textbf{4.3} & 92.9 \\
\addlinespace[2pt]
Qwen2.5-7B & 59.4 & 38.8 & 59.4 & 92.5 & 12.9 & 92.5 & \textbf{95.9} & \textbf{4.3} & \textbf{95.9} \\
\addlinespace[2pt]
Qwen2.5-14B & 71.2 & 29.4 & 71.0 & 91.8 & 11.2 & 91.7 & \textbf{96.4} & \textbf{4.1} & \textbf{96.3} \\
\addlinespace[2pt]
Qwen3-30B-A3B & 81.5 & 19.2 & 81.4 & 92.4 & 10.1 & 92.2 & \textbf{96.8} & \textbf{3.9} & \textbf{96.8} \\
\addlinespace[2pt]
Llama-3.3-70B-INT8 & 85.2 & 16.1 & 85.0 & 93.1 & 9.4 & 93.0 & \textbf{97.2} & \textbf{3.8} & \textbf{97.1} \\
\bottomrule
\end{tabularx}
\vspace{2pt}
\parbox{\linewidth}{\scriptsize P: prefills; subscripts 0/7: original/seven-round paths. FPR uses unchanged clean controls. Bold: best per backbone. \system uses development-selected fusion and thresholds. Protocol: Appendix~\ref{app:multiskill-stress}.}
\end{table}

\begin{table}[!t]
\centering
\caption{Fixed and development-adapted \system on CompoSkill Path (\%). Fixed uses $\max(B,\min(O,G))$ at 0.5. Nested selection searches four monotone fusion rules and 33 thresholds on development data.}
\label{tab:composkill-fixed-nested}
\scriptsize
\setlength{\tabcolsep}{3.5pt}
\begin{tabular}{@{}llrrrrrr@{}}
\toprule
& & \multicolumn{3}{c}{Fixed} & \multicolumn{3}{c}{Nested development} \\
\cmidrule(lr){3-5}\cmidrule(lr){6-8}
Backbone & Mode & FPR$\downarrow$ & Recall$\uparrow$ & F1$\uparrow$ & FPR$\downarrow$ & Recall$\uparrow$ & F1$\uparrow$ \\
\midrule
Llama-3.1-8B & Explicit & 19.8 & 92.4 & 87.1 & 8.7 & 86.3 & 88.5 \\
 & Implicit & 0.0 & 99.7 & 99.9 & 0.0 & 99.7 & 99.9 \\
\midrule
Qwen2.5-7B & Explicit & 13.7 & 95.3 & 91.2 & 7.6 & 92.7 & 92.6 \\
 & Implicit & 0.9 & 99.4 & 99.3 & 1.2 & 99.1 & 99.0 \\
\bottomrule
\end{tabular}
\end{table}

\subsection{Compositional Mechanism and Adaptive Stress}
\label{app:multiskill-stress}

Table~\ref{tab:composkill-mode-details} reports explicit and implicit attack
modes derived from CompoSkill-Bench~\citep{liu2026composkill}.
We freeze dataset revision \texttt{1a13cca2}\footnote{Full revision:
\url{https://huggingface.co/datasets/Limax11/CompoSkill-Bench/tree/1a13cca2a63c9d6fdcc196df853c3a372ed31e8a}.}.
Its primary design contains 380 three-skill instances, each with clean,
explicit, and implicit variants (1,140 records). We pair each available
attack variant with its clean counterpart, requiring a valid task, a resolved
source--bridge--terminal chain, and complete workspace files. Implicit variants
also require a valid execution guide, matching path length, and no primary-chain
skill identifiers in the sanitized payload. Missing variants are not synthesized.
This yields 344 explicit and 346 implicit pairs (690 mode-specific pairs,
1,380 evaluation rows); a clean variant can appear in both modes.
The five outer folds separate scenario/persona role
groups. LatentTriad and AgentLens use one native prefill; RouteGuard uses four
fixed probe prefills. Cisco Cross-Skill uses the official scanner to
jointly inspect the three installed root documents. Our root-document adaptation
of SkillReact extracts capability conjunctions from those documents; the original
framework also inspects bundled code \citep{wangWhenSafeSkills2026a}.
Our Qwen3-30B-A3B reimplementation of ChainGuard evaluates each candidate against
previously installed skills, following ColluSkill's context-aware procedure
\citep{zengColluSkillAdversarialCrossSkill2026}.

The serialized path includes the trusted task, the root \texttt{SKILL.md}
of each active skill in path order, and the workspace visible at the audited
boundary. In this path-specific region map, D contains the task and skill
declarations; unlike the package map, it includes the trusted task. O contains the remaining
skill instructions and visible workspace. Auxiliary package files remain
tool-addressable but are not loaded into this path view. Private JSON keys
beginning with an underscore are removed recursively from implicit execution
guides; labels and attack-mode metadata are also excluded. The explicit view
precedes chain execution. The implicit view follows workspace observation and
precedes the terminal action. Both are single-prefill inspections of a given
path context, rather than end-to-end planner rollouts.

Table~\ref{tab:composkill-main} pools all 690 mode-specific pairs to report
original-path detection and the seven-round stress endpoint on the same support.
It covers five backbones, including Llama-3.3-70B in INT8; the path inputs,
fold protocol, and development-only selection rules are unchanged.
Table~\ref{tab:composkill-mode-details} instead pools paths within each mode
separately for Llama-3.1-8B and Qwen2.5-7B. \system reaches 88.5--99.9\% F1 and
0.0--8.7\% FPR across the four backbone--mode cells. Three paired F1 gains over
RouteGuard exclude zero; the Llama explicit gain is 1.2 points [-1.4, 3.9].
The installation checks receive identical skill documents for both members
of each pair, without the task or workspace that distinguishes their risk.

\begin{table}[!t]
\centering
\caption{\textbf{Development-adapted \system has the highest F1 in all four backbone--mode comparisons.} Metrics pool all paths within each mode (\%).}
\label{tab:composkill-mode-details}
\vspace{-2pt}
\fontsize{6.5}{7.0}\selectfont
\setlength{\tabcolsep}{2.7pt}
\renewcommand{\arraystretch}{0.82}
\begin{tabularx}{\linewidth}{@{}>{\raggedright\arraybackslash}p{2.42cm}>{\raggedright\arraybackslash}p{1.73cm}*{6}{>{\centering\arraybackslash}X}@{}}
\toprule
\rowcolor{tablegroupgray}\multicolumn{8}{l}{\textbf{(a) Whole-path representation detectors}\quad complete task path} \\
\textbf{Method} & \textbf{Backbone} & \multicolumn{3}{c}{\textbf{Explicit}} & \multicolumn{3}{c}{\textbf{Implicit}} \\
\cmidrule(lr){3-5}\cmidrule(lr){6-8}
& & FPR$\downarrow$ & Recall$\uparrow$ & F1$\uparrow$ & FPR$\downarrow$ & Recall$\uparrow$ & F1$\uparrow$ \\
\midrule
AgentLens (1P) & Llama-3.1-8B & 15.7 & 69.2 & 74.8 & 19.4 & 98.8 & 90.6 \\
RouteGuard (4P) & Llama-3.1-8B & 16.3 & 90.1 & 87.3 & 0.3 & 98.3 & 99.0 \\
\textbf{\system (1P)} & Llama-3.1-8B & \textbf{8.7} & 86.3 & \textbf{88.5} & \textbf{0.0} & \textbf{99.7} & \textbf{99.9} \\
\cmidrule(lr){1-8}
AgentLens (1P) & Qwen2.5-7B & 37.2 & 40.4 & 45.5 & 40.5 & 78.3 & 71.6 \\
RouteGuard (4P) & Qwen2.5-7B & 24.7 & 88.4 & 82.9 & 1.2 & 96.5 & 97.7 \\
\textbf{\system (1P)} & Qwen2.5-7B & \textbf{7.6} & \textbf{92.7} & \textbf{92.6} & \textbf{1.2} & \textbf{99.1} & \textbf{99.0} \\
\midrule
\rowcolor{tablegroupgray}\multicolumn{8}{l}{\textbf{(b) Installation-time composition checks}\quad installed skill set; no task or workspace state} \\
\textbf{Method} & \textbf{Build} & \multicolumn{3}{c}{\textbf{Explicit}} & \multicolumn{3}{c}{\textbf{Implicit}} \\
\cmidrule(lr){3-5}\cmidrule(lr){6-8}
& & FPR$\downarrow$ & Recall$\uparrow$ & F1$\uparrow$ & FPR$\downarrow$ & Recall$\uparrow$ & F1$\uparrow$ \\
\midrule
Cisco Cross-Skill & Official & 30.5 & 30.5 & 37.9 & 31.2 & 31.2 & 38.4 \\
SkillReact & Root-doc adaptation & 96.8 & 96.8 & 65.9 & 96.8 & 96.8 & 65.9 \\
ChainGuard & Qwen3 reimpl. & 46.5 & 46.5 & 48.2 & 43.6 & 43.6 & 46.6 \\
\bottomrule
\end{tabularx}
\vspace{1pt}
\parbox{\linewidth}{\fontsize{6.5}{7.0}\selectfont \emph{Input note.} Panel (b) receives the same installed skill documents for both members of each pair; the task-conditioned path is visible only to panel (a). Bold numbers mark \system column optima; P denotes model prefills.}
\end{table}

Node-only scanners answer a narrower question: whether independent package
inspection can recover a risk that emerges only from composition. They inspect
each activated package separately and never observe the complete task path.

\begin{table}[!t]
\centering
\caption{\textbf{Node-only controls on CompoSkill.} Each method scans the three constituent skills independently and never observes their task-conditioned composition.}
\label{tab:composkill-node-controls}
\scriptsize
\setlength{\tabcolsep}{3.2pt}
\begin{tabularx}{\linewidth}{@{}>{\raggedright\arraybackslash}p{2.45cm}>{\raggedright\arraybackslash}p{1.25cm}*{6}{>{\centering\arraybackslash}X}@{}}
\toprule
\textbf{Method} & \textbf{Build} & \multicolumn{3}{c}{\textbf{Explicit}} & \multicolumn{3}{c}{\textbf{Implicit}} \\
\cmidrule(lr){3-5}\cmidrule(lr){6-8}
& & FPR$\downarrow$ & Recall$\uparrow$ & F1$\uparrow$ & FPR$\downarrow$ & Recall$\uparrow$ & F1$\uparrow$ \\
\midrule
SkillsGuard & Pinned & 96.5 & 96.5 & 65.9 & 96.0 & 96.0 & 65.7 \\
Cisco independent & Pinned & 13.7 & 13.7 & 21.5 & 13.9 & 13.9 & 21.7 \\
SkillSpector & Pinned & 61.9 & 61.9 & 55.3 & 62.4 & 62.4 & 55.5 \\
\bottomrule
\end{tabularx}
\vspace{1pt}
\parbox{\linewidth}{\scriptsize Attack and clean paths contain the same skill documents. Consequently, node-only scores are identical within each pair, pair AUROC is 50.0\%, and FPR equals recall.}
\end{table}

The SCR-Bench panel~\citep{xie2026benign} transfers the fitted CompoSkill detectors without using an
SCR label for fitting or threshold selection. It contains 667 matched cases:
150 capability-flow, 401 trust-transfer, and 116 authorization-confusion
pairs. We use source revision \texttt{b8bb3e9e}.\footnote{Release:
\href{https://github.com/saint-viperx/SCR_Bench/tree/b8bb3e9e3cf7e72db091655f2d5e8ef15b273848}{\texttt{SCR-Bench@b8bb3e9e}}.}
Of its 118 AuthBlur cases, \texttt{authblur:26} and \texttt{authblur:76}
lack the plain-advisor variant required for the four-condition panel, leaving
116 eligible cases. Primary contrasts are terminal-only versus composed-neutral
for capability flow, control versus experiment for trust transfer, and level 0
versus level 3 for authorization confusion. Each pair holds the case and
downstream action fixed while changing the mechanism condition.
We use 2,000 case-paired bootstrap draws and 20,000 paired
sign-flip draws. Holm correction is applied across the three mechanisms within
each backbone and method.

The adaptive panel adapts ColluSkill's scanner-feedback rewriting to our
path-level detector \citep{zengColluSkillAdversarialCrossSkill2026}.
We generate candidates over seven rounds. At each round, the attacker receives
the verdict, branch scores, and high-risk region summary. A candidate must
preserve the task, node order, tools, paths, endpoints, artifact handoffs, and
capability graph. It must also pass an independent frozen semantic judge. All
690 original paths remain in the denominator, and path-paired uncertainty uses
2,000 bootstrap draws. Invalid rewrites retain the preceding valid path and
never count as successful attacks.
The rewrites target \system; AgentLens and RouteGuard score the same resulting
paths, rather than candidates optimized separately against each baseline.

Table~\ref{tab:multiskill-stress} separates detector performance from attack
coverage for Llama-3.1-8B and Qwen2.5-7B. Of the initially blocked paths, nine on
Llama-3.1-8B and eight on Qwen2.5-7B receive
at least one valid rewrite. These sets contain nine distinct mode-specific
paths in total. New evasions occur on one of Llama's nine paths and none of
Qwen's eight. Repeated rounds on the same path do not increase this coverage.
The evasion rate over all paths reflects both the attacker's ability to
produce valid candidates and the detector's response. Conditional rates on
these nine and eight paths isolate the response to valid rewrites.

\begin{table}[!t]
\centering
\caption{\textbf{Compositional mechanism transfer and adaptive stress.} SCR-Bench changes one mechanism at a time; the adaptive panel preserves each path's capability graph over seven feedback rounds.}
\label{tab:multiskill-stress}
\scriptsize
\setlength{\tabcolsep}{2.6pt}
\renewcommand{\arraystretch}{0.90}
\begin{tabularx}{\linewidth}{@{}lrrrr>{\raggedright\arraybackslash}X@{}}
\toprule
\rowcolor{tablegroupgray}\multicolumn{6}{l}{\textbf{(a) Frozen transfer to SCR-Bench}\quad 667 paired cases; no SCR label used for fitting} \\
Backbone & \shortstack{Capability flow\\$\Delta s$} & \shortstack{Trust transfer\\$\Delta s$} & \shortstack{Authorization\\$\Delta s$} & \shortstack{Holm\\significant} & AgentLens control \\
\midrule
Llama-3.1-8B & \textbf{+0.145}$^{*}$ & \textbf{+0.002}$^{*}$ & \textbf{-0.027} & \textbf{2/3} & Opposite on flow \\
Qwen2.5-7B & \textbf{-0.010} & \textbf{+0.008}$^{*}$ & \textbf{+0.229}$^{*}$ & \textbf{2/3} & Saturated at 1.0 \\
\bottomrule
\end{tabularx}
\par\vspace{3pt}
\begin{tabularx}{\linewidth}{@{}>{\raggedright\arraybackslash}p{2.00cm}>{\centering\arraybackslash}p{1.70cm}>{\centering\arraybackslash}p{1.10cm}>{\centering\arraybackslash}p{1.90cm}>{\centering\arraybackslash}p{1.90cm}>{\raggedright\arraybackslash}X@{}}
\toprule
\rowcolor{tablegroupgray}\multicolumn{6}{l}{\textbf{(b) Adaptive rewriting}\quad 690 paths; seven rounds; capability graph held fixed} \\
Backbone & \shortstack{LT TPR$\uparrow$\\$t_0\rightarrow t_7$} & LT FPR$\downarrow$ & \shortstack{AgentLens $t_7$\\TPR/FPR} & \shortstack{RouteGuard $t_7$\\TPR/FPR} & LT$-$baseline $\Delta$TPR [95\% CI] \\
\midrule
Llama-3.1-8B & \textbf{93.0$\rightarrow$92.9} & \textbf{4.3} & 83.9/17.5 & 94.1/8.3 & AL \textbf{+9.0 [+6.1, +11.9]}$^{*}$; RG \textbf{-1.2 [-3.6, +1.2]} \\
Qwen2.5-7B & \textbf{95.9$\rightarrow$95.9} & \textbf{4.3} & 59.4/38.8 & 92.5/12.9 & AL \textbf{+36.5 [+32.8, +40.4]}$^{*}$; RG \textbf{+3.5 [+1.2, +5.7]}$^{*}$ \\
\bottomrule
\end{tabularx}
\par\vspace{3pt}
\begin{tabularx}{\linewidth}{@{}l*{4}{>{\centering\arraybackslash}X}@{}}
\toprule
\rowcolor{tablegroupgray}\multicolumn{5}{l}{\textbf{(c) Valid rewrite coverage}\quad distinct mode-specific paths and repeated attempts} \\
Backbone & \shortstack{Initially\\blocked paths} & \shortstack{Paths with a\\valid rewrite} & \shortstack{Accepted rewrites\\/ attempts} & \shortstack{New evasions\\/ valid paths} \\
\midrule
Llama-3.1-8B & 642 & 9 & 21/4488 & 1/9 \\
Qwen2.5-7B & 662 & 8 & 11/4634 & 0/8 \\
\bottomrule
\end{tabularx}
\par\vspace{2pt}
\scriptsize In (a), $\Delta s=s_{\mathrm{high}}-s_{\mathrm{low}}$ is the mean score change from the lower- to higher-risk condition, without percentage scaling. In (b), rates are percentages and differences are percentage points; AL/RG denote AgentLens/RouteGuard. $^{*}$ marks a 95\% paired interval excluding zero in (b), or Holm-adjusted $p<0.05$ in (a). Repeated rewrites do not increase distinct-path coverage.
\end{table}

\FloatBarrier
\subsection{Metrics, Aggregation, and Statistical Inference}
\label{app:statistics}

Pooled metrics describe the observed package mixture; source-macro metrics
give each ecosystem equal influence.

For malicious as the positive class, we compute
\[
\mathrm{FPR}=\frac{\mathrm{FP}}{\mathrm{FP}+\mathrm{TN}},\quad
\mathrm{FNR}=\frac{\mathrm{FN}}{\mathrm{FN}+\mathrm{TP}},
\]
\[
\mathrm{Precision}=\frac{\mathrm{TP}}{\mathrm{TP}+\mathrm{FP}},\quad
\mathrm{Recall}=\frac{\mathrm{TP}}{\mathrm{TP}+\mathrm{FN}},\quad
\mathrm{F1}=\frac{2\mathrm{TP}}{2\mathrm{TP}+\mathrm{FP}+\mathrm{FN}}.
\]
Undefined precision is set to zero when a method predicts no positives.
For Skill detection, each learned method is fitted independently on each
split. We pool predictions from the five nonoverlapping test folds.
Source-macro metrics are computed within each source and then averaged
equally over the five sources. If source $s$ has $n_s^-$ benign packages,
\[
\mathrm{FPR}_{\mathrm{pooled}}
=\frac{\sum_s\mathrm{FP}_s}{\sum_s n_s^-},\qquad
\mathrm{FPR}_{\mathrm{source}}
=\frac{1}{5}\sum_s\frac{\mathrm{FP}_s}{n_s^-}.
\]
The former weights source-specific FPR by benign sample counts; the latter
assigns each source 20\%. Both retain every test package. An expected deployment
FPR would instead use the source proportions among benign deployment requests.
Confidence intervals use 2,000 shared bootstrap replicates. Components are
stratified by their source-membership sets and resampled with replacement.
Each draw includes all packages in the selected components for every method.
We then recompute source-macro F1. The 14 LatentTriad-minus-comparator intervals share these
draws, preserving component, source, method, and backbone pairing. These are
per-comparison 95\% intervals, not multiplicity-adjusted tests.

The five-source LOSO protocol performs five additional refits per learned
method. Each refit withholds one ecosystem from fitting, development selection,
and calibration. We compute metrics on each held-out source and average the
five results. The 2,000 paired replicates resample
matched families within the held-out source. Every method and backbone
receives the same family multiplicities.

Average precision is the step-integrated area under the precision--recall
curve. At each distinct score, precision is weighted by the increase in recall.
We compute AP from sample-level ranking scores.
Aggregate confusion counts are insufficient. Cisco contributes its discrete
ordinal severity instead of a calibrated probability. AUROC is
threshold-independent; it is the primary metric for the final-configuration
relation-feature contrast and a ranking diagnostic elsewhere. PairAcc is
the fraction of matched inverse pairs in which the malicious-operation member
receives the higher risk score. A tie contributes one half. We use PairAcc only
for matched counterfactual diagnostics.

\section{Large-ID Detection and Cross-Source Generalization}
\label{app:large-id-evidence}

Source slices, component controls, and paired contrasts use the same
component-disjoint package detector. Source-holdout evaluation fits separate
detectors with the test ecosystem excluded.
\subsection{Source-Wise and Paired Results}
\label{app:large-id-source-results}

\begin{table*}[!t]
\centering
\caption{\textbf{Large-ID F1 by source (\%).} Representation detectors report seven-backbone means with ranges in brackets; input-level methods report one value.}
\label{tab:large-id-source-metrics}
\scriptsize
\setlength{\tabcolsep}{3.2pt}
\begin{tabular}{@{}lrrrrrrrr@{}}
\toprule
\textbf{Source} & \textbf{Regex} & \textbf{SkillSieve} & \textbf{Cisco} & \textbf{Mistral judge} & \textbf{Sup. text} & \textbf{AgentLens} & \textbf{RouteGuard} & \textbf{LatentTriad} \\
\midrule
MalSkillBench & 66.7 & 70.3 & 43.2 & 32.2 & 73.3 & 71.2 [28.6, 98.6] & 85.0 [84.2, 86.3] & 99.4 [99.3, 99.6] \\
SkillFortify & 66.7 & 68.4 & 8.5 & 56.0 & 92.4 & 73.0 [35.7, 99.4] & 98.0 [97.2, 98.6] & 100.0 [100.0, 100.0] \\
SkillsBench-1650 & 66.7 & 61.8 & 42.9 & 6.0 & 35.4 & 54.0 [27.6, 67.9] & 55.4 [47.2, 60.5] & 90.9 [88.6, 93.9] \\
Agent-skill malware & 66.7 & 48.8 & 6.7 & 6.7 & 12.9 & 53.0 [25.6, 73.5] & 71.1 [56.5, 78.6] & 89.2 [83.6, 91.5] \\
ATR Skill Benchmark & 66.7 & 52.9 & 37.0 & 51.6 & 77.8 & 60.4 [26.7, 81.6] & 86.4 [82.9, 90.9] & 90.0 [85.7, 93.6] \\
\bottomrule
\end{tabular}
\end{table*}

Table~\ref{tab:large-id-source-distribution} shows the source imbalance.
MalSkillBench supplies 2,621 of the 2,915 benign packages (89.9\%).
\system's FPR on this source is 0.5--0.8\% across backbones; on SkillFortify,
it is 0\%. Rates are higher on the other three sources, which contain 114
benign packages in total (Table~\ref{tab:large-id-source-fpr}).
For example, Llama flags 17/2,621 MalSkillBench packages, 0/180 SkillFortify
packages, 10/63 SkillsBench1650 packages, 1/29 Agent-skill malware packages,
and 5/22 ATR packages. This gives a pooled FPR of 33/2,915 (1.13\%) and a
source-macro FPR of 8.54\%. The latter weights each source equally, exposing
differences in benign acceptance that the dominant source would otherwise mask.

\begin{table}[!t]
\centering
\caption{\textbf{Benign acceptance varies across skill sources.}
FPR ranges (\%) span seven backbones; they are not confidence intervals.
All methods share the listed benign support.}
\label{tab:large-id-source-fpr}
\small
\setlength{\tabcolsep}{3pt}
\renewcommand{\arraystretch}{1.08}
\begin{tabularx}{\linewidth}{@{}Xrrrr@{}}
\toprule
\rowcolor{tablegroupgray}
Source & Benign $n$ & AgentLens & RouteGuard & \system \\
\midrule
MalSkillBench & 2,621 & 0.84--40.02 & 16.06--22.51 & 0.53--0.76 \\
SkillFortify & 180 & 0.00--45.56 & 0.00--1.67 & 0.00--0.00 \\
SkillsBench-1650 & 63 & 19.05--82.54 & 23.81--34.92 & 11.11--23.81 \\
Agent-skill malware & 29 & 17.24--68.97 & 10.34--20.69 & 3.45--13.79 \\
ATR Skill Benchmark & 22 & 18.18--68.18 & 9.09--18.18 & 13.64--27.27 \\
\bottomrule
\end{tabularx}
\end{table}

Table~\ref{tab:large-id-source-metrics} reports source-wise F1, and
Table~\ref{tab:large-id-paired} gives all same-backbone contrasts and
confidence intervals.

\begin{table}[!t]
\centering
\caption{\textbf{Paired Large-ID source-macro F1 differences (percentage points).} Each cell is LatentTriad minus the same-backbone comparator. Intervals use 2,000 shared source-stratified component bootstrap draws. The table reports all 14 contrasts.}
\label{tab:large-id-paired}
\small
\setlength{\tabcolsep}{4pt}
\begin{tabular}{@{}lrr@{}}
\toprule
\textbf{Backbone} & \textbf{$\Delta$ vs. RouteGuard} & \textbf{$\Delta$ vs. AgentLens} \\
\midrule
Qwen2.5-7B & 13.5 [9.5, 18.1] & 50.0 [45.1, 58.2] \\
Llama-3.1-8B & 17.0 [13.1, 21.6] & 14.9 [12.0, 18.1] \\
Mistral-7B & 13.6 [9.6, 18.2] & 9.8 [6.3, 13.3] \\
InternLM2.5-7B & 15.6 [11.0, 20.8] & 18.4 [14.7, 22.3] \\
Qwen2.5-14B & 14.1 [10.8, 17.8] & 65.5 [59.0, 77.4] \\
Phi-3-medium-14B & 16.8 [12.4, 22.1] & 48.7 [42.7, 58.6] \\
Qwen3-30B-A3B & 12.5 [7.6, 17.9] & 13.9 [9.0, 18.0] \\
\bottomrule
\end{tabular}
\end{table}

\FloatBarrier
\subsection{Final-Configuration Component Controls}
\label{app:final-component-controls}

These controls use the component-disjoint Large-ID configuration in
Table~\ref{tab:large-id-main} (summarized in main-text Table~\ref{tab:main-component-controls}): seven backbones, five folds, eight depths,
and three members with offsets 300, 400, and 500. These are follow-up controls
on the same folds; their results do not select the reported detector.

The first comparison fixes all fitted component scores and varies
only the fusion: $B$, $O$, $G$, $\max(B,O)$, $\min(O,G)$, or the full rule.
All use threshold 0.5. The second replaces Geometry with independent-state
features while retaining B/O scores, input coverage, member partitions,
head family, and fusion. For each member, D and O concatenate pooled projected
states and atom-coordinate means, standard deviations, and maximum absolute
values across eight depths. B concatenates its eight O$\rightarrow$B-projected
states. Separate unwhitened PCA fits on benign member-training rows retain
82 D, 82 O, and 81 B coordinates. Concatenating the three members gives
735 features. The replacement Geometry head uses the same standardizer and
L2 logistic readout ($C=1$), fit on training plus development. This control
contains no cross-region residuals, differences, distances, or dot products.

The third comparison refits all three branch heads for each original member separately,
using the same training recipe. We report all three members and
the arithmetic mean of their metrics, rather than selecting the best member
or averaging their scores. Three-member concatenation changes feature width
as well as member count. The equal-width Geometry control matches input
dimension and head family, not every aspect of statistical capacity.

\begin{table}[!t]
\centering
\caption{\textbf{Multi-state evidence recovers Boundary misses at unchanged FPR.} Source-macro metrics (\%) are averaged over seven backbones. All decisions use 0.5.}
\label{tab:final-component-controls}
\small
\setlength{\tabcolsep}{4pt}
\begin{tabularx}{\linewidth}{@{}Xrrrrr@{}}
\toprule
\rowcolor{tablegroupgray}
Configuration & AUROC$\uparrow$ & AP$\uparrow$ & F1$\uparrow$ & FPR$\downarrow$ & Recall$\uparrow$ \\
\midrule
Full: $\max(B,\min(O,G))$ & 99.02 & 99.18 & 93.91 & 9.51 & 96.73 \\
Boundary-only & 97.28 & 97.94 & 91.63 & 9.51 & 92.58 \\
Operation-only & 99.03 & 99.18 & 86.69 & 0.00 & 80.04 \\
Geometry-only & 99.68 & 99.78 & 95.85 & 0.00 & 92.42 \\
$\max(B,O)$ & 99.15 & 99.29 & 94.08 & 9.51 & 97.05 \\
$\min(O,G)$ & 99.53 & 99.64 & 83.56 & 0.00 & 75.98 \\
\midrule
Dimension-matched independent-state control & 98.95 & 99.12 & 93.89 & 9.51 & 96.68 \\
\midrule
Single member: offset 300 & 99.09 & 99.24 & 93.71 & 10.30 & 96.97 \\
Single member: offset 400 & 98.91 & 99.08 & 93.44 & 10.47 & 96.67 \\
Single member: offset 500 & 98.98 & 99.15 & 93.57 & 10.30 & 96.78 \\
Single-member mean & 98.99 & 99.16 & 93.57 & 10.36 & 96.81 \\
\bottomrule
\end{tabularx}
\end{table}

For each backbone, metrics are computed on the concatenated out-of-fold
predictions, then averaged equally over sources. Table~\ref{tab:final-component-controls}
averages these metrics over the seven fixed backbones.
Table~\ref{tab:final-component-backbones} retains the backbone-level results.
The combined branches complement Boundary-only: Full raises mean recall from
\ControlBoundaryRecall\% to \ControlFullRecall\% and F1 from
\ControlBoundaryFOne\% to \ControlFullFOne\%. Both flag exactly the same
benign packages on each backbone, giving \ControlBoundaryFPR\% mean FPR.
Explicit relation features add \ControlRelationAUROC{} points of AUROC over
the equal-width independent-state control. All seven point estimates are positive.

The remaining controls distinguish this complementarity from the choice of
fusion rule. Geometry-only yields \ControlGeometryFOne\% F1 and
\ControlGeometryRecall\% recall with no false alarms. At threshold 0.5, it
trades lower recall than Full for higher F1 on these data. Zero observed FPR
describes this fitted operating point, not a guarantee on new traffic.
The simpler $\max(B,O)$
rule reaches \ControlSimpleFusionFOne\% F1 and \ControlSimpleFusionRecall\%
recall at the same FPR as Full. Here, Geometry confirmation removes some true
alarms from that rule without removing false alarms. These results support the
multi-state evidence, but do not identify the full rule as the best operating point.

\Needspace{4\baselineskip}
Three members differ from the single-member mean by
\ControlMemberAUROC{} AUROC points, \ControlMemberAP{} AP points, and
\ControlMemberFOne{} F1 points. All three intervals include zero
(Table~\ref{tab:final-component-paired}), so this comparison does not establish
a mean ensemble gain.

\begin{table}[!t]
\centering
\caption{\textbf{Relation features add a small ranking gain beyond independent-state features.} Differences are percentage points with 95\% paired component-bootstrap intervals.}
\label{tab:final-component-paired}
\small
\setlength{\tabcolsep}{4pt}
\begin{tabularx}{\linewidth}{@{}Xrrr@{}}
\toprule
\rowcolor{tablegroupgray}
Contrast & $\Delta$AUROC & $\Delta$AP & $\Delta$F1 \\
\midrule
{\raggedright Full $-$ independent-state \mbox{control}\par} & +0.076 [+0.027, +0.136] & +0.063 [+0.023, +0.112] & +0.026 [+0.001, +0.075] \\
Full $-$ single-member mean & +0.032 [-0.061, +0.144] & +0.029 [-0.041, +0.112] & +0.340 [-0.275, +0.917] \\
Full $-$ Boundary-only & +1.744 [+1.229, +2.328] & +1.240 [+0.874, +1.643] & +2.286 [+1.595, +3.131] \\
Full $-$ $\max(B,O)$ & -0.125 [-0.243, -0.027] & -0.106 [-0.194, -0.030] & -0.168 [-0.350, -0.036] \\
\bottomrule
\end{tabularx}
\end{table}

\Needspace{8\baselineskip}
The primary contrast is the seven-backbone mean source-macro AUROC difference
between Full and the independent-state control. Its 95\% interval is
\ControlRelationAUROCCI{} percentage points. We use 2,000 paired component-bootstrap
draws, stratified by source-membership signature, with shared component
multiplicities across methods, members, and backbones. Every draw recomputes
AUROC and AP from the saved scores, including ties. Intervals use the 2.5th
and 97.5th percentiles and are conditional on the fitted models. A two-sided
test with 5,000 shared component-level method swaps gives raw
$p=\ControlRelationRawP$; seed 20260910 fixes both procedures.
The intervals in Table~\ref{tab:final-component-paired} are pointwise.

\begin{table}[!t]
\centering
\caption{\textbf{Multi-state evidence complements Boundary on all seven backbones.} Source-macro metrics (\%); the single-member row averages offsets 300, 400, and 500.}
\label{tab:final-component-backbones}
\small
\setlength{\tabcolsep}{4pt}
\begin{tabularx}{\linewidth}{@{}lXrrrrr@{}}
\toprule
\rowcolor{tablegroupgray}
Backbone & Configuration & AUROC$\uparrow$ & AP$\uparrow$ & F1$\uparrow$ & FPR$\downarrow$ & Recall$\uparrow$ \\
\midrule
Llama-3.1-8B & Full & 99.11 & 99.26 & 94.37 & 8.54 & 96.78 \\
 & Boundary-only & 97.63 & 98.19 & 92.52 & 8.54 & 93.37 \\
 & Independent-state control & 99.11 & 99.26 & 94.37 & 8.54 & 96.78 \\
 & Single-member mean & 98.95 & 99.12 & 93.73 & 8.88 & 96.00 \\
\cmidrule(lr){1-7}
Qwen2.5-7B & Full & 99.15 & 99.30 & 94.31 & 10.39 & 98.20 \\
 & Boundary-only & 97.01 & 97.79 & 91.95 & 10.39 & 93.90 \\
 & Independent-state control & 99.03 & 99.20 & 94.31 & 10.39 & 98.20 \\
 & Single-member mean & 99.06 & 99.24 & 93.82 & 11.43 & 98.09 \\
\cmidrule(lr){1-7}
Mistral-7B-v0.3 & Full & 98.66 & 98.92 & 93.42 & 10.48 & 96.53 \\
 & Boundary-only & 97.19 & 97.96 & 91.37 & 10.48 & 92.69 \\
 & Independent-state control & 98.59 & 98.88 & 93.42 & 10.48 & 96.53 \\
 & Single-member mean & 98.71 & 98.97 & 93.10 & 11.53 & 96.83 \\
\cmidrule(lr){1-7}
InternLM2.5-7B & Full & 99.25 & 99.36 & 94.60 & 8.99 & 97.60 \\
 & Boundary-only & 97.53 & 98.15 & 92.62 & 8.99 & 93.89 \\
 & Independent-state control & 99.01 & 99.17 & 94.42 & 8.99 & 97.26 \\
 & Single-member mean & 99.29 & 99.38 & 93.86 & 11.32 & 98.05 \\
\cmidrule(lr){1-7}
Qwen2.5-14B & Full & 98.90 & 99.12 & 94.33 & 8.04 & 96.28 \\
 & Boundary-only & 97.56 & 98.22 & 93.04 & 8.04 & 93.88 \\
 & Independent-state control & 98.88 & 99.11 & 94.33 & 8.04 & 96.28 \\
 & Single-member mean & 98.93 & 99.13 & 93.82 & 9.60 & 96.59 \\
\cmidrule(lr){1-7}
Phi-3-medium-14B & Full & 99.22 & 99.24 & 93.34 & 8.97 & 95.41 \\
 & Boundary-only & 97.30 & 97.66 & 89.23 & 8.97 & 88.39 \\
 & Independent-state control & 99.16 & 99.18 & 93.34 & 8.97 & 95.41 \\
 & Single-member mean & 99.12 & 99.14 & 93.33 & 7.99 & 94.65 \\
\cmidrule(lr){1-7}
Qwen3-30B-A3B & Full & 98.89 & 99.09 & 93.01 & 11.19 & 96.30 \\
 & Boundary-only & 96.73 & 97.63 & 90.66 & 11.19 & 91.95 \\
 & Independent-state control & 98.87 & 99.06 & 93.01 & 11.19 & 96.30 \\
 & Single-member mean & 98.88 & 99.09 & 93.35 & 11.77 & 97.43 \\
\bottomrule
\end{tabularx}
\end{table}

\FloatBarrier

\IfFileExists{Tables/large_id_loso.tex}{%
\subsection{Five-Source Large-ID Leave-One-Source-Out Evaluation}
\label{app:large-id-loso}
With the complete test ecosystem withheld from fitting, \system retains
meaningful detection performance, although it falls below within-ecosystem
results. It averages 73.2\% source-macro F1 and 80.3\% AUROC, retaining higher F1
than same-backbone RouteGuard and AgentLens on all seven backbones (all 14 paired
bootstrap intervals exclude zero). Recall and AP average 76.6\% and 82.0\%. Tables~\ref{tab:large-id-loso}
and~\ref{tab:large-id-loso-paired} give the aggregate results and
paired representation contrasts with 2,000-draw bootstrap confidence intervals.

The advantage is metric-dependent. The supervised Qwen text readout has
lower F1 (58.7\%) but higher AUROC (88.1\%) and AP (90.0\%). \system therefore
leads the same-backbone representation monitors in F1, not every comparator
on every metric. Its mean F1 also exceeds the 66.7\% positive-class F1 of an
all-positive rule, which has 100\% FPR on the source-balanced data.
Deployment still requires measuring recall and false alarms at the chosen
operating point; F1 alone does not establish a cold-start alert budget.
\begin{table*}[!t]
\centering
\caption{\textbf{Five-source Large-ID leave-one-source-out evaluation (\%).}
Sources are equally weighted. Brackets: 95\% intervals from 2,000
source-stratified family-bootstrap draws.}
\label{tab:large-id-loso}
\scriptsize
\setlength{\tabcolsep}{3.5pt}
\renewcommand{\arraystretch}{0.94}
\begin{tabular}{@{}llrrrr@{}}
\toprule
\textbf{Method} & \textbf{Backbone} & \textbf{Recall} &
\textbf{F1 [95\% CI]} & \textbf{AUROC} & \textbf{AP} \\
\midrule
Regex & Shared & 100.0 & 66.7 [66.7, 66.7] & 68.6 & 68.6 \\
SkillSieve L1 & Shared & 53.0 & 60.5 [54.3, 65.9] & 69.0 & 69.3 \\
Cisco scanner & Shared & 17.5 & 27.6 [21.2, 33.1] & 63.4 & 64.7 \\
Mistral judge & Mistral-7B & 20.3 & 30.5 [24.7, 35.8] & 81.0 & 81.1 \\
Supervised Qwen text & Qwen2.5-7B & 49.6 & 58.7 [52.8, 64.0] & 88.1 & 90.0 \\
\midrule
AgentLens & Llama-3.1-8B & 48.2 & 35.8 [34.1, 37.2] & 50.3 & 55.1 \\
RouteGuard & Llama-3.1-8B & 60.0 & 56.1 [50.7, 60.8] & 60.5 & 62.2 \\
\textbf{\system} & \textbf{Llama-3.1-8B} & \textbf{70.8} & \textbf{72.3 [66.9, 76.4]} & \textbf{78.3} & \textbf{80.9} \\
AgentLens & Qwen2.5-7B & 40.0 & 26.7 [26.7, 26.7] & 47.0 & 48.9 \\
RouteGuard & Qwen2.5-7B & 63.6 & 59.3 [54.0, 64.0] & 66.6 & 66.8 \\
\textbf{\system} & \textbf{Qwen2.5-7B} & \textbf{78.2} & \textbf{72.3 [69.5, 74.9]} & \textbf{77.8} & \textbf{79.5} \\
AgentLens & Mistral-7B & 82.3 & 63.0 [59.8, 66.1] & 59.8 & 63.6 \\
RouteGuard & Mistral-7B & 61.0 & 58.8 [53.7, 63.4] & 66.3 & 67.1 \\
\textbf{\system} & \textbf{Mistral-7B} & \textbf{74.1} & \textbf{73.6 [68.9, 77.4]} & \textbf{77.6} & \textbf{79.0} \\
AgentLens & InternLM2.5-7B & 69.8 & 49.9 [47.6, 51.7] & 53.7 & 52.5 \\
RouteGuard & InternLM2.5-7B & 60.6 & 57.7 [52.4, 62.7] & 64.6 & 64.2 \\
\textbf{\system} & \textbf{InternLM2.5-7B} & \textbf{75.7} & \textbf{73.4 [69.0, 77.2]} & \textbf{80.7} & \textbf{82.1} \\
AgentLens & Qwen2.5-14B & 53.6 & 37.8 [35.3, 40.0] & 45.3 & 51.5 \\
RouteGuard & Qwen2.5-14B & 64.0 & 60.9 [55.3, 66.1] & 65.9 & 64.7 \\
\textbf{\system} & \textbf{Qwen2.5-14B} & \textbf{78.1} & \textbf{73.1 [70.0, 75.8]} & \textbf{80.6} & \textbf{82.0} \\
AgentLens & Phi-3-medium-14B & 37.5 & 26.6 [25.8, 27.4] & 48.1 & 49.5 \\
RouteGuard & Phi-3-medium-14B & 59.3 & 55.7 [49.9, 60.6] & 65.8 & 65.0 \\
\textbf{\system} & \textbf{Phi-3-medium-14B} & \textbf{70.4} & \textbf{72.2 [67.9, 75.8]} & \textbf{82.1} & \textbf{84.1} \\
AgentLens & Qwen3-30B-A3B & 54.3 & 38.8 [37.3, 40.0] & 50.8 & 50.6 \\
RouteGuard & Qwen3-30B-A3B & 66.9 & 66.2 [61.3, 70.4] & 75.5 & 76.0 \\
\textbf{\system} & \textbf{Qwen3-30B-A3B} & \textbf{88.8} & \textbf{75.5 [72.9, 77.8]} & \textbf{84.8} & \textbf{86.4} \\
\bottomrule
\end{tabular}
\end{table*}

\begin{table}[!t]
\centering
\caption{\textbf{Paired five-source LOSO F1 differences (percentage points).}
\system minus each comparator, with 95\% source-stratified family-bootstrap CIs.}
\label{tab:large-id-loso-paired}
\scriptsize
\setlength{\tabcolsep}{4pt}
\begin{tabular}{@{}lrr@{}}
\toprule
\textbf{Backbone} & \textbf{vs. RouteGuard} & \textbf{vs. AgentLens} \\
\midrule
Llama-3.1-8B & +16.1 [+10.0, +22.3] & +36.5 [+31.0, +40.9] \\
Qwen2.5-7B & +13.0 [+8.4, +18.5] & +45.6 [+42.9, +48.2] \\
Mistral-7B & +14.7 [+8.7, +20.8] & +10.6 [+4.7, +16.0] \\
InternLM2.5-7B & +15.7 [+9.7, +22.2] & +23.5 [+18.7, +28.1] \\
Qwen2.5-14B & +12.2 [+6.3, +18.3] & +35.3 [+31.8, +38.7] \\
Phi-3-medium-14B & +16.5 [+10.3, +22.7] & +45.6 [+41.4, +49.3] \\
Qwen3-30B-A3B & +9.3 [+4.6, +14.5] & +36.7 [+33.7, +39.5] \\
\bottomrule
\end{tabular}
\end{table}

\FloatBarrier
}{%
}

\section{Robustness and Enforcement Results}
\label{app:supporting-evidence}

The white-box and broker studies use the pair-disjoint application probes
defined in Appendix~\ref{app:data-governance}. They measure rewrite sensitivity
and request-level containment with the probes held fixed.

\subsection{White-Box Adversarial Robustness}
\label{app:adaptive}

The panel asks whether an attacker can manipulate visible D/O/B geometry while
preserving the original operation. It contains 256 family-unique matched pairs
from all five Large-ID sources. Source quotas are set by the evaluation protocol. Eligible
package roots are at most 16,384 bytes and admit a safe Operation newline split.
Each pair contributes clean, region-padding, cross-file-split, and six
Geometry-candidate views. This yields 4,608 package views per backbone. The
attacker knows the native tokenizer, D/O/B binding, fold-specific detector
bundle, fusion, and threshold. It may rewrite package documentation and layout
before capability loading, but it cannot modify model or detector parameters.
Geometry evasion chooses the lowest Geometry-margin candidate from the malicious
package alone. We then apply the same candidate index to its matched benign
control.

The three candidate rewrite mechanisms are detailed in Box~D.1. Exact deletion or ordered concatenation reconstructs every original package. Every original non-root file is byte-identical, and no original operation byte is deleted. No package content is executed.

\begin{protocolbox}[title={\textbf{Protocol Box D.1:} White-Box Adversarial Rewrite Transformations},label={box:whitebox-rewrites}]
\small
\begin{itemize}[leftmargin=*,itemsep=2.5pt,topsep=1pt]
  \item \textbf{Region Padding:} Injects 1,024--2,048 marked inert characters split across $\mathcal{D}$ and $\mathcal{O}$ spans to alter relative token coordinates and test sensitivity to document expansion.
  \item \textbf{Cross-File Splitting:} Moves the complete Operation suffix into at most four ordered Markdown fragments separated at safe newline boundaries outside code fences.
  \item \textbf{Geometry Evasion Search:} Searches six fixed candidate 768-character inert placements spanning description, operation, and early or late auxiliary positions; selects the candidate minimizing Geometry logit margin $m_G(x)$ on the malicious package.
\end{itemize}
\end{protocolbox}

Every view is scored by the detector fitted for its original Large-ID test
fold, without refitting or a threshold change. The inference unit is
the matched family, and metrics average the five sources equally.
Intervals use 2,000 source-stratified family-bootstrap draws. Fused and
Geometry logit-margin shifts use 10,000 source-stratified family sign flips. We apply
separate Holm corrections across the 21 backbone--attack tests. The robustness
criterion requires at least 80\% TPR retention and an empirical-bootstrap
95\% upper endpoint of at most 25\% for attack success.

In Table~\ref{tab:whitebox-adversarial-full}, $\Delta m$ is the attacked-minus-clean
fused logit margin, without percentage scaling. $p_{\mathrm H}(m)$ and
$p_{\mathrm H}(m_G)$ test fused and Geometry margin shifts, respectively.

TPR retention is the attacked source-macro TPR divided by the clean
source-macro TPR, expressed as a percentage. It can exceed 100\% when rewriting
makes previously missed attacks detectable. ASR first divides successful
pass-after-rewrite evasions by clean true positives within each source, then
averages the defined source rates. Thus its denominator is the clean-detected
subset of the 256 malicious packages, not all package views. Benign FPR uses
the 256 matched benign packages with equal source weighting.
When no evasion is observed, every bootstrap replicate also contains zero
evasions, producing $[0,0]$. This interval reflects the observed sample;
it does not bound the population evasion rate at zero.

For a zero-event cell with $n_s$ clean true positives in source $s$, a
source-wise binomial model gives the one-sided upper endpoint
$u_s=1-(0.05/5)^{1/n_s}$. Bonferroni coverage then yields the conservative
95\% upper bound $U_{\mathrm{macro}}=\frac{1}{5}\sum_s u_s$.
Table~\ref{tab:whitebox-zero-event-bounds} reports the eight zero-event cells
using their actual source denominators. These bounds assume independent
Bernoulli family trials with a common evasion probability within each source,
conditional on clean detection, the selected support, and the tested rewrite.
They require no independence across sources and apply per cell, not
simultaneously across all 21 cells. Dependence between content-linked families
(Appendix~\ref{app:data-governance}) remains outside that model.

\begin{table}[!t]
\centering
\caption{\textbf{One-sided 95\% macro-ASR bounds for zero-event cells.}
Counts follow the source order: agent-skill malware, ATR, MalSkillBench,
SkillFortify, SkillsBench-1650. Listed rewrites denote separate cells.}
\label{tab:whitebox-zero-event-bounds}
\small
\setlength{\tabcolsep}{4pt}
\begin{tabular}{@{}lllr@{}}
\toprule
Backbone & Rewrite & Clean-TP counts & Upper bound\\
\midrule
Llama-3.1-8B & Padding & 23, 21, 70, 103, 31 & 12.48\%\\
Qwen2.5-7B & Padding, Geometry & 23, 22, 70, 103, 31 & 12.32\%\\
Mistral-7B & Padding, Geometry & 27, 21, 70, 103, 32 & 11.90\%\\
InternLM2.5-7B & Padding, Geometry & 26, 21, 70, 103, 30 & 12.18\%\\
Qwen2.5-14B & Padding & 22, 22, 69, 103, 31 & 12.48\%\\
\bottomrule
\end{tabular}
\end{table}

The supplementary bounds span 11.90--12.48\%. The table below also reports
empirical-bootstrap intervals for all cells; the binomial bounds apply only
to zero-event cells.

\begin{table}[!t]
\centering
\caption{\textbf{Operation-preserving rewrites retain detection but increase benign alerts.} Source-macro rates (\%) on 256 matched pairs; ASR conditions on clean true positives. Brackets: 95\% source-stratified family-bootstrap CIs. $\Delta m$ is a logit shift.}
\label{tab:whitebox-adversarial-full}
\scriptsize
\setlength{\tabcolsep}{2.2pt}
\begin{tabular}{@{}llrrrrrrc@{}}
\toprule
Backbone & Attack & TPR ret. & ASR [95\% CI] & Benign FPR $c{\to}a$ & $\Delta m$ [95\% CI] & $p_{\mathrm{H}}(m)$ & $p_{\mathrm{H}}(m_G)$ & Pass \\
\midrule
Llama-3.1-8B & Region padding & 105.3 & 0.0 [0.0, 0.0] & $34.4{\to}76.4$ & 5.8 [5.1, 6.5] & 0.0021 & 0.0021 & yes \\
 & Cross-file split & 92.8 & 11.0 [6.5, 16.3] & $34.4{\to}55.0$ & -0.6 [-1.3, 0.2] & 0.9179 & 0.0021 & yes \\
 & Geometry evasion & 101.6 & 3.1 [0.6, 6.1] & $34.4{\to}65.7$ & 3.7 [2.9, 4.5] & 0.0021 & 0.0021 & yes \\
\addlinespace[1.5pt]
Qwen2.5-7B & Region padding & 105.0 & 0.0 [0.0, 0.0] & $32.3{\to}78.1$ & 6.9 [6.2, 7.6] & 0.0021 & 0.7895 & yes \\
 & Cross-file split & 92.3 & 9.0 [4.2, 14.0] & $32.3{\to}57.1$ & -3.7 [-4.3, -3.1] & 0.0021 & 0.0021 & yes \\
 & Geometry evasion & 105.0 & 0.0 [0.0, 0.0] & $32.3{\to}78.5$ & 7.5 [6.7, 8.4] & 0.0021 & 0.0021 & yes \\
\addlinespace[1.5pt]
Mistral-7B & Region padding & 102.3 & 0.0 [0.0, 0.0] & $40.8{\to}80.4$ & 4.6 [4.0, 5.2] & 0.0021 & 0.0021 & yes \\
 & Cross-file split & 93.2 & 7.1 [3.2, 11.6] & $40.8{\to}57.9$ & -3.6 [-4.1, -3.0] & 0.0021 & 0.0021 & yes \\
 & Geometry evasion & 100.7 & 0.0 [0.0, 0.0] & $40.8{\to}73.1$ & 5.5 [4.8, 6.3] & 0.0021 & 0.0021 & yes \\
\addlinespace[1.5pt]
InternLM2.5-7B & Region padding & 104.4 & 0.0 [0.0, 0.0] & $38.1{\to}73.6$ & 5.4 [4.9, 6.0] & 0.0021 & 0.7895 & yes \\
 & Cross-file split & 100.4 & 1.1 [0.0, 2.8] & $38.1{\to}65.6$ & -0.3 [-0.9, 0.3] & 1.0000 & 0.0021 & yes \\
 & Geometry evasion & 104.4 & 0.0 [0.0, 0.0] & $38.1{\to}75.4$ & 6.0 [5.4, 6.6] & 0.0021 & 0.0021 & yes \\
\addlinespace[1.5pt]
Qwen2.5-14B & Region padding & 103.9 & 0.0 [0.0, 0.0] & $35.8{\to}68.1$ & 1.9 [1.3, 2.5] & 0.0021 & 0.0021 & yes \\
 & Cross-file split & 98.5 & 3.4 [0.6, 6.7] & $35.8{\to}57.4$ & 0.1 [-0.6, 0.7] & 1.0000 & 0.0021 & yes \\
 & Geometry evasion & 105.1 & 1.2 [0.0, 3.3] & $35.8{\to}61.0$ & 3.7 [3.0, 4.4] & 0.0021 & 0.0021 & yes \\
\addlinespace[1.5pt]
Phi-3-medium-14B & Region padding & 103.6 & 0.9 [0.0, 2.7] & $34.3{\to}74.7$ & 3.5 [2.9, 4.1] & 0.0021 & 0.0021 & yes \\
 & Cross-file split & 93.6 & 8.2 [4.3, 12.8] & $34.3{\to}64.1$ & 1.2 [0.7, 1.6] & 0.0021 & 0.0021 & yes \\
 & Geometry evasion & 101.9 & 1.8 [0.0, 4.5] & $34.3{\to}57.5$ & 2.7 [2.2, 3.2] & 0.0021 & 0.0021 & yes \\
\addlinespace[1.5pt]
Qwen3-30B-A3B & Region padding & 101.6 & 2.1 [0.0, 4.8] & $35.1{\to}62.8$ & -1.1 [-1.7, -0.5] & 0.0021 & 0.0021 & yes \\
 & Cross-file split & 101.4 & 1.7 [0.0, 4.3] & $35.1{\to}70.9$ & 0.1 [-0.6, 0.8] & 1.0000 & 0.0021 & yes \\
 & Geometry evasion & 102.9 & 0.9 [0.0, 2.9] & $35.1{\to}49.3$ & 0.3 [-0.4, 0.9] & 1.0000 & 0.0021 & yes \\
\bottomrule
\end{tabular}
\end{table}

All 21 backbone--attack cells satisfy this empirical anti-evasion criterion. Cross-file
splitting is the strongest attack. Its equal-source ASR reaches 11.0\% [6.5,
16.3] on Llama, while TPR retention remains at least 92.3\% on Qwen2.5-7B.
Region padding and Geometry search achieve TPR retention of at least 101.6\%
and 100.7\%,
respectively. Geometry-search ASR is at most 3.1\%. The selected rewrite changes
the Geometry margin in 19/21 cells after Holm correction, but these changes
rarely produce successful evasion. Manipulating one exposed component therefore
does not imply control of the fused verdict.

The matched benign controls expose a different cost: the attacks raise alert
load even when they rarely evade detection. Clean equal-source FPR is
32.3--40.8\% on this length- and split-eligible subset and becomes 49.3--80.4\%
after the same inert rewrites. This panel therefore measures both evasion
resistance and the benign interruption cost of the fixed rule. Canonicalization
and alert-budget calibration are deployment choices to evaluate against that
cost, not validated mitigations in this panel.
\FloatBarrier

\subsection{Capability-Broker Enforcement}
\label{app:capability-enforcement}

\begin{table}[!t]
\centering
\caption{\textbf{Request-level enforcement across seven backbones.} Cells give mean ASR $\downarrow$ / valid benign allowance $\uparrow$ (\%) on 256 matched pairs, using all trials of each label as denominators.}
\label{tab:real-agent-enforcement}
\small
\setlength{\tabcolsep}{2.5pt}
\begin{tabular}{lrrrrr}
\toprule
Budget & Random & AgentLens & RouteGuard & \textbf{\system} & Oracle \\
\midrule
10\% & 41.9/42.2 & 42.8/41.5 & 36.5/46.9 & \textbf{36.7/47.1} & 31.7/47.1 \\
20\% & 37.3/37.4 & 36.1/39.5 & 27.3/46.4 & \textbf{26.9/47.0} & 20.3/47.1 \\
30\% & 32.9/32.6 & 28.2/38.3 & 19.6/45.8 & \textbf{17.2/46.7} & 11.2/47.1 \\
\bottomrule
\end{tabular}
\end{table}

\begin{table}[!t]
\centering
\caption{\textbf{Enforcement by acting backbone.} Valid: typed-request validity (\%). Other cells: ASR $\downarrow$ / valid benign allowance $\uparrow$, as in Table~\ref{tab:real-agent-enforcement}.}
\label{tab:real-agent-enforcement-backbones}
\scriptsize
\setlength{\tabcolsep}{1.8pt}
\renewcommand{\arraystretch}{0.96}
\begin{tabular}{lrrrrr}
\toprule
Acting backbone & Budget & Valid & AgentLens & RouteGuard & \textbf{\system} \\
\midrule
Qwen2.5-7B & 10\% & 86.3 & 71.1/9.0 & 56.2/15.2 & \textbf{60.9/15.2} \\
 & 20\% &  & 64.1/7.0 & 43.0/14.8 & \textbf{45.7/15.2} \\
 & 30\% &  & 52.3/7.0 & 31.6/14.5 & \textbf{32.8/15.2} \\
Llama-3.1-8B & 10\% & 99.0 & 61.7/21.1 & 63.7/21.1 & \textbf{61.3/21.1} \\
 & 20\% &  & 46.9/20.7 & 48.8/21.1 & \textbf{45.7/21.1} \\
 & 30\% &  & 35.5/19.5 & 32.8/21.1 & \textbf{28.9/21.1} \\
Mistral-7B & 10\% & 95.5 & 19.9/82.0 & 19.5/81.6 & \textbf{20.3/82.0} \\
 & 20\% &  & 13.3/81.6 & 14.8/80.9 & \textbf{14.1/81.6} \\
 & 30\% &  & 9.4/79.7 & 10.9/80.1 & \textbf{9.0/80.9} \\
InternLM2.5-7B & 10\% & 65.8 & 55.5/23.4 & 44.5/25.0 & \textbf{43.4/25.0} \\
 & 20\% &  & 49.6/22.7 & 36.3/25.0 & \textbf{29.3/25.0} \\
 & 30\% &  & 37.9/22.7 & 28.1/24.2 & \textbf{19.5/25.0} \\
Qwen2.5-14B & 10\% & 98.0 & 47.3/68.4 & 35.5/85.5 & \textbf{35.5/85.9} \\
 & 20\% &  & 43.8/59.4 & 21.5/84.4 & \textbf{25.8/85.5} \\
 & 30\% &  & 34.8/59.4 & 13.3/82.8 & \textbf{14.1/85.2} \\
Phi-3-medium & 10\% & 32.6 & 22.3/10.9 & 21.1/11.3 & \textbf{20.7/11.3} \\
 & 20\% &  & 18.8/9.4 & 16.0/11.3 & \textbf{16.8/11.3} \\
 & 30\% &  & 15.2/9.0 & 12.1/11.3 & \textbf{10.5/11.3} \\
Qwen3-30B-A3B & 10\% & 99.4 & 21.9/75.8 & 14.8/88.7 & \textbf{14.8/89.1} \\
 & 20\% &  & 16.4/75.8 & 10.9/87.5 & \textbf{10.9/89.1} \\
 & 30\% &  & 12.1/70.7 & 8.2/86.3 & \textbf{5.5/88.3} \\
\bottomrule
\end{tabular}
\end{table}

\FloatBarrier

This experiment measures request-level policy enforcement on
256 family-unique matched safe--unsafe pairs from all five ecosystems. Sample
selection uses neither detector scores nor acting-model outputs. Each of the
seven protected models loads the complete package through its native chat and
tool template. It then encodes the candidate operation as one typed \texttt{request\_capability} action (Box~D.2). The schema requires scope, destination, authorization, and persistence arguments. Malformed requests fail closed.

\begin{codebox}[title={\textbf{Action Box D.2:} Typed Broker Action Schema: \texttt{request\_capability}},label={box:broker-schema}]
{
  "action": "request_capability",
  "arguments": {
    "scope": "filesystem" | "network" | "subprocess" | "credentials",
    "destination": "<target_file_path | remote_endpoint_url>",
    "authorization": "<policy_token | user_confirmation | none>",
    "persistence": true | false
  }
}
\end{codebox}

A policy-independent parser records each request before applying a policy.
AgentLens, RouteGuard, and \system route the same recorded requests using their
package scores under 10, 20, and 30\% containment budgets. Only the routing score
changes. The inert broker records \textsc{allow} or \textsc{contain}; even
\textsc{allow} executes no package payload, generated command, or network
request. All models use greedy decoding. Qwen3 disables its optional reasoning
trace for this deterministic
typed-encoding task. Confidence intervals use 10,000 paired bootstrap draws
over source-stratified operation pairs.

Let $V$ indicate a parser-valid typed request, $U$ an unsafe argument
combination, and $C$ selection for score-based containment. The categorical
checker marks unrelated or private resources, unapproved or private destinations, absent
or bypassed authorization, persistence, and unknown values as unsafe.
For the budgeted policies, allowance is $A=V(1-C)$. This Boolean allowance is
request allowance, not payload execution.
For package label $y$, the reported endpoints are
\[
\mathrm{ASR}=\frac{\sum_{y=1}VU A}{N_{\mathrm{unsafe}}},\qquad
\mathrm{BenignRequestAllowance}
=\frac{\sum_{y=0}V(1-U)A}{N_{\mathrm{safe}}}.
\]
Here $N_{\mathrm{unsafe}}=N_{\mathrm{safe}}=256$ per acting model.
Benign request allowance counts valid, checker-safe requests that the broker
allows; it does not measure completion of an executed task.
Request validity is $\sum V/512$;
score-based policy rejection is $\sum C/512$, with malformed requests
additionally contained. Conditional attack success divides the same unsafe
allowance count by $\sum_{y=1}VU$. The table reports the all-unsafe denominator.
For paired intervention comparisons, the prevented-attack count is
$\sum_{y=1}VUC$; $1-\mathrm{ASR}$ also includes invalid or non-unsafe requests
and is not the broker's blocking rate.

Table~\ref{tab:real-agent-enforcement} averages package-level rates over seven
acting models. Random averages 20 seeds; Oracle uses labels.
Allow-all records 82.4\% valid requests, 46.7\% unsafe-request
allowance, and 47.1\% benign-request allowance. Against AgentLens, \system
reduces unsafe-request allowance by 6.1, 9.2, and 11.0 points at the three
budgets and increases benign-request allowance by 5.6, 7.5, and 8.4 points.
Both metrics favor \system in a fraction 1.0 of bootstrap draws at each budget.
At 30\% containment, the corresponding differences from RouteGuard are
$-2.4$ points (95\% CI $[-4.4,-0.4]$) and $+0.95$ points
($[0.22,1.67]$); both metrics favor \system in a fraction 0.990 of draws.

\section{Operation Sensitivity and Localization}
\label{app:mechanism-evidence}
\subsection{Balanced Counterfactual Diagnosis}
\label{app:diagnostics}

The balanced $2\times2$ design crosses visible surface with operation
semantics. B/B denotes benign surface and benign operation. B/M keeps the
benign surface but inserts a malicious operation. S/B combines suspicious
surface with benign operation, while S/M is suspicious in both dimensions.
Each quadrant has 57 packages, so lexical appearance and operation labels are
independently balanced.

\begin{figure}[!t]
  \centering
  \includegraphics[width=\linewidth]{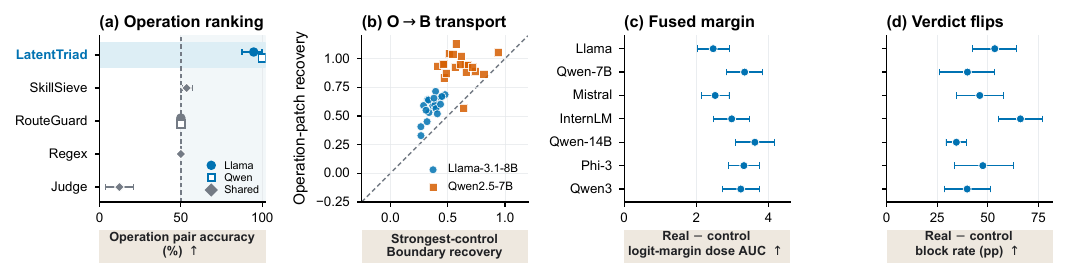}
  \caption{\textbf{Operation interventions change Boundary states and frozen
  policy verdicts.} (a) Operation ranking. (b) One-layer Boundary recovery.
  (c,d) Seven-depth intervention effects on margins and eligible verdicts.
  Bars are 95\% intervals; (c,d) use source-stratified family bootstrap.}
  \label{fig:mechanism-interventions}
\end{figure}

Llama \system ranks malicious operations above their matched benign
counterparts in 94.7\% of pairs, and Qwen orders every pair correctly.
RouteGuard ties every pair. These results separate operation-sensitive score
ordering from package-surface classification. Figure~\ref{fig:mechanism-interventions}a
shows all methods and paired intervals.

\FloatBarrier
\subsection{Auxiliary Risk Localization}
\label{app:region-attribution}

The localization experiment predicts the region containing risky content. It
reuses hidden states from the same native prefill and fits two $C=1$
L2-logistic readouts: one on Declaration features and one on Code/Operation
features. Thresholding both outputs at 0.5 gives four classes:
\textsc{clean} $(0,0)$, \textsc{decl} $(1,0)$, \textsc{code} $(0,1)$, and
\textsc{both} $(1,1)$. These auxiliary predictions enter neither the three-branch
score nor the policy threshold. The tables' attribution labels denote
content localization, not causal attribution of the primary decision.

The natural panel contains 610 packages from 150 grouped families: 210
\textsc{clean}, 115 \textsc{decl}, 175 \textsc{code}, and 110 \textsc{both}.
It spans five source ecosystems and six malicious-risk categories. The
controlled panel contains 128 base families, each represented by clean,
Declaration-only, Code-only, and dual-region variants. All members of a family
remain in one of five outer folds. The reported intervals resample families
within source 2,000 times.

In Table~\ref{tab:region-attribution-full}, D/C denotes the Declaration and
Code heads. ``Blocked'' counts primary-detector blocks among 400 malicious
packages; conditional exact attribution measures correctness within those
blocks. Joint success requires both the correct primary verdict and exact
attribution among all 610 natural packages.

\begin{table*}[!t]
\centering
\caption{\textbf{Risk localization across seven backbones.}
Rates are percentages; D/C AUROC uses 0--1.
Natural Macro-F1 includes 95\% family-bootstrap intervals.}
\label{tab:region-attribution-full}
\scriptsize
\setlength{\tabcolsep}{3.0pt}
\begin{tabular}{lrrrrrrr}
\toprule
Backbone & Nat. Acc. & Nat. Macro-F1 [95\% CI] & D/C AUROC & Ctrl. Macro-F1 & Blocked & Cond. exact & Joint \\
\midrule
Llama-3.1-8B & 85.7 & 86.2 [83.2, 88.8] & 0.998/0.992 & 90.9 & 375/400 & 92.0 & 85.7 \\
Qwen2.5-7B & 82.1 & 82.4 [79.6, 85.3] & 0.998/0.984 & 89.2 & 370/400 & 89.5 & 82.1 \\
Mistral-7B-v0.3 & 82.1 & 82.7 [79.7, 85.5] & 0.998/0.985 & 86.2 & 369/400 & 88.6 & 82.1 \\
InternLM2.5-7B & 79.7 & 80.7 [77.7, 83.5] & 0.996/0.981 & 86.2 & 363/400 & 89.5 & 79.7 \\
Qwen2.5-14B & 87.7 & 88.1 [85.5, 90.6] & 0.999/0.994 & 90.4 & 373/400 & 93.6 & 87.7 \\
Phi-3-medium-14B & 83.1 & 84.0 [81.0, 86.9] & 0.997/0.988 & 90.9 & 373/400 & 89.0 & 83.1 \\
Qwen3-30B-A3B & 87.5 & 88.0 [85.4, 90.5] & 0.999/0.993 & 92.3 & 371/400 & 94.3 & 87.5 \\
\bottomrule
\end{tabular}
\end{table*}

\begin{table}[!t]
\centering
\caption{\textbf{Auxiliary attribution cost.} Times measure only the two
diagnostic heads after the native prefill. The heads reuse the detector's hidden
states and add no model forward.}
\label{tab:region-attribution-runtime}
\scriptsize
\setlength{\tabcolsep}{3.5pt}
\begin{tabular}{lrrrrr}
\toprule
Backbone & Med. ms & P95 ms & Prefill \% & VRAM MB & Fwds. \\
\midrule
Llama-3.1-8B & 1.8 & 3.1 & 1.26 & 14.2 & 1 \\
Qwen2.5-7B & 1.6 & 2.8 & 1.24 & 12.8 & 1 \\
Mistral-7B-v0.3 & 1.7 & 2.9 & 1.30 & 13.5 & 1 \\
InternLM2.5-7B & 1.7 & 3.0 & 1.26 & 13.1 & 1 \\
Qwen2.5-14B & 2.4 & 4.1 & 1.07 & 18.6 & 1 \\
Phi-3-medium-14B & 2.3 & 3.9 & 1.05 & 17.9 & 1 \\
Qwen3-30B-A3B & 3.1 & 5.2 & 0.99 & 24.5 & 1 \\
\bottomrule
\end{tabular}
\end{table}

Because the two heads reuse cached D/O states, their median cost is 1.6--3.1\,ms
and their peak additional memory is below 25\,MB on every backbone.

\FloatBarrier
\subsection{Exact-Slot \texorpdfstring{O$\rightarrow$B}{O-to-B} Causal Confirmation}
\label{app:mechanism-interventions}

We test whether Operation states affect the pre-action Boundary, beyond
supporting a diagnostic readout. The test contains 40 source--target families. Within each
family, operation slots have equal token counts and identical absolute
Operation and Boundary positions under both native tokenizers. A score-blind
odd/even split assigns 20 families to development and 20 to confirmation.
Development selects the best of seven normalized intervention layers by its
smallest mean margin over the controls (Llama layer 3 and Qwen layer 9). No confirmation
family informs layer selection, controls, or gates.

At the selected layer, the exact source Operation token states replace the
aligned target states. Let $\mathbf b_t$, $\mathbf b_s$, and $\mathbf b_p$
denote the target, donor, and patched Boundary vectors. Recovery is the
normalized projection of the patch-induced displacement onto the donor direction:
\[
\mathrm{Recovery}=
\frac{(\mathbf b_p-\mathbf b_t)^\top(\mathbf b_s-\mathbf b_t)}
{\|\mathbf b_s-\mathbf b_t\|_2^2+10^{-8}}.
\]
It measures directional displacement, not fractional distance reduction.
Four controls isolate specificity: eight same-energy random directions at the same
slot, the semantic delta at an equal-length neutral slot, a permutation of the
source token order, and a per-token rescaled delta from another family.
Confidence intervals use 10,000 paired family-bootstrap replicates.

Exact replacement produces positive Boundary recovery in all 20 confirmation
families on each backbone. Llama's median recovery is 0.592. Its mean recovery gains
over the random, neutral-position, token-permutation, and mismatched-semantic
controls are 0.420 [0.379, 0.457], 0.439 [0.407, 0.474], 0.323 [0.277, 0.373],
and 0.219 [0.186, 0.251], respectively. Qwen's median recovery is 0.944. Its
corresponding gains are 0.743 [0.672, 0.804], 0.876 [0.823, 0.929], 0.540
[0.464, 0.613], and 0.366 [0.277, 0.451]. Every confirmation gate passes on
both backbones. Replacing Operation states therefore moves the downstream
Boundary toward the donor state on held-out families. This measures causal
O$\rightarrow$B influence; the learned transport residual is a separate
statistical proxy. The following experiment measures application-probe response to
multi-depth Operation interventions.

\subsection{Detector-Aligned Score and Verdict Test}
\label{app:detector-aligned-readout}

This experiment uses 100 distinct Large-ID matched families, with 20 pairs drawn
from each source by a deterministic, score-blind rule. Every pair uses the
fixed application probe for which its matched family was outer-test data.
It recomputes the eight-depth input with a compact limit of eight D and 24 O
atoms, reusing the Large-ID projections, member encoders, and final readouts.
No head is refit and no additional ensemble is formed.
The intervention starts from the benign package and inserts
the matched malicious Operation trajectory at doses 0.25, 0.50, and 1.00. It
patches the seven causally upstream depths of the detector's eight-depth
trajectory; the final depth is excluded because it cannot alter the same-layer
Boundary state.

At each depth, Operation atoms are aligned by normalized order. Their
malicious-minus-benign mean difference is added to all target-atom tokens. Two
controls use independent Gaussian directions at the same atoms and depths, with
every atom-level norm matched to the real intervention. We then recompute the
Boundary state, the three component margins, and the fixed-threshold verdict.
For component $k\in\{B,O,G\}$, the logit margin is
$m_k=\log[p_k/(1-p_k)]$. We report the fused margin
\begin{equation}
m(x)=\max\{m_B(x),\min(m_O(x),m_G(x))\}
     =\operatorname{logit}s(x).
\label{eq:fused-margin}
\end{equation}
Thus $m(x)\geq0$ gives the same verdict as $s(x)\geq0.5$ in
Equation~\ref{eq:detector-gate}. The intervention generates no tokens and
executes no payload.

Point estimates weight the five sources equally. Confidence intervals use
10,000 source-stratified family-bootstrap replicates. One-sided family sign-flip
tests compare the real intervention with its random controls; Holm correction is
applied across seven backbones separately for each endpoint. Eligibility for the
verdict analysis is fixed before intervention: the benign score must be below
the operating threshold and its matched malicious score at or above it. For each
pair, dose AUC is the trapezoidal integral of the real-minus-control response
over doses $\{0,0.25,0.50,1.00\}$.

Branch activity is measured on unpatched benign targets. Verdict $\Delta$
is the full-dose real-minus-random pass-to-block rate among eligible pairs.

\begin{table}[!t]
\centering
\caption{\textbf{Operation interventions shift Boundary and fused margins.} Dose AUC is relative to norm-matched random controls. Brackets: 95\% source-stratified family-bootstrap CIs; $^{*}$: Holm-corrected positive sign-flip test.}
\label{tab:readout-path-effects}
\scriptsize
\setlength{\tabcolsep}{3.2pt}
\renewcommand{\arraystretch}{1.13}
\begin{tabular}{@{}lrrrrrr@{}}
\toprule
Backbone & \shortstack{Natural\\PairAcc (\%)} & \shortstack{Boundary\\recovery} & \shortstack{Boundary\\margin} & \shortstack{Operation\\margin} & \shortstack{Geometry\\margin} & \shortstack{Fused\\margin $m(x)$} \\
\midrule
Llama-3.1-8B & \shortstack{$89.0^{*}$\\$[83.0,94.0]$} & \shortstack{$0.096^{*}$\\$[0.077,0.116]$} & \shortstack{$2.880^{*}$\\$[2.281,3.473]$} & \shortstack{$1.496^{*}$\\$[1.149,1.839]$} & \shortstack{$0.639$\\$[0.082,1.197]$} & \shortstack{$2.465^{*}$\\$[2.035,2.920]$} \\
Qwen2.5-7B & \shortstack{$90.0^{*}$\\$[84.0,95.0]$} & \shortstack{$0.174^{*}$\\$[0.151,0.196]$} & \shortstack{$4.436^{*}$\\$[3.836,5.028]$} & \shortstack{$1.632^{*}$\\$[1.245,2.013]$} & \shortstack{$2.703^{*}$\\$[2.175,3.230]$} & \shortstack{$3.336^{*}$\\$[2.826,3.841]$} \\
Mistral-7B & \shortstack{$86.0^{*}$\\$[79.0,92.0]$} & \shortstack{$0.107^{*}$\\$[0.091,0.123]$} & \shortstack{$2.554^{*}$\\$[2.077,3.043]$} & \shortstack{$2.429^{*}$\\$[2.032,2.842]$} & \shortstack{$1.833^{*}$\\$[1.392,2.272]$} & \shortstack{$2.520^{*}$\\$[2.134,2.902]$} \\
InternLM2.5-7B & \shortstack{$88.0^{*}$\\$[82.0,94.0]$} & \shortstack{$0.125^{*}$\\$[0.111,0.138]$} & \shortstack{$3.219^{*}$\\$[2.593,3.829]$} & \shortstack{$2.117^{*}$\\$[1.718,2.527]$} & \shortstack{$1.300^{*}$\\$[0.794,1.794]$} & \shortstack{$2.978^{*}$\\$[2.470,3.479]$} \\
Qwen2.5-14B & \shortstack{$88.0^{*}$\\$[82.0,94.0]$} & \shortstack{$0.111^{*}$\\$[0.093,0.130]$} & \shortstack{$4.452^{*}$\\$[3.868,5.040]$} & \shortstack{$2.221^{*}$\\$[1.782,2.677]$} & \shortstack{$1.959^{*}$\\$[1.408,2.515]$} & \shortstack{$3.621^{*}$\\$[3.089,4.159]$} \\
Phi-3-medium-14B & \shortstack{$89.0^{*}$\\$[83.0,94.0]$} & \shortstack{$0.125^{*}$\\$[0.107,0.143]$} & \shortstack{$4.500^{*}$\\$[3.827,5.187]$} & \shortstack{$2.188^{*}$\\$[1.745,2.646]$} & \shortstack{$0.349$\\$[-0.137,0.830]$} & \shortstack{$3.316^{*}$\\$[2.877,3.755]$} \\
Qwen3-30B-A3B & \shortstack{$92.0^{*}$\\$[87.0,97.0]$} & \shortstack{$0.119^{*}$\\$[0.105,0.133]$} & \shortstack{$4.930^{*}$\\$[4.298,5.567]$} & \shortstack{$2.100^{*}$\\$[1.667,2.532]$} & \shortstack{$1.606^{*}$\\$[1.162,2.048]$} & \shortstack{$3.230^{*}$\\$[2.731,3.755]$} \\
\bottomrule
\end{tabular}
\end{table}

\begin{table}[!t]
\centering
\caption{\textbf{Operation interventions change fixed-probe verdicts.} Verdict $\Delta$: full-dose real-minus-random pass-to-block rate on threshold-eligible pairs (pp). Brackets: 95\% source-stratified family-bootstrap CIs.}
\label{tab:readout-path-verdicts}
\scriptsize
\setlength{\tabcolsep}{5.0pt}
\begin{tabular}{@{}lrrrrrr@{}}
\toprule
Backbone & B active (\%) & O active (\%) & G active (\%) & Eligible pairs & Verdict $\Delta$ pp $[95\%\ \mathrm{CI}]$ & $p_{\mathrm{H}}$ \\
\midrule
Llama-3.1-8B & 33.0 & 51.0 & 16.0 & 54/100 & 53.4 [42.4, 64.3] & $<10^{-3}$ \\
Qwen2.5-7B & 22.0 & 48.0 & 30.0 & 58/100 & 39.9 [25.9, 53.1] & $<10^{-3}$ \\
Mistral-7B & 29.0 & 53.0 & 18.0 & 54/100 & 46.0 [34.5, 57.5] & $<10^{-3}$ \\
InternLM2.5-7B & 40.0 & 23.0 & 37.0 & 61/100 & 66.1 [55.1, 77.2] & $<10^{-3}$ \\
Qwen2.5-14B & 26.0 & 36.0 & 38.0 & 51/100 & 34.5 [29.3, 39.6] & $<10^{-3}$ \\
Phi-3-medium-14B & 30.0 & 38.0 & 32.0 & 60/100 & 47.5 [33.6, 62.9] & $<10^{-3}$ \\
Qwen3-30B-A3B & 24.0 & 44.0 & 32.0 & 58/100 & 39.8 [28.5, 51.2] & $<10^{-3}$ \\
\bottomrule
\end{tabular}
\end{table}

Natural malicious-over-benign PairAcc is 86--92\%. The real trajectory patch
produces positive Boundary-recovery, Boundary-head, Operation-head, and
fused logit-margin dose responses on all seven backbones after Holm correction
($p_{\mathrm H}<10^{-3}$ throughout). Geometry also responds on five. The final
logit-margin dose AUC ranges from 2.465 to 3.621. Depending on the backbone, 51--61 of
the 100 pairs are threshold-eligible. Within this set, the real patch raises
pass-to-block flips by 34.5--66.1 percentage points over the norm-matched
controls; every interval excludes zero. The single-depth experiment tests
O$\rightarrow$B influence. The multi-depth experiment measures joint three-branch
response and verdict sensitivity under compact extraction; neither isolates
Boundary as the necessary mediator of the fused decision.

\FloatBarrier
\section{Runtime and Resource Accounting}
\label{app:runtime-resources}
\subsection{Measurement Protocol and Costs}
\label{app:runtime}

Runtime measurements use the compact CDF-consensus configuration, rather
than the package detector. Table~\ref{tab:configuration-map} lists the
extraction and scoring settings for each study.

\begin{table}[!t]
\centering
\caption{\textbf{Extraction and readout configurations by study.}
Atom limits are D/O maxima.}
\label{tab:configuration-map}
\small
\setlength{\tabcolsep}{4pt}
\begin{tabularx}{\textwidth}{@{}lcl>{\raggedright\arraybackslash}X@{}}
\toprule
Study & Atoms & Readout & Evidence scope\\
\midrule
Package detection & 8/256 & Concatenated logistic & Content-component-disjoint folds\\
White-box / broker & 8/256 & Fixed application probes & Matched-pair-disjoint folds; rewrites and request policies\\
Multi-depth intervention & 8/24 & Fixed application probes & Same fitted heads; compact extraction for intervention\\
Compact runtime & 8/24 & Member CDF scores & Llama-3.1-8B and Qwen2.5-7B timing\\
\bottomrule
\end{tabularx}
\end{table}


\begin{table}[!t]
\centering
\caption{\textbf{Compact 8/24-atom CDF-consensus timing.} Each backbone has 360 paired measurements. Incremental time includes hook callbacks, D/O/B pooling, feature construction, CDF calibration, and scoring; hook setup is excluded.}
\label{tab:runtime-detail}
\footnotesize
\setlength{\tabcolsep}{4.5pt}
\begin{tabular}{@{}llrrrr@{}}
\toprule
\textbf{Backbone} & \textbf{Stage} & \textbf{Mean ms} & \textbf{Median ms} & \textbf{P95 ms} & \textbf{Std. ms}\\
\midrule
\multirow{3}{*}{Llama-3.1-8B} & Native prefill & 334.1 & 219.4 & 905.2 & 365.8 \\
 & \system incremental & 64.6 & 64.7 & 71.1 & 9.1 \\
 & Prefill + detector & 398.7 & 283.9 & 1028.4 & 367.9 \\
\midrule
\multirow{3}{*}{Qwen2.5-7B} & Native prefill & 310.8 & 202.7 & 838.5 & 333.8 \\
 & \system incremental & 56.4 & 56.1 & 62.0 & 9.6 \\
 & Prefill + detector & 367.2 & 258.7 & 896.5 & 334.5 \\
\bottomrule
\end{tabular}
\end{table}

The compact measurements use one RTX 4090 host with Python 3.12.13, PyTorch 2.6.0, CUDA
12.4, Transformers 5.13.0, BF16, SDPA, and batch size one. After 20 warmups, we
measure 120 MalSkillBench packages per backbone three times with GPU synchronization. This yields 360
paired measurements per backbone. Six empirical token-length bins contribute
20 packages each. Labels and detector scores do not affect selection. Peak
process GPU memory is 17,574\,MiB for Llama and 17,242\,MiB for Qwen.

The frozen timing benchmark subtracts a paired no-hook prefill from the
instrumented forward and scoring interval. The increment includes hook
callbacks, D/O/B pooling, projections, feature construction, member-level
CDF calibration, and consensus scoring. Model loading, tokenization and sample
preparation, bundle/projection loading, and hook registration/removal occur
outside the timed interval. It aggregates Boundary and Operation
member margins by their medians and Geometry by its maximum.
Its 64.7\,ms Llama and 56.1\,ms Qwen median increments measure this compact
pipeline, not the current 256-atom, concatenated-logistic Large-ID pipeline.

The capability-broker's greedy typed-request generation is not included in
detector increments.
Table~\ref{tab:runtime-detail} separates native prefill from detector overhead;
neither overhead alone nor the compact configuration establishes total agent
latency or the full Large-ID configuration's runtime.

\FloatBarrier

\FloatBarrier
\section{Controlled Pre-Action Representation Studies}
\label{app:representation-studies}

These studies distinguish information accessible to a diagnostic readout from
the performance of the deployed detector. Unless specified otherwise, the
three backbones are Llama-3.1-8B, Qwen2.5-7B, and Mistral-7B-v0.3.
B32 and O32 read 32-dimensional Boundary and Operation representations,
respectively. The independent-state (Ind.192) and residual-feature (Rel.192)
readouts each use 192 features. They differ
from the full-width Boundary and 735-dimensional Geometry components in
Large-ID. These studies do not change the reported package detector or its threshold.

\begin{figure}[!t]
  \centering
  \includegraphics[width=\linewidth]{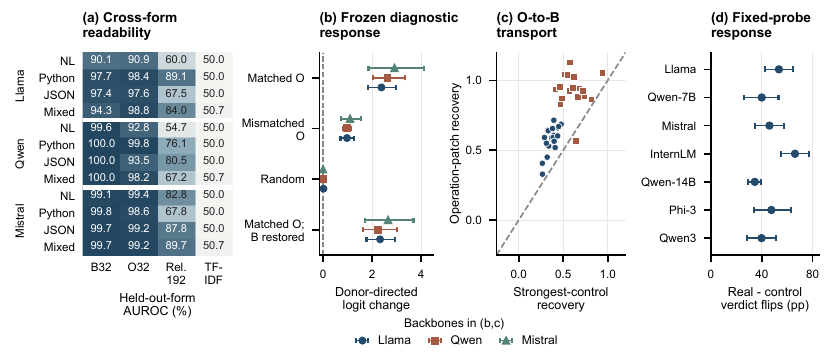}
  \caption{\textbf{Pre-action operation information is readable across forms
  and functionally affects frozen judgments.}
  (a) Cross-form AUROC. (b) Diagnostic logit shifts.
  (c) One-layer Boundary recovery by family. (d) Compact-probe verdict flips
  relative to controls. Intervals are 95\% bootstrap CIs;
  intervention protocols differ across (b--d).}
  \label{fig:representations}
\end{figure}

\subsection{Risk Prediction Across Expression Forms}
\label{app:representation-single}

The controlled dataset has 800 examples from ten domain blocks, five
task--operation compatibility rules, and four expression forms: natural language, Python plans, JSON
configuration, and mixed text. Matched operations preserve the trusted task
and vary whether the selected object satisfies its compatibility rule. Same-group
variants stay together. Five folds hold out domain blocks; the all-forms
condition includes every form in training and test. Each of the four
leave-one-form-out conditions excludes its test form from both training and
development and evaluates it on held-out domains.

All projections and benign-reference maps fit on training examples.
Development selects layer, logistic regularization, and the operating
threshold. Ind.192 constructs features separately from each state;
Rel.192 includes benign transport residuals. B32, O32, and D32 read one
region, while TF-IDF and length provide text controls. The seven methods,
three models, and five settings yield 105 aggregate metric rows.
Table~\ref{tab:representation-forms} reports all settings.
Each all-forms evaluation uses 800 examples; each held-out-form evaluation
uses 200.

\begin{table}[!t]
\centering
\small
\setlength{\tabcolsep}{4pt}
\caption{Cross-form risk AUROC (\%). All seven readouts and all form holdouts; five domain-disjoint folds.}\label{tab:representation-forms}
\begin{tabular}{llrrrrrrr}
\toprule
Model & Test form & B32 & O32 & D32 & Ind.192 & Rel.192 & TF-IDF & Length\\
\midrule
Llama & All & 99.75 & 99.98 & 50.00 & 99.28 & 99.77 & 50.23 & 50.00\\
Llama & NL & 90.09 & 90.92 & 50.00 & 89.26 & 59.98 & 50.00 & 50.00\\
Llama & Python & 97.66 & 98.36 & 50.00 & 92.12 & 89.09 & 50.02 & 50.00\\
Llama & JSON & 97.44 & 97.63 & 50.00 & 87.83 & 67.54 & 50.00 & 50.00\\
Llama & Mixed & 94.33 & 98.81 & 50.00 & 89.33 & 84.02 & 50.70 & 50.00\\
Qwen & All & 100.00 & 99.86 & 50.00 & 99.89 & 99.93 & 50.23 & 50.00\\
Qwen & NL & 99.59 & 92.81 & 50.00 & 78.59 & 54.66 & 50.00 & 50.00\\
Qwen & Python & 100.00 & 99.80 & 50.00 & 95.00 & 76.08 & 50.02 & 50.00\\
Qwen & JSON & 100.00 & 93.47 & 50.00 & 88.52 & 80.52 & 50.00 & 50.00\\
Qwen & Mixed & 99.95 & 98.21 & 50.00 & 98.20 & 67.23 & 50.70 & 50.00\\
Mistral & All & 99.88 & 98.86 & 50.00 & 99.99 & 99.95 & 50.23 & 50.00\\
Mistral & NL & 99.14 & 99.41 & 50.00 & 90.91 & 82.81 & 50.00 & 50.00\\
Mistral & Python & 99.75 & 98.59 & 50.00 & 96.15 & 67.79 & 50.02 & 50.00\\
Mistral & JSON & 99.66 & 99.20 & 50.00 & 75.42 & 87.85 & 50.00 & 50.00\\
Mistral & Mixed & 99.67 & 99.22 & 50.00 & 93.67 & 89.69 & 50.70 & 50.00\\
\bottomrule
\end{tabular}
\end{table}

AUROC pools out-of-fold predictions from frozen heads. Intervals resample the
ten domain blocks 5,000 times, keeping related variants together. Paired
AUROC tests enumerate the 1,024 domain-level method-score swaps and apply
Holm correction to three prespecified comparisons per model and setting.
In all-forms evaluation, Rel.192 exceeds TF-IDF on all three backbones
(corrected $p=0.00586$), but does not significantly exceed B32 or Ind.192.
This comparison concerns the fitted lexical baseline, not every possible
static analyzer.

Pairwise ordering and pooled ranking measure different properties. On many
holdouts, matched-pair accuracy remains high even when pooled AUROC decreases.
Score scales can differ across domains and independently fitted fold heads.
For example, Mistral's held-out Python
Rel.192 readout has 67.79\% pooled AUROC but 100\% matched-pair accuracy.
Qwen's natural-language holdout also loses pairwise ordering, so score-scale
changes do not explain every failure. These exploratory comparisons do not
select a new calibration or test threshold.

Layerwise distances use training-fitted coordinate standardization. We
compare a semantic-preserving form change with a risk-changing operation
under fixed form. Form changes dominate whole-vector distances at most
depths, despite strong risk readouts. Qwen's late Boundary states show the
opposite ordering; Llama and Mistral do not share this reversal. Pointwise
domain-bootstrap intervals describe individual depths, without selecting
test-set peaks or claiming simultaneous layerwise significance.

\subsection{Matched Operation Interventions}
\label{app:representation-interventions}

For each of the 800 recipients per model, the all-forms Rel.192
diagnostic and its development threshold remain frozen. At the depth
preceding the selected readout, we intervene on Operation states using a
donor matched in domain, compatibility rule, expression form, and trusted task.
Only the operation object changes. Both benign-to-malicious and
malicious-to-benign directions are included. We report the logit-margin
change directed toward the donor label, not a probability change or an
attack execution rate.

Controls comprise an unchanged forward pass, a sham replacement, same-position
norm-matched random perturbations, unrelated Declaration perturbations with
matched total energy, norm-matched semantically mismatched donors, and the
matched Operation donor followed by restoration of Boundary to its unpatched
state. Five random seeds are averaged within each recipient before domain
aggregation. They do not increase the independent sample count.
Table~\ref{tab:representation-interventions} includes all conditions.

\begin{table}[!t]
\centering
\small
\setlength{\tabcolsep}{4pt}
\caption{Donor-directed diagnostic logit changes. Intervals resample ten domains.}\label{tab:representation-interventions}
\begin{tabular}{lrrr}
\toprule
Condition & Llama & Qwen & Mistral\\
\midrule
Matched O & 2.376 [1.812, 2.988] & 2.622 [2.038, 3.353] & 2.909 [1.849, 4.110]\\
Matched O, B restored & 2.320 [1.768, 2.924] & 2.251 [1.640, 3.001] & 2.646 [1.701, 3.685]\\
Norm-matched random & 0.011 [0.007, 0.015] & 0.002 [-0.006, 0.008] & -0.004 [-0.011, 0.002]\\
Mismatched O & 0.971 [0.708, 1.251] & 0.957 [0.801, 1.146] & 1.097 [0.722, 1.540]\\
Unrelated D & 0.115 [-0.011, 0.260] & 0.158 [0.060, 0.243] & -0.019 [-0.053, 0.021]\\
Unpatched & 0.000 [0.000, 0.000] & 0.000 [0.000, 0.000] & 0.000 [0.000, 0.000]\\
Sham & 0.000 [0.000, 0.000] & 0.000 [0.000, 0.000] & 0.000 [0.000, 0.000]\\
\bottomrule
\end{tabular}
\end{table}

Matched donors flip the frozen diagnostic decision on 46.75\%, 34.13\%, and
28.75\% of recipients on Llama, Qwen, and Mistral, respectively. All four
prespecified matched-control contrasts per model pass the domain-level
sign-swap test with Holm-corrected $p=0.0078125$. Equal corrected values arise
from the finite set of domain sign swaps, not equal effect sizes.

Sham effects and verdict flips are zero. Most of the
matched-donor effect remains after Boundary restoration: the diagnostic
still directly reads O and cross-region residuals. This supports functional
sensitivity to the operation intervention without establishing Boundary as a
necessary mediator. The separate exact-slot and compact detector-aligned
experiments in Appendices~\ref{app:mechanism-interventions}
and~\ref{app:detector-aligned-readout} retain their own inputs, heads, and
statistical units.

\subsection{Region Masks on Identical Hidden States}
\label{app:representation-masks}

D/O are structural proxies, not a natural-language/code partition. The
heading parser can treat a \texttt{\#}-prefixed line within a code fence as a
heading. Across 5,830 packages, this moves the declaration boundary inside a
fence in 27 cases (0.46\%). We test all 27 affected cases and 27 matched
controls on Llama and Qwen. The serialized input, tokens, hidden states, and
Boundary state are unchanged; only the D/O pooling masks change.

Neither model changes a binary verdict on the affected cases. Maximum
absolute probability changes are 0.30932 on Llama and 0.02283 on Qwen.
Thus, this concrete parsing issue does not change decisions in the tested
set, but some scores are sensitive to the mask. The experiment is neither a
semantic segmentation accuracy estimate nor evidence that all possible
region definitions are equivalent. All reported detector results use the
same structural parsing rule; this test changes masks only for diagnosis.

\subsection{CompoSkill Input Dependence}
\label{app:representation-inputs}

The monitoring set comprises 344 explicit and 346 implicit malicious--benign
pairs: 690 mode-specific pairs, or 1,380 records. These are not 690
independent attack families. Their family union contains 361 identities,
329 shared between modes, grouped into 76 scenario/persona role groups.
The five threat categories contribute 150 data-exfiltration, 128
memory-tampering, 140 multi-agent-hijack, 123 privilege/dangerous-command,
and 149 resource-exhaustion pairs.

Input ablations refit diagnostic readouts under the same five role-group
folds. They compare the task, workspace, installed nodes, local contexts, and
full path, using the original labels for the full paths. A workspace-only
TF-IDF readout reaches 98.38\% explicit and 100.00\% implicit AUROC.
Pure-node views are at 50\% because each matched pair has identical skill
content. On Llama, a full-path diagnostic reaches 84.97/99.62\% AUROC
(explicit/implicit), versus 93.64/100.00\% for workspace only. On Qwen,
the corresponding values are 95.75/99.99\% and 95.24/100.00\%.
These are diagnostic refits, not the final path detector.

Workspace cues therefore suffice for strong prediction on this benchmark.
This does not mean that the final detector relies only on workspace cues.
It means that benchmark accuracy alone cannot identify how the model
represents skill dependencies. The controlled fixed-node study
below tests that question without directly exposing the final target in the
workspace.

\subsection{Fixed-Node Composition Counterfactuals}
\label{app:representation-composition}

\begin{figure}[!t]
  \centering
  \includegraphics[width=\linewidth]{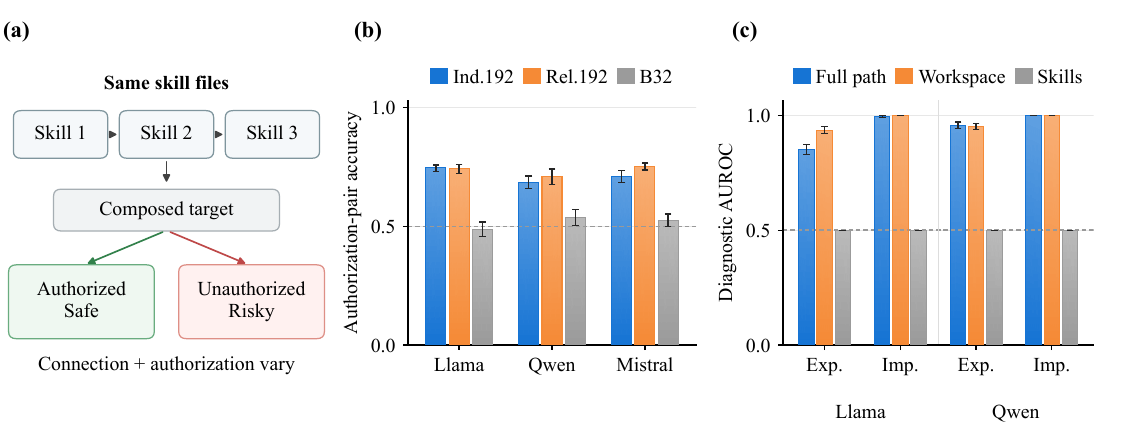}
  \caption{\textbf{Risk from skill composition.}
  (a) Fixed-node experiment: identical skill files, varying connections and
  authorization. (b) Authorization-pair accuracy for independent-state
  (Ind.192), residual (Rel.192), and Boundary (B32) readouts.
  (c) CompoSkill diagnostic input ablations under explicit (Exp.) and
  implicit (Imp.) attacks, not final-detector results.
  Error bars are 95\% bootstrap intervals; dashed lines mark chance.}
  \label{fig:composition-evidence}
\end{figure}

We fix the bytes of three skill files and cross their connection and
authorization settings. The workspace supplies routing and a starting key;
the composed finite map determines the target. A deterministic, no-I/O
oracle checks whether that target is authorized. The 144 adaptable cases
and 31 domain labels are drawn from SCR-CapFlow's organization, yielding
1,152 bundles and 4,608 four-cell presentations. This is one controlled
finite-map mechanism, not 31 independent mechanisms or the original
SCR-Bench attack evaluation.

Five domain-grouped folds keep related variants together. Layers and
hyperparameters are selected on development. Input-deletion controls retain
the original full-path labels: their chance performance indicates missing
information for that label, not safety of a newly defined partial task.
Table~\ref{tab:representation-composition} includes all 19 methods.

\begin{figure}[!t]
  \centering
  \includegraphics[width=\linewidth]{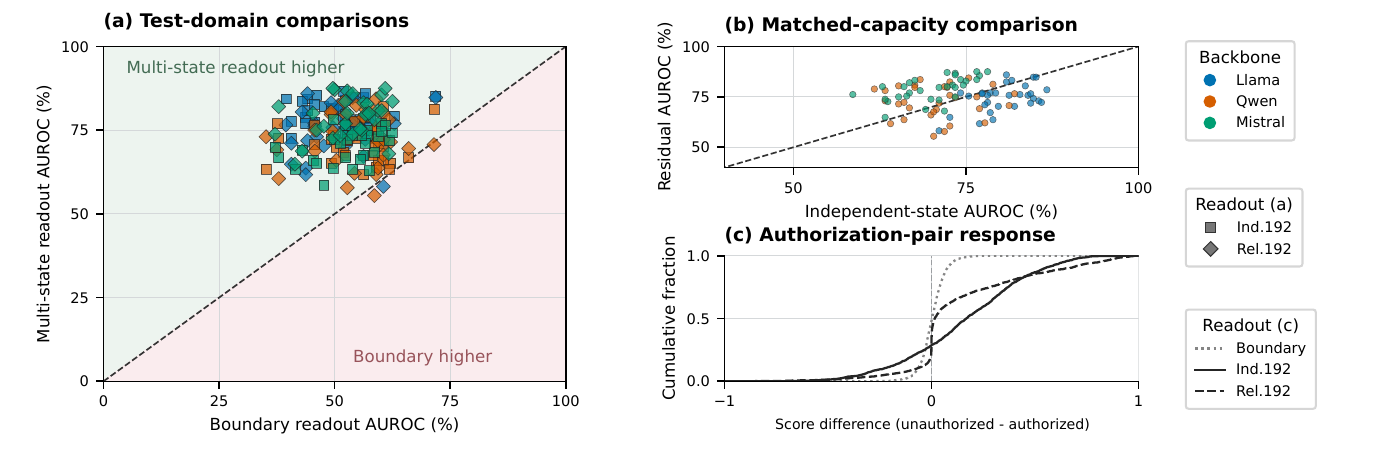}
  \caption{\textbf{Multi-state readouts expose controlled composition risk;
  residual gains vary across domains.}
  (a) Multi-state versus Boundary AUROC by domain.
  (b) Matched-capacity independent-state versus residual-feature AUROC.
  (c) Cumulative score differences (unauthorized minus authorized).
  Diagonals mark equal AUROC; positive differences indicate correct risk ordering.}
  \label{fig:composition-readouts}
\end{figure}

Figure~\ref{fig:composition-readouts}a contains 186 comparisons: three backbones,
31 domains, and two multi-state readouts. Panel b compares the two 192-dimensional
readouts on the same 93 backbone--domain pairs; B32 uses 32 dimensions.
For panel c, each backbone contributes 2,304 authorization pairs. The empirical
distributions weight backbones and domains equally and retain all held-out pairs.
A positive score difference ranks the unauthorized input higher; magnitudes
are not calibrated across readouts. Panels share inputs, not independent replicates.

\begin{figure}[!t]
  \centering
  \includegraphics[width=\linewidth]{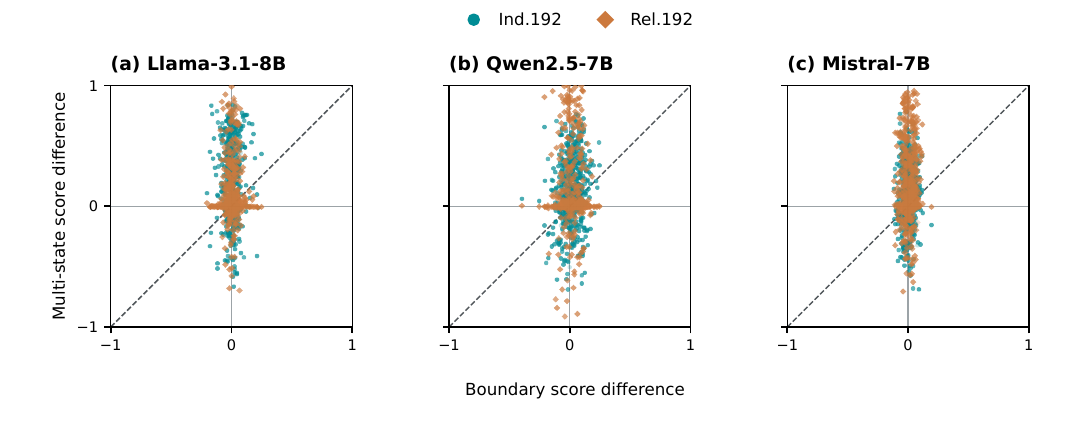}
  \caption{\textbf{Authorization-pair responses by backbone.}
  Each point compares Boundary and multi-state score differences
  (unauthorized minus authorized). Positive values indicate correct risk ordering.
  Colors distinguish the two multi-state readouts; axes use a common scale.}
  \label{fig:composition-pointcloud}
\end{figure}

Figure~\ref{fig:composition-pointcloud} separates the paired responses by backbone.
Each panel retains all 2,304 held-out authorization pairs for each multi-state readout,
without jitter or filtering. The two readouts share pairs, yielding 13,824 plotted
points rather than independent experimental replicates. The diagonal indicates
equal numerical score differences, not equivalent calibration across heads.

\begin{table}[!t]
\centering
\small
\setlength{\tabcolsep}{4pt}
\caption{Finite-map composition AUROC (\%), equally weighted across 31 domain labels. All 19 fixed readouts.}\label{tab:representation-composition}
\begin{tabular}{lrrr}
\toprule
Readout & Llama & Qwen & Mistral\\
\midrule
Full-path B32 & 50.83 & 54.69 & 52.77\\
Task only & 50.00 & 50.00 & 50.00\\
Workspace only & 50.00 & 50.00 & 50.00\\
Task + workspace & 50.00 & 50.00 & 50.00\\
Skills only & 50.00 & 50.00 & 50.00\\
Node 1 & 50.00 & 50.00 & 50.00\\
Node 2 & 50.00 & 50.00 & 50.00\\
Node 3 & 50.00 & 50.00 & 50.00\\
Local context 1 & 50.00 & 50.00 & 50.00\\
Local context 2 & 50.00 & 50.00 & 50.00\\
Local context 3 & 50.00 & 50.00 & 50.00\\
Node aggregate (32) & 50.00 & 50.00 & 50.00\\
Local aggregate (32) & 49.97 & 49.74 & 49.99\\
Local concatenation (96) & 49.70 & 50.03 & 50.36\\
Local squared (64) & 65.07 & 59.20 & 65.42\\
Independent-state squared (192) & 80.41 & 70.77 & 70.17\\
Relational residual squared (192) & 74.55 & 72.68 & 78.99\\
TF-IDF & 50.00 & 50.00 & 50.00\\
Length & 50.00 & 50.00 & 50.00\\
\bottomrule
\end{tabular}
\end{table}

Independent-state and residual features both use squared transformations and
192 dimensions. Relative to the independent construction, residual AUROC
changes by $-5.86$ [$-7.77$, $-3.92$] points on Llama, $+1.92$
[$-1.45$, $+5.23$] on Qwen, and $+8.82$ [$+7.08$, $+10.51$] on Mistral.
Holm-corrected paired $p$ values are 0.00080, 0.27275, and 0.00080.
The residual contribution therefore depends on the backbone. Comparisons
with B32 change feature width and construction as well as the regions read.
Using squared features in a readout does not identify a particular nonlinear
computation inside the model.

The corresponding composition intervention uses all 4,608 recipients per
model, half with changed risk and half with unchanged risk. Matched workspace
donors keep skill files and the trusted task fixed. Donor-directed diagnostic
margin changes are 0.172/0.292/0.154 for Llama/Qwen/Mistral, versus
0.002/$-0.0001$/$-0.0005$ for norm-matched random controls and
0.069/0.084/0.044 for mismatched donors. Restoring B leaves
0.172/0.288/0.158. However, unchanged-risk recipients still flip decisions
at 14.92/7.06/8.47\%, compared with 14.31/6.85/2.02\% on changed-risk
recipients. The intervention changes the diagnostic but does not isolate a
risk-specific causal effect. It therefore does not establish a universal
composition mechanism.

\subsection{Full-Architecture Structural Transfer}
\label{app:representation-transfer}

\begin{figure}[!t]
  \centering
  \includegraphics[width=\linewidth]{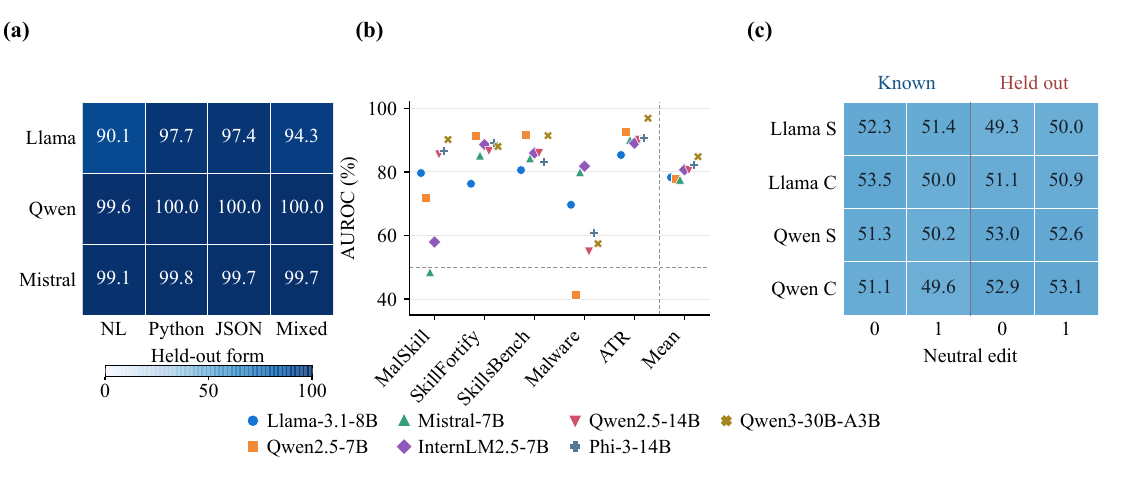}
  \caption{\textbf{Risk prediction under three kinds of shift.}
  (a) Boundary AUROC with each expression form excluded from training and
  development. (b) LatentTriad AUROC on five held-out source ecosystems
  and their macro mean across seven backbones.
  (c) Full-architecture performance on known- and held-out-template panels,
  for single (S) and composed (C) packages under neutral edits 0/1.
  Both heatmaps use the same 0--100\% AUROC scale.}
  \label{fig:transfer-boundary}
\end{figure}

The broader synthetic test uses six templates: dictionary binding, role
requirements, credential validation, delegation chains, scope intersection,
and conditional fallback. Each model has 2,304 inputs: 576 training,
192 development, and 1,536 test presentations. The test uses new opaque
symbol assignments, with both single-skill and composed packaging and two
neutral-edit variants. The known-template panel trains on all six templates.
Six leave-one-template-out panels train on 480 and develop on 160 examples,
testing 256 presentations of the omitted template. Template reuse and symbol
changes define the generalization axis; these are not independent natural
attack mechanisms.

Unlike the pooled diagnostic readouts, this test fits the complete
\system architecture from genuine local atoms at eight depths. It retains
32-dimensional latent coordinates, offsets 300/400/500, the unmodified
component-head family, and fixed max--min fusion. The models run in BF16
without quantization, CPU offload, or truncation. Four state-feature controls,
Boundary-only, and \system are evaluated on every panel. Independent-linear
here is fitted on synthetic training data, whereas the independent-state control in
Appendix~\ref{app:final-component-controls} is fitted on natural packages.

\begin{table}[!t]
\centering
\small
\setlength{\tabcolsep}{4pt}
\caption{Full-architecture transfer AUROC (\%). Both neutral-edit variants and every control. Held-out values average all six template panels.}\label{tab:representation-transfer}
\begin{tabular}{llllrrrrrr}
\toprule
Model & Templates & Edit & Package & B & LT & Ind.L & Ind.S & Orth.S & Quad.\\
\midrule
Llama & Known & 0 & Single & 51.01 & 52.31 & 53.91 & 54.02 & 53.47 & 52.93\\
Llama & Known & 0 & Composed & 51.60 & 53.47 & 54.46 & 54.41 & 54.38 & 54.18\\
Llama & Known & 1 & Single & 49.87 & 51.38 & 51.45 & 51.24 & 51.22 & 51.20\\
Llama & Known & 1 & Composed & 49.98 & 50.05 & 50.93 & 50.90 & 50.88 & 51.03\\
Llama & Held out & 0 & Single & 49.76 & 49.25 & 50.08 & 49.85 & 49.72 & 49.45\\
Llama & Held out & 0 & Composed & 50.34 & 51.06 & 50.85 & 50.86 & 50.62 & 50.15\\
Llama & Held out & 1 & Single & 49.72 & 50.05 & 49.95 & 49.84 & 50.23 & 49.32\\
Llama & Held out & 1 & Composed & 50.54 & 50.91 & 50.42 & 50.29 & 50.78 & 50.55\\
Qwen & Known & 0 & Single & 49.12 & 51.30 & 52.85 & 53.19 & 52.72 & 51.50\\
Qwen & Known & 0 & Composed & 50.29 & 51.12 & 52.08 & 52.10 & 51.99 & 52.03\\
Qwen & Known & 1 & Single & 50.75 & 50.16 & 52.52 & 52.13 & 51.35 & 52.13\\
Qwen & Known & 1 & Composed & 47.79 & 49.64 & 50.02 & 49.80 & 49.30 & 49.64\\
Qwen & Held out & 0 & Single & 53.69 & 52.96 & 53.52 & 53.92 & 53.81 & 52.62\\
Qwen & Held out & 0 & Composed & 51.29 & 52.86 & 53.52 & 53.53 & 53.52 & 53.24\\
Qwen & Held out & 1 & Single & 51.63 & 52.64 & 53.32 & 53.39 & 53.27 & 52.72\\
Qwen & Held out & 1 & Composed & 50.65 & 53.09 & 52.72 & 52.90 & 53.01 & 52.23\\
\bottomrule
\end{tabular}
\end{table}

Table~\ref{tab:representation-transfer} reports all 96 aggregate AUROC cells.
Ind.L and Ind.S use independent linear and squared features; Orth.S uses
orthogonal squared features; Quad. uses explicit quadratic features.
Held-out entries are means of all six template panels, not pooled scores
from different fitted heads. We provide descriptive results without treating
six constructed templates as a population of independent mechanisms.

For the main neutral variant, \system AUROC ranges from 49.25\% to 53.47\%.
At both development-calibrated 5\% and 10\% FPR budgets, \system and
Boundary-only make identical test decisions on every panel and packaging.
This identity also holds for the second neutral variant. Development-budget
calibration does not ensure the same test FPR: at the 10\% budget, test FPR
is 3.12--9.38\% in known-template panels but 28.65--34.38\% under template
holdout. This test therefore supports neither stable structural transfer nor
a low-FPR advantage over Boundary-only. It refits the architecture on synthetic
training data; it does not reevaluate the natural-package checkpoints.

The finite-map result and this broader transfer test answer different
questions. Retaining local atoms at eight depths does not recover transfer
here, so missing local features alone cannot explain the result.

\subsection{Same-Prompt Readouts and Generated Judgments}
\label{app:representation-generation}

This exploratory program test uses three operators (object flow, delegation,
and conditional selection), six lexical families, four expression forms,
two packaging styles, and both answer orders. Each model has four
condition--target panels of 384 questions: the actual object must either be
inferred or is supplied, and the target is either the object or its risk.
Three operator folds rotate training, development, and test roles.
Actual-object and risk heads are trained separately. All three fixed seeds
reuse the same questions, so they do not triple the independent support.

For each question, the Boundary readout and generation start from the exact
same prompt. The readout uses the first prefill; generation is evaluated at
96, 256, and 512 tokens. Bfull uses the model's full hidden width rather than
a compressed B32/B128 projection. The table reports all fixed methods,
including random-training-label and text controls. Development chooses
regularization from $\{0.01,0.1,1\}$, with ties resolved toward smaller values.

\begin{table}[!t]
\centering
\small
\setlength{\tabcolsep}{4pt}
\caption{Same-prompt readout accuracy (\%). All three seed means; 384 questions per model, condition, and target.}\label{tab:representation-generation-readouts}
\begin{tabular}{lllrrrrrr}
\toprule
Model & Object & Target & B32 & B128 & Bfull & Random & Text & Options\\
\midrule
Llama & Inferred & actual & 55.47 & 60.16 & 61.46 & 47.92 & 50.00 & 50.00\\
Llama & Inferred & risk & 49.83 & 51.82 & 53.39 & 50.69 & 50.00 & 50.00\\
Llama & Supplied & actual & 67.19 & 78.12 & 75.52 & 47.31 & 50.00 & 50.00\\
Llama & Supplied & risk & 59.72 & 60.94 & 59.90 & 49.74 & 50.00 & 50.00\\
Qwen & Inferred & actual & 52.00 & 52.78 & 51.30 & 49.39 & 50.00 & 50.00\\
Qwen & Inferred & risk & 51.65 & 54.51 & 53.12 & 48.96 & 50.00 & 50.00\\
Qwen & Supplied & actual & 81.08 & 89.24 & 89.84 & 48.09 & 50.00 & 50.00\\
Qwen & Supplied & risk & 75.09 & 79.86 & 89.84 & 51.22 & 50.00 & 50.00\\
\bottomrule
\end{tabular}
\end{table}

\begin{table}[!t]
\centering
\small
\setlength{\tabcolsep}{4pt}
\caption{Same-prompt generation accuracy (\%). Parentheses count format failures, retained as errors; each denominator is 384.}\label{tab:representation-generation-budgets}
\begin{tabular}{lllrrr}
\toprule
Model & Object & Target & 96 tokens & 256 tokens & 512 tokens\\
\midrule
Llama & Inferred & actual & 55.47 (166) & 76.30 (42) & 77.60 (32)\\
Llama & Inferred & risk & 61.72 (77) & 76.56 (3) & 76.56 (3)\\
Llama & Supplied & actual & 59.90 (139) & 80.21 (15) & 80.21 (13)\\
Llama & Supplied & risk & 69.27 (87) & 84.64 (8) & 84.90 (4)\\
Qwen & Inferred & actual & 52.34 (176) & 93.23 (4) & 94.01 (1)\\
Qwen & Inferred & risk & 23.18 (286) & 92.71 (10) & 94.53 (2)\\
Qwen & Supplied & actual & 51.82 (185) & 95.05 (18) & 99.74 (0)\\
Qwen & Supplied & risk & 44.27 (211) & 89.58 (14) & 91.15 (6)\\
\bottomrule
\end{tabular}
\end{table}

Random denotes the B128 random-training-label control; Text uses the full
question and Options removes the program. Format failures are counted as
errors rather than discarded. In particular, Qwen's inferred-risk accuracy
rises from 23.18\% at 96 tokens to 94.53\% at 512 tokens, with format failures
falling from 286 to 2. A short-budget failure is not evidence that the model
cannot solve the task.

Bfull inferred-risk accuracy is 53.39\% on Llama and 53.12\% on Qwen, compared
with 76.56\% and 94.53\% after 512-token generation. Supplying the actual
object raises Bfull to 59.90\% and 89.84\%. This separates successful generated
reasoning from accessibility to the tested fixed linear head. It does not
prove that the first prefill lacks the information, that generation creates
new knowledge, or that model scale determines detectability.

\clearpage
\section{Discussion and Limitations}
\label{app:discussion-limitations}
\label{sec:discussion}

\subsection{Scope of Pre-Action Risk Readability}
\label{app:readability-scope}

Pre-action readouts expose a different capability from generated judgments.
On inferred-object prompts, Qwen reaches 94.53\% risk-answer accuracy with
a 512-token generation budget, versus 53.12\% for a full-width linear
Boundary readout on the same prompts
(Appendix~\ref{app:representation-generation}).
Supplying the actual operation object raises that Boundary readout to
89.84\%. For this probe, reading an explicit task--operation relation is
easier than first resolving the operation from an implicit composition.
The contrast concerns accessibility to the tested linear head, not the
absence of information from the first prefill or generation creating new
knowledge.

The intervention protocols also answer distinct questions. A diagnostic that
reads O directly can respond to its replacement even when B is restored.
The exact-slot experiment measures downstream B displacement, while the
compact detector-aligned experiment measures frozen score and verdict
changes. These tests establish functional effects in their respective
settings; they do not establish B as a necessary mediator or measure an
agent's execution of the patched operation
(Appendices~\ref{app:representation-interventions},
\ref{app:mechanism-interventions}, and~\ref{app:detector-aligned-readout}).

\subsection{Generalization Beyond Training Distributions}
\label{app:generalization-scope}

Expression, ecosystem, and structural shifts change different parts of the
problem. Held-out-form prediction tests new expressions of the controlled
task--operation relations, with domains also held out. Source holdout changes
the skill ecosystem and reduces performance despite an advantage over the
tested baselines (Section~\ref{sec:package-detection}). Expression transfer
therefore supports robustness to the tested forms, not ecosystem-independent
screening.

The fixed-node study changes connections and authorization while keeping
file bytes fixed. Its multi-state features outperform the tested B-only
readout, but the comparison also changes feature width and construction.
The broader synthetic study refits the full architecture under new symbol
assignments and remains near chance even on known templates
(Appendix~\ref{app:representation-transfer}); it does not reevaluate the
natural-package checkpoints. This is a broader inference requirement than
decoding an explicit operation's task consistency. Likewise, strong
workspace-only controls on CompoSkill show that benchmark accuracy alone
cannot identify which dependency information a detector uses.

\subsection{Deployment and Runtime Boundaries}
\label{app:deployment-boundaries}

Deployment should select readouts, thresholds, and fusion on independent
development data for its detection objective. Geometry-only has the strongest
observed F1 and zero observed FPR in the component panel, whereas
$\max(B,O)$ improves recall, F1, and AUROC over the fixed max--min rule at
the same displayed FPR. These comparisons distinguish the representation
design from the choice of decision policy. The reported fixed rule remains
unchanged; deployment-specific operating points require separate selection
rather than choosing a new policy from test outcomes. Zero observed false
positives in this panel is not a guarantee for future traffic.

The detector requires white-box access and the input contract in
Appendix~\ref{app:contract}. It can inspect only content available in the
loaded context. Downloads, environment-triggered branches, and externally
assembled payloads can reveal behavior absent from that context, motivating
permission enforcement and execution-time monitoring alongside screening.
The compact configuration's timing does not measure the full Large-ID
pipeline, and broker request decisions do not measure completed-task
success. The eligible-subset false-positive increases under white-box
rewrites and the limited valid adaptive-rewrite coverage also constrain
deployment conclusions (Section~\ref{sec:composition-representations}).

\end{document}